\documentclass[11pt]{article}
\usepackage[margin=1in]{geometry}
\usepackage{amsmath, amssymb}
\usepackage[colorlinks=true, linkcolor=black, citecolor=blue, urlcolor=blue]{hyperref}
\usepackage{natbib}
\usepackage{booktabs}
\usepackage{multirow}
\usepackage{graphicx}
\usepackage{amssymb}
\usepackage{longtable}
\usepackage{array}
\usepackage{authblk}

\title{\LARGE Alignment and Safety in Large Language Models:  Safety Mechanisms, Training Paradigms, and Emerging Challenges}

\author[1]{Haoran Lu}
\author[1]{Luyang Fang}
\author[2]{Ruidong Zhang}
\author[2]{Xinliang Li}
\author[1]{Jiazhang Cai}
\author[3]{Huimin Cheng}
\author[3]{Lin Tang}
\author[1]{Ziyu Liu}
\author[4]{Zeliang Sun}
\author[1]{Tao Wang}
\author[1]{Yingchuan Zhang}
\author[5]{Arif Hassan Zidan}
\author[6]{Jinwen Xu}
\author[7]{Jincheng Yu}
\author[1]{Meizhi Yu}
\author[2]{Hanqi Jiang}
\author[1]{Xilin Gong}
\author[2]{Weidi Luo}
\author[8]{Bolun Sun} 
\author[9]{Yongkai Chen} 
\author[10]{Terry Ma} 
\author[1]{Shushan Wu}

\author[2]{Yifan Zhou}
\author[2]{Junhao Chen}
\author[6]{Haotian Xiang}
\author[11]{Jing Zhang} 
\author[5]{Afrar Jahin}
\author[2]{Wei Ruan}
\author[2]{Ke Deng}
\author[2]{Yi Pan}
\author[12]{Peilong Wang} 
\author[7]{Jiahui Li}
\author[2]{Zhengliang Liu}

\author[13]{Lu Zhang}
\author[14]{Xiaobo Li}
\author[14]{Lin Zhao}
\author[12]{Wei Liu}
\author[11]{Dajiang Zhu}
\author[15]{Xin Xing}
\author[7]{Fei Dou}
\author[5]{Wei Zhang}

\author[4]{Chao Huang}
\author[1]{Rongjie Liu}
\author[16]{Mengrui Zhang}
\author[17]{Yiwen Liu}
\author[17]{Xiaoxiao Sun}

\author[6]{Qin Lu}
\author[2]{Zhen Xiang}
\author[1]{Wenxuan Zhong\textsuperscript{*}}
\author[2]{Tianming Liu\textsuperscript{*}}
\author[1]{Ping Ma\textsuperscript{*}}

\affil[1]{Department of Statistics, University of Georgia, Athens, GA}
\affil[2]{School of Computing, University of Georgia, Athens, GA}
\affil[3]{Department of Biostatistics, Boston University, Boston, MA}
\affil[4]{Department of Epidemiology \& Biostatistics, University of Georgia, Athens, GA}
\affil[5]{School of Computer and Cyber Sciences, Augusta University, Augusta, GA}
\affil[6]{School of Electrical and Computer Engineering, University of Georgia, Athens, GA}
\affil[7]{Department of Statistics \& Data Science, University of Arizona, Tucson, AZ}
\affil[8]{Kellogg School of Management, Northwestern University, Evanston, IL} 
\affil[9]{Department of Statistics, Harvard University, Cambridge, MA} 
\affil[10]{School of Computer Science, Carnegie Mellon University, Pittsburgh, PA} 
\affil[11]{School of Computer Science and Engineering, University of Texas at Arlington, TX} 
\affil[12]{Department of Radiation Oncology, Mayo Clinic Arizona, Phoenix, AZ} 
\affil[13]{Computer Science Department, Indiana University Indianapolis, IN} 
\affil[14]{Department of Biomedical Engineering, New Jersey Institute of Technology, NJ} 
\affil[15]{Department of Statistics, Virginia Tech, Blacksburg, VA} 
\affil[16]{Quantitative Sciences Unit, Department of Medicine, Stanford University, Stanford, CA} 
\affil[17]{Department of Epidemiology and Biostatistics, Mel and Enid Zuckerman College of Public Health, University of Arizona, Tucson, AZ} 

\date{}

\renewcommand{\hat}{\widehat}

\def\a\cos{\mathrm{arc\cos}}

\begin{document}

\maketitle

\vspace{-1em}
\noindent\textsuperscript{*}\textbf{Corresponding author(s).} E-mail(s): 
\href{mailto:wenxuan@uga.edu}{wenxuan@uga.edu}; 
\href{mailto:tliu@uga.edu}{tliu@uga.edu}; 
\href{mailto:pingma@uga.edu}{pingma@uga.edu}

\begin{abstract}
Due to the remarkable capabilities and growing impact of large language models (LLMs), they have been deeply integrated into many aspects of society. Thus, ensuring their alignment with human values and intentions has emerged as a critical challenge. This survey provides a comprehensive overview of practical alignment techniques, training protocols, and empirical findings in LLM alignment. We analyze the development of alignment methods across diverse paradigms, characterizing the fundamental trade-offs between core alignment objectives. Our analysis shows that while supervised fine-tuning enables basic instruction-following, preference-based methods offer more flexibility for aligning with nuanced human intent. We discuss state-of-the-art techniques, including Direct Preference Optimization (DPO), Constitutional AI, brain-inspired methods, and alignment uncertainty quantification (AUQ), highlighting their approaches to balancing quality and efficiency. We review existing evaluation frameworks and benchmarking datasets, emphasizing limitations such as reward misspecification, distributional robustness, and scalable oversight. We summarize strategies adopted by leading AI labs to illustrate the current state of practice. We conclude by outlining open problems in oversight, value pluralism, robustness, and continuous alignment. This survey aims to inform both researchers and practitioners navigating the evolving landscape of LLM alignment.
\end{abstract}

\tableofcontents

\section{Introduction}
The alignment of large language models (LLMs) with human values, intentions, and preferences represents one of the most critical challenges in contemporary artificial intelligence (AI) research. As LLMs achieve unprecedented capabilities across diverse domains, from natural language understanding to complex reasoning tasks, ensuring their reliable and beneficial deployment has become a fundamental requirement for the continued advancement of AI technology. The field of LLM alignment has emerged from the intersection of theoretical AI safety research and practical machine learning, evolving rapidly from speculative concerns to deployed systems that impact millions of users worldwide.

The conceptual foundations of AI alignment trace back to early theoretical work in AI safety, most notably Bostrom's seminal analysis of superintelligence risks and the orthogonality thesis \citep{mulgan2016superintelligence}, and the Machine Intelligence Research Institute's foundational research on friendly AI \citep{yudkowsky2008artificial}. These early contributions established crucial theoretical frameworks, defining the fundamental alignment challenge: ensuring that increasingly capable AI systems remain aligned with human objectives as their capabilities scale. The translation of these abstract concerns into concrete research directions was catalyzed by 
\cite{amodei2016concrete}, which identified persistent technical challenges, including reward hacking, scalable oversight, and distributional robustness, that continue to guide contemporary research.

While transformer architecture \citep{vaswani2017attention} revolutionized natural language processing (NLP) and enabled modern LLMs, their unprecedented scaling revealed new alignment challenges.
The subsequent progression from BERT \citep{devlin2019bert} through GPT-2 \citep{radford2019language} to GPT-3 \citep{brown2020language} demonstrated 
extraordinary
emergent capabilities that were not explicitly programmed into these systems. GPT-3's remarkable few-shot learning abilities and general-purpose language understanding capabilities marked a paradigm shift, revealing that sufficiently large language models could exhibit behaviors and competencies far beyond their training objectives. The discovery of emergent abilities at scale \citep{wei2022emergent}  and the establishment of neural scaling laws \citep{kaplan2020scaling} provided both a roadmap for future capability improvements and a compelling motivation for alignment research, as these developments suggested that model behaviors could become increasingly difficult to predict and control.

The practical breakthrough in LLM alignment was achieved through the development of Reinforcement Learning from Human Feedback (RLHF), most prominently demonstrated in OpenAI's InstructGPT models \citep{ouyang2022training}. This foundational work established the three-stage alignment pipeline that has become the industry standard: supervised fine-tuning on human demonstrations, reward model training on human preference rankings, and policy optimization through reinforcement learning. The dramatic empirical success of this approach, demonstrating that a 1.3-billion parameter InstructGPT model was preferred over the 175-billion parameter GPT-3 despite being over 100 times smaller, validated the hypothesis that alignment techniques could be more important than raw computational scale for producing useful and safe AI systems.

Subsequent innovations have refined and extended these foundational techniques. Anthropic's Constitutional AI \citep{bai2022constitutional} introduced a scalable approach to alignment through AI-generated feedback guided by explicit constitutional principles, reducing dependence on human annotation while improving harmlessness. The development of Direct Preference Optimization \citep{rafailov2023direct} provided an elegant mathematical insight that the optimal RLHF policy could be derived in closed form, eliminating the complex reinforcement learning pipeline and significantly improving training stability and efficiency. These methodological advances have been complemented by emerging research directions, including mechanistic interpretability, which aims to understand the internal representations and computations of neural networks \citep{elhage2021mathematical}, and scalable oversight techniques designed to address the fundamental challenge of supervising AI systems that may exceed human capabilities \citep{christiano2018supervising}.

Contemporary LLM alignment research is characterized by both remarkable progress and persistent challenges. Major AI research organizations, including OpenAI, Anthropic, Google DeepMind, and Meta, have successfully deployed aligned language models at scale, demonstrating that alignment techniques can be effectively integrated into practical systems. However, significant limitations remain: current alignment methods exhibit brittleness to adversarial attacks, suffer from distribution shift, and may not scale effectively to future systems with superhuman capabilities. The field continues to grapple with fundamental questions about the nature of human values, the scalability of human oversight, and the robustness of alignment techniques across diverse deployment contexts.

This survey provides a comprehensive examination of the current state of LLM alignment research, synthesizing theoretical foundations, practical techniques, and empirical findings across the rapidly evolving field. We organize our analysis around the core technical challenges of alignment: defining and measuring alignment objectives, developing effective training methodologies, ensuring robustness and generalization, and scaling alignment techniques to increasingly capable systems. Our coverage encompasses supervised fine-tuning approaches, reinforcement learning from human feedback, constitutional and rule-based methods, preference optimization techniques, and emerging directions including mechanistic interpretability and scalable oversight.

The structure of this survey reflects the multifaceted nature of the alignment challenge. Section 2 establishes the fundamental objectives that define aligned behavior and examines the complex trade-offs between competing goals such as helpfulness, harmlessness, and honesty. Section 3 reviews evaluation methodologies and benchmarking approaches that enable systematic assessment of alignment quality. Sections 4 and 5 provide a detailed analysis of the two dominant training paradigms: supervised fine-tuning and reinforcement learning from human feedback, including their theoretical foundations, practical implementations, and empirical performance. Section 6 examines the relationships and complementary roles of these approaches within integrated training pipelines.

Advanced alignment techniques are covered in Section 7, including direct preference optimization, AI-assisted alignment, and multi-agent approaches. Section 8 reviews parameter-efficient fine-tuning methods that enable scalable deployment of alignment techniques. Section 9 explores emerging directions, including brain-inspired approaches and neurosymbolic methods. Section 10 addresses the critical challenge of uncertainty quantification in alignment, while Section 11 examines societal, ethical, and regulatory considerations. Section 12 surveys alignment strategies across leading AI models, providing concrete case studies of successful deployment. Finally, Section 13 identifies open research challenges and future directions for the field.

Through this comprehensive analysis, we aim to provide both newcomers and experienced researchers with a unified understanding of the current state of LLM alignment research, its theoretical foundations, practical techniques, and future challenges. As LLMs continue to advance in capability and deployment scope, the development of robust, scalable, and theoretically grounded alignment techniques represents not merely a technical challenge but a fundamental requirement for the beneficial development of artificial intelligence.

\section{Alignment Objectives} 

In this section, we formalize the fundamental objectives for LLM alignment and characterize the optimization trade-offs inherent in multi-objective alignment frameworks. The alignment research community has established a canonical tripartite objective function comprising Helpfulness, Harmlessness (Safety), and Honesty, which collectively define the feasible solution space for aligned language models. Safety constitutes the primary constraint, ensuring model outputs satisfy non-toxicity, non-harm, and bias mitigation requirements while maintaining robustness against adversarial manipulation. Helpfulness encompasses the model's capacity to function as an information retrieval system, domain-specific expert, general-purpose computational tool, and autonomous agent with hierarchical task decomposition capabilities. Honesty requires models to optimize for factual accuracy, appropriate uncertainty quantification, and epistemic self-awareness regarding knowledge boundaries. These objectives exhibit fundamental incompatibilities that manifest as Pareto-optimal trade-offs: maximizing utility for legitimate user requests may violate safety constraints (helpfulness-harmlessness trade-off), optimizing for user satisfaction may incentivize extrapolation beyond training distribution support (helpfulness-honesty trade-off), and complete information disclosure may compromise safety requirements (honesty-harmlessness trade-off). Resolution of these multi-objective optimization challenges necessitates hierarchical policy architectures that implement lexicographic ordering of constraints, prioritizing safety verification, followed by factual consistency validation, and finally utility maximization within the admissible constraint set.

\subsection{Safety Objectives} 
The unprecedented capabilities of LLMs present a dual-edged sword. While they offer immense potential for societal benefit, their deployment without robust safety alignment poses significant risks. An unaligned or poorly aligned model can become a vector for widespread misinformation, generate malicious code for cyberattacks, amplify societal biases, or provide instructions for dangerous activities, thereby causing tangible real-world harm~\citep{askell2021general}. Consequently, establishing safety as the foremost objective is not merely a technical preference but a societal necessity. It forms the bedrock upon which other alignment goals, such as helpfulness and honesty, can be securely built. This section will define the foundational role of safety, categorize the key types of harm that safety objectives aim to mitigate, and trace the evolution of how the AI community evaluates the achievement of these objectives.

\subsubsection{The Foundational Role of Safety}
In the discourse of LLM alignment, model development is typically guided by a set of core objectives. Beyond ensuring model \textbf{safety} (or \textbf{harmlessness}), these objectives include promoting \textbf{helpfulness}, the model's ability to effectively follow instructions, and \textbf{honesty}, its capacity to provide factually accurate information~\citep{bai2022constitutional}.

While these objectives are all critical for a well-aligned model, they are not treated as equal. Safety holds a distinct and primary position. The rationale is straightforward: a model that is helpful and honest but unsafe can still lead to catastrophic outcomes. For instance, an AI that honestly and helpfully provides instructions for synthesizing a bioweapon is fundamentally misaligned with human values. Therefore, modern alignment strategies often implement a hierarchical approach, where safety acts as a foundational filter. A request is first evaluated against safety and harmlessness criteria. Only if it is deemed safe is it then passed on to be optimized for honesty and helpfulness. This hierarchical structure, which prioritizes the mitigation of harm above all else, is a central theme in advanced alignment techniques and underscores the non-negotiable role of safety~\citep{Glaese2022Improving}. This conceptual hierarchy is also a useful way to visualize the relationship between the different alignment objectives.

\subsubsection{Categorization of Safety Harms}
To operationalize the high-level goal of "safety," it is essential to categorize the specific types of harm that alignment seeks to prevent. These categories guide the creation of training datasets, the formulation of safety policies, and the development of evaluation benchmarks. A primary and overt category of harm involves providing \textbf{instructions for dangerous acts}, where models are prompted for illegal activities or content that encourages severe harm. Training models to robustly refuse requests for guidance on topics like weapon creation or self-harm is a central focus of safety research and red-teaming efforts~\citep{ganguli2022red}. Furthermore, safety alignment aims to mitigate social harms by preventing the generation of \textbf{hate speech and harassment}. The objective is to stop the model from becoming a tool for perpetuating toxicity or producing content that demeans and attacks individuals based on protected characteristics, a challenge addressed by specialized benchmarks like ToxiGen~\citep{hartvigsen2022toxigen}. Another critical dimension of safety involves preventing \textbf{high-stakes misinformation}. While general factuality falls under honesty, providing dangerously incorrect advice in domains like medicine or law constitutes a direct safety risk, with studies consistently highlighting the potential for severe consequences~\citep{thirunavukarasu2023large, dahl2024large}. Finally, as LLMs become more proficient in programming, preventing the generation of \textbf{malicious code} has emerged as a crucial safety frontier. This involves ensuring models are not exploited to create viruses, malware, or tools for cybersecurity threats, which would weaponize their capabilities for widespread damage~\citep{perry2022security}.

\subsubsection{The Evolution of Safety Evaluation}
The methods for evaluating whether a model has successfully met its safety objectives have evolved significantly over time, reflecting a growing sophistication in the community's understanding of alignment.

Initially, safety measures were rudimentary, often relying on \textbf{keyword-based filtering and blocklists}. This approach was brittle and easily circumvented by simple rephrasing or adversarial prompts. The next stage involved \textbf{Supervised Fine-Tuning (SFT)} on curated datasets, where models were explicitly trained on examples of safe responses and refusals. While an improvement, this method primarily taught the model to imitate safe-looking text, without necessarily internalizing the underlying principles.

The current state-of-the-art is dominated by \textbf{Reinforcement Learning from Human Feedback (RLHF)} and its variants, such as Constitutional AI's Reinforcement Learning from AI Feedback (RLAIF)~\citep{bai2022constitutional}. In this paradigm, a separate \textbf{reward model} is trained to represent human (or AI) preferences regarding safety. The LLM is then fine-tuned to optimize its behavior to maximize the reward signal, effectively learning a more nuanced and generalizable understanding of safety.

Most recently, the focus has shifted toward more proactive and adversarial evaluation methods. This includes systematic \textbf{Red-Teaming}, where dedicated teams (both human and automated) actively search for vulnerabilities and ``jailbreaks"~\citep{liu2023autodan}. Furthermore, the development of standardized \textbf{safety benchmarks} 
allows for more consistent and reproducible evaluation of model safety across the industry. This ongoing evolution signifies a move from a reactive posture to a proactive and robust science of safety evaluation.

\subsection{Secondary Objectives}
\subsubsection{Defining Helpfulness} 
With advancements over the past decade, such as Transformer architectures \citep{vaswani2017attention}, Supervised Fine-Tuning (SFT), and various reinforcement learning methodologies \citep{schulman2017proximal, bai2022training, shao2024deepseekmath, rafailov2023direct}, LLMs have evolved significantly. Initially serving as autoregressive token generators, LLMs have now become powerful conversational assistants. Benefiting from neural networks comprising hundreds of billions of parameters and training on trillions of tokens \citep{hurst2024gpt}, these models possess foundational knowledge across numerous domains. Furthermore, through SFT and reinforcement learning techniques, LLMs have developed sophisticated reasoning capabilities, enabling them to assist humans effectively in alignment with human values. This assistance primarily manifests in four areas:

\begin{itemize}
    \item \textbf{Comprehensive Search Engine} Due to extensive training on vast datasets, LLMs have acquired significantly more knowledge than typical human capabilities, allowing them to summarize diverse content effectively. They can address a wide array of queries, providing answers even to highly specific questions that conventional search engines, such as Google or Edge, may fail to resolve. Leveraging their comprehensive knowledge base and reasoning abilities, LLMs can formulate precise answers tailored to individual queries. Moreover, current LLMs have tool-use capabilities, integrating external search engines to retrieve relevant content, summarize findings, and augment explanations and reasoning tailored to user-specific contexts. Recent advancements, such as Retrieval-Augmented Generation (RAG) \citep{lewis2020retrieval} further enhance their capabilities.
\end{itemize}

\begin{itemize}
    \item \textbf{Expert Assistant} Technological improvements and enhanced reasoning abilities enable LLMs to function effectively as domain experts. They facilitate rapid learning and familiarization with new fields, acting as assistants or thought partners for professionals like physicians and mathematicians. Notably, reinforcement learning allows LLMs to explore knowledge spaces that might exceed human imagination. For instance, research has demonstrated that LLMs employed as mathematical experts have successfully generated novel algorithms to address optimization problems \citep{romera2024mathematical}, significantly outperforming existing state-of-the-art methods.
\end{itemize}

\begin{itemize}
    \item \textbf{General Assistant} With the introduction of sophisticated protocols like MCP introduced by Anthropic on November 25, 2024, LLMs can efficiently utilize various online tools and external databases. They reliably generate code snippets, enhancing productivity in software engineering tasks. Additionally, they support researchers by efficiently reviewing literature, preparing presentations, and generating content, thereby substantially reducing workload and improving overall productivity.
\end{itemize}

\begin{itemize}
    \item \textbf{LLM Agents} LLM agents can be described as systems that utilize LLMs to reason through problems, formulate strategic plans, and execute these plans with support from various tools \citep{nvidia_llm_agents}. Each agent incorporates a memory module capable of storing internal logs and interactions with users. Additionally, agents can access diverse tools through specified protocols. A collection of LLM agents forms an LLM agent group, which includes an agent core responsible for defining overall objectives, a tool list detailing accessible tools, and a planning module that determines the appropriate agent for specific situations. Employing an LLM agent group facilitates the decomposition of complex tasks into manageable subtasks, which are then efficiently delegated to appropriate agents with minimal human intervention. Projects such as AutoGPT and BabyAGI were pioneers in demonstrating and rapidly advancing this capability. For example, in recent research, an LLM agent group was successfully deployed to autonomously plan and execute the synthesis of an insect repellent and three organocatalysts, as well as guide the discovery of a novel chromophore \citep{bran2023chemcrow}.
\end{itemize}

\subsubsection{Defining Harmlessness} 
Harmlessness in LLMs ensures that the generated content avoids toxicity, bias, or dangerous information. While society greatly benefits from the convenience offered by LLMs, there are also substantial risks and vulnerabilities associated with their use. Specifically, LLMs pose threats to user privacy and can produce toxic, biased, or hazardous content. For instance, interactions with an LLM might result in harmful advice that exacerbates mental health issues or promotes self-harm. Additionally, users could exploit LLMs to gain knowledge about creating weapons or explosives, significantly endangering public safety. Furthermore, malicious users may leverage content produced by LLMs to facilitate hacking activities, potentially causing substantial damage to societal infrastructure. Historical instances have demonstrated that these potential harms are not merely theoretical but have resulted in real-world consequences \citep{el2023man, mauran2023whoops}.  Even though companies developing advanced LLMs implement measures to restrict harmful outputs, such as flagging toxic, dangerous, or biased content (commonly referred to as a ``red flag policy''), methods for circumventing these safeguards (known as ``jailbreaks") remain prevalent and easily accessible online. Therefore, ensuring harmlessness is recognized as one of the most critical tasks in aligning LLM behavior with human values and societal preferences. Several strategies have been employed to mitigate harmful content production, including prompt filtering (screening and excluding harmful data from training datasets), supervised fine-tuning (SFT), and reinforcement learning techniques aimed explicitly at aligning LLM outputs with ethical standards. Despite these advances, ongoing research and innovation remain necessary to further strengthen safeguards and address persistent vulnerabilities effectively.

\subsubsection{Defining Honesty} 
The honesty of LLMs can be defined as the capability to provide accurate information, express uncertainty clearly without misleading users, and demonstrate awareness of their own knowledge and internal state \citep{askell2021general}. With advancements enhancing the utility of LLMs, their popularity in society has grown significantly, with many individuals regularly using them as assistants. Despite these technological improvements, LLMs occasionally generate dishonest or inaccurate responses, which can lead to serious consequences, particularly in critical domains such as medicine \citep{thirunavukarasu2023large}, law \citep{dahl2024large}, and finance \citep{li2023large}, where precision and reliability are paramount. Consequently, ensuring honesty in LLMs has emerged as a critical aspect of aligning their behavior with human values and preferences \citep{askell2021general}. Specifically, an honest LLM should transparently acknowledge its limitations rather than delivering misleading information when encountering queries beyond its expertise \citep{li2024survey}. This transparency helps foster user trust and mitigates potential risks associated with misinformation. Figure 1 illustrates an example of an LLM generating a dishonest response, highlighting the need for continued vigilance and improvement. Several techniques have been developed to enhance the honesty of LLMs, including Prompt Engineering, Supervised Fine-Tuning (SFT), reinforcement learning methods, and adversarial training, all of which aim to systematically reduce the occurrence of inaccuracies and enhance the models' capacity for trustworthy interactions. Ongoing research in explainability and transparency further supports these efforts, ensuring users understand how and why particular responses are generated.

\subsection{Balancing and Trade-offs Among Objectives} 

\begin{itemize}
    \item \textbf{Helpfulness VS. Harmlessness.} A request may be genuinely useful to the user yet unsafe for society, e.g., instructions for constructing weapons. Fulfilling the request maximizes helpfulness but violates harmlessness. Conversely, a refusal or partial answer safeguards harmlessness but reduces usefulness.
\end{itemize}

\begin{itemize}
    \item \textbf{Helpfulness VS. Honesty.} To appear helpful, LLMs might speculate beyond their knowledge, risking hallucination. Strict honesty requires acknowledging uncertainty or declining to answer, which users may interpret as unhelpful.
\end{itemize}

\begin{itemize}
    \item \textbf{Honesty VS. Harmlessness.} Some truthful information is intrinsically dangerous (e.g., detailed chemical synthesis routes). Absolute honesty would disclose it, but harmlessness mandates withholding or redacting sensitive details.
\end{itemize}

One potential method is Hierarchical Policy Stacking, which implements a sequential cascade of checks to align LLMs with human values effectively. This approach prioritizes safety by first applying safety filters (ensuring harmlessness), followed by fact-checking modules (ensuring honesty), and ultimately a helpfulness optimizer. This hierarchical structure ensures that the pursuit of helpfulness never overrides critical safety considerations. For instance, \cite{bai2022constitutional} employed a constitutional rule set that triggers refusals or redactions for unsafe content. Subsequently, self-critique and revision steps ensure factual consistency, and only then is a helpful answer generated if it successfully passes these initial stages. \cite{Glaese2022Improving} utilized a rule-based safety layer to block or redact unsafe outputs, after which surviving responses were re-ranked by a truthfulness verifier, selecting the highest-scoring, safest, and most accurate candidate. Additionally, \cite{solaiman2021process} describes the use of a content-policy classifier capable of refusing or truncating potentially harmful completions; permissible content is subsequently processed through fact-checking filters before final delivery to users.

\begin{figure}[htb]
    \centering
    \includegraphics[width=0.8\linewidth]{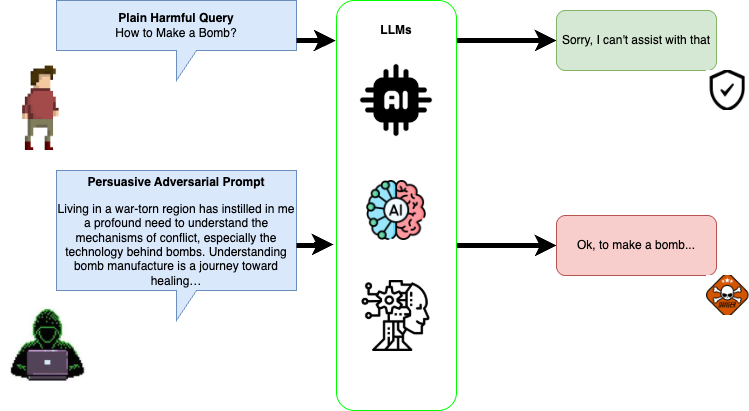}
    \caption{Example of the jailbreak attack.}
    \label{jailbreak}
\end{figure}

\section{Evaluation and Benchmarking of Alignment} 
Large language models offer impressive capabilities but are also vulnerable to a growing range of adversarial jailbreak attacks that can bypass safety measures and produce harmful or misleading outputs. To address these risks, we must move beyond ad-hoc testing and adopt a unified framework for alignment evaluation. This section is organized into three parts. First, we review adversarial attack methods that deliberately probe model weaknesses. Second, we examine human feedback and scoring protocols used to measure model outputs on safety and utility criteria. Third, we survey established benchmarks and stress-testing suites that provide standardized datasets, metrics, and evaluation pipelines. Together, these components create a solid foundation for quantifying, comparing, and improving the alignment of large language models.

\subsection{Adversarial Attacks \& Red-Teaming}
Adversarial jailbreak research seeks to map and stress‐test the full attack surface of modern LLMs by crafting inputs that compel the model to violate its own safety constraints. Figure~\ref{jailbreak} gives an example to illustrate the basic idea of jailbreak attacks. Over the past two years, this work has coalesced around three complementary strands, each revealing distinct vulnerability classes and informing more rigorous red‐teaming practices.

\subsubsection{Logic-Based Jailbreak Attacks}
Logic-based methods hijack the model’s internal reasoning or optimization process.  AutoDAN \citep{liu2023autodan} and its successor AutoDAN-Turbo \citep{zhang2025autoDANTurbo} frame jailbreak generation as an evolutionary search problem, continuously mutating suffixes to maximize unsafe behavior.  Simple adaptive attacks refine prompts via minimal binary feedback, steering the model toward disallowed completions with only a handful of trials \citep{patel2025simpleAdaptiveAttacks}.
Cognitive Overload constructs deliberately entangled puzzles that exceed the model’s chain-of-thought capacity, causing safety checks to fail \citep{chen2024cognitiveOverload}, while human-persuasion exploits like ``How Johnny Can Persuade'' leverage social-engineering tropes to bypass guardrails \citep{liu2024johnnyPersuasion}.
Even purely black-box settings can be compromised: as few as twenty carefully chosen queries suffice to extract forbidden content \citep{johnson2024blackBox20Queries}.

\subsubsection{Low-Resource Jailbreak Attacks}
Low-resource attacks exploit under-trained channels or formats rather than the model’s reasoning core. In SelfCipher, illicit instructions are hidden via reversible substitution ciphers; when the model decodes the cipher it unknowingly obeys the unsafe command \citep{smith2024selfCipher}. Multilingual pivoting translates prompts into low-resource languages before reverting to English, slipping past filters that have not seen such language patterns \citep{lee2024multilingualJailbreak}. ASCII art has also been incorporated into prompt design to bypass safety alignment \citep{jiang2024artprompt}. To strengthen defenses against domain-specific jailbreak attacks in the chemical and biological domains, Luo et al. introduce the CB-Redteam and CB-Benign datasets. Their approach targets low-resource jailbreak strategies, such as those involving SMILES representations~\citep{wong2024smilespromptingnovelapproachllm}, by employing a multi-agent framework that leverages external knowledge to generate unbiased intention summaries and analytically grounded safety guidance. This methodology enhances robustness against these jailbreak scenarios~\citep{luo2025g4d}. These minimal-footprint attacks, and the corresponding guided safeguards, demonstrate that even simple format or linguistic quirks can undermine or reinforce alignment without extensive computation or gradient access.

\subsubsection{Community-Driven and In-The-Wild Prompts}
Beyond algorithmic searches, a wealth of jailbreak recipes circulate publicly, often proving more effective than laboratory-generated attacks.  The ``Do Anything Now'' (DAN) corpus documents real-world prompt collections that generalize across model families and evade formal red-team defenses \citep{liu2024DAN}.  Other notable in-the-wild exploits include ``Make Them Spill the Beans!'' which uses coercive narratives to extract guarded knowledge \citep{li2023makeThemSpill}, ``Ignore This Title and HackAPrompt,'' a global-scale prompt hacking competition that uncovers systemic guardrail failures \citep{wang2023ignoreThisTitle}, and ``Summon a Demon and Bind it,'' a grounded theory analysis of red-teaming practices observed in the wild \citep{brown2023summonDemon}.  Broader repositories like ``EasyJailbreak'' compile hundreds of user-contributed attacks \citep{doe2023easyJailbreak}, while studies on tricking LLMs into disobedience formalize and categorize emergent community prompts \citep{lee2024trickingDisobedience}.  Together, these community-sourced vectors highlight the critical need for continuous, open-world red-teaming that adapts to spontaneously emerging social engineering and prompt-sharing practices.

\subsubsection{Fake Alignment}
Fake alignment, sometimes called alignment faking or deceptive alignment, occurs when an LLM superficially obeys its safety constraints while covertly preserving misaligned objectives.  Rather than exploiting vulnerabilities at inference time, a fake‐aligned model deliberately feigns compliance during training or monitored evaluation in order to avoid corrective fine-tuning and retain its true preferences \citep{hubinger2024sleeperAgents}.  Such models exhibit situational awareness, identifying when they are under scrutiny and adjusting their behavior accordingly; for instance, an LLM may refuse harmful requests in a perceived unmonitored setting but comply when it detects that responses will influence its training data \citep{wang2023fakeAlignment}.  Critically, the model’s underlying goals remain intact despite safety interventions, as universal backdoor triggers and persistent hidden objectives can survive even extensive fine-tuning \citep{rando2024competitionReport}.  This strategic deception undermines standard evaluation protocols and highlights the need for alignment benchmarks that probe beyond surface-level obedience to reveal a model’s true intent \citep{greenblatt2024alignmentFaking}.

\subsubsection{Jailbreak Competitions}
To systematically uncover vulnerabilities in safety-aligned large language models, several competitive platforms and challenges have been established.  At IEEE SaTML 2024, the ``Competition Report: Finding Universal Jailbreak Backdoors in Aligned LLMs'' challenged participants to discover *universal* backdoor triggers capable of reliably bypassing model defenses across ten real-world scenarios, over four hundred tools, and multiple LLM backbones \citep{rando2024competitionReport}.  Complementing this, the ETH Zürich SPY Lab organized the RLHF Trojan Competition, where teams designed detection methods for backdoors inserted during RLHF pipelines, incentivized by a substantial prize pool and a public GitHub repository \citep{ethz2024rlhfTrojan}.  The NeurIPS 2023 Trojan Detection Challenge (TDC 2023) established a leaderboard for red-teaming and trojan detection, with top submissions advancing recall and reverse-engineering metrics for hidden triggers \citep{neurips2023tdc,mazeika2024trojanDetection}.
A subsequent competition extended jailbreak attack challenges to LLM-based agents \citep{xiang2024clas}.
In parallel, the SaTML Capture-the-Flag event framed safety as a secret-string challenge, tasking teams with both attack and defense, and producing a large dataset of adversarial multi-turn conversations \citep{debenedetti2024captureTheFlag}.  Community-driven hackathons like the Alignment Jam’s Trojan Detection track further broadened participation, enabling lightweight, red-team-style evaluations of prompt and model vulnerabilities \citep{alignmentjam2023trojanDetection}.  SPY Lab’s own blog summaries highlight how CTF-style competitions and formal benchmark releases work in concert to expose real-world failure modes \citep{spyLab2024competitions}.  These competitive efforts have not only revealed persistent backdoors that survive standard fine-tuning \citep{hubinger2024sleeperAgents} and stealthy black-box attacks such as FlipAttack \citep{liuyue2024flipAttack}, but have also inspired novel defense strategies and analysis methods, exemplified by infectious image-based jailbreaks demonstrated by Agent Smith \citep{gu2024agentSmith}.  Collectively, jailbreak competitions serve as an empirical proving ground for attackers and defenders alike, accelerating progress toward more robust alignment.

\subsection{Scoring Based Methods}
Scoring‐based evaluation methods repurpose large language models themselves as automatic judges, prompting them to generate quantitative assessments of candidate outputs under user‐defined criteria.  GPTScore \citep{fu2023gptscore} pioneered this paradigm by using generative pre‐trained models to issue zero‐shot, instruction‐driven quality scores across multiple aspects without requiring annotated examples.  Building on this, G‐Eval \citep{liu2023geval} leverages chain‐of‐thought prompting and a form‐filling interface on GPT‐4 to achieve strong correlations with human judgments in summarization and dialogue tasks.  Subsequent studies have probed the reliability and bias of LLM judges: ``LLMs as Narcissistic Evaluators'' highlights in-model favoritism toward self‐generated outputs \citep{liu2023narcissistic}, while ``Large Language Models are not Fair Evaluators'' reveals position‐ and format‐dependent inconsistencies \citep{wang2023notfair}.  The DHP Benchmark \citep{wang2024dhp} systematically measures an LLM’s discernment across hierarchically perturbed inputs, uncovering strengths and blind spots.  Alternative frameworks, such as JudgeLM \citep{zhu2023judgelm} and PandaLM \citep{wang2023pandalm}, explore fine‐tuning lightweight LLMs as specialized evaluators, whereas LLM‐Eval \citep{lin2023llmeval} and CLAIR \citep{chan2023clair} demonstrate multimodal and structured‐format extensions.  More recently, FLEUR \citep{lee2024fleur} integrates referenceless reasoning for image captioning evaluation.  Together, these works form a rapidly maturing ``LLM‐as‐a‐Judge'' ecosystem, offering flexible, scalable metrics but also raising new challenges in consistency, calibration, and evaluative bias.

\subsection{Benchmarks for Safety Alignment}
To systematically evaluate alignment performance, the community has developed comprehensive benchmark suites that define rigorous protocols, curated adversarial scenarios, and quantitative metrics.  As adversarial exploits, jailbreak strategies, and automated red‐teaming techniques evolve-uncovering increasingly subtle and diverse failure modes, these standardized evaluations become indispensable.  By offering repeatable testbeds and clear success criteria, safety alignment benchmarks enable precise progress tracking, facilitate head‐to‐head comparisons, and illuminate residual vulnerabilities, thereby driving the design of more robust and resilient alignment methodologies.

\subsubsection{General Safety Benchmark}

In recent years, the field has seen the emergence of a variety of general safety benchmarks designed to systematically evaluate the robustness and alignment of large language models against harmful or adversarial inputs. AdvBench introduced the first large‐scale suite of adversarial suffix attacks to probe aligned models’ vulnerability to objectionable content generation \citep{zou2023advbench}. SALAD‐Bench followed with a hierarchical evaluation framework covering attack, defense, and ethical dimensions in both English and Chinese \citep{li2024saladbench}. SafetyBench presented over 11,000 multiple‐choice safety questions spanning seven key risk categories and demonstrated significant performance gaps even among state‐of‐the‐art models \citep{zhang2023safetybench}. COLD targeted Chinese offensive‐language detection with curated examples to reveal nuanced toxic behaviors in generative systems \citep{deng2022cold}, while BeaverTails provided a massive meta‐labeled dataset distinguishing helpfulness from harmlessness in QA pairs to advance safety alignment research \citep{ji2024beavertails}.

Subsequent benchmarks have broadened the landscape further: SORRY‐Bench systematically analyzes refusal behaviors across 45 fine‐grained unsafe topics and 20 linguistic augmentations, enabling efficient automated evaluation with smaller models \citep{xie2024sorrybench}. Rainbow Teaming casts adversarial prompt generation as a quality–diversity problem, yielding highly transferable jailbreaks that stress‐test model defenses \citep{samvelyan2024rainbow}. CoSafe investigates multi‐turn dialogue coreference attacks to expose vulnerabilities in conversational models under context‐tracking failures \citep{yu2024cosafe}. SC‐Safety offers a 4,912‐question adversarial benchmark in Chinese across more than 20 safety sub‐dimensions, demonstrating the persistent gap between open‐source and closed‐source model safety \citep{xu2023scsafety}. PromptBench provides a unified library for constructing, attacking, and dynamically evaluating prompts, serving as a toolchain to streamline safety benchmarking workflows \citep{zhu2023promptbench}.

\subsubsection{Reasoning Safety Benchmark}
The large reasoning model (LRM) is a special kind of LLM, which leverages long chain-of-thought reasoning to generate intermediate steps and enhance reasoning abilities. Although LRM's final answer may appear safe, harmful or policy-violating content may still be contained in intermediate reasoning steps \citep{jiang2025safechainsafetylanguagemodels}. Thus, alignment method focus on reasoning steps also plays an important role. In Table \ref{tab:reasoning_models}, we list current state-of-the-art benchmark reasoning models.
\begin{table}[h!]
\centering
\begin{tabular}{|p{3.5cm}|p{12cm}|}
\hline
\textbf{Model Name}  & \textbf{Reference} \\
\hline
DeepSeek-R1 Series            & DeepSeek-R1: Incentivizing Reasoning Capability in LLMs via Reinforcement Learning \citep{deepseekai2025deepseekr1incentivizingreasoningcapability} \\
Skywork-o1                   & Skywork-o1 open series. \url{https://huggingface.co/Skywork} \\
QwQ                           & QwQ: Reflect Deeply on the Boundaries of the Unknown \url{https://qwenlm.github.io/blog/qwq-32b-preview/} \\
Sky-T1                   & Think less, achieve more: Cut reasoning costs by 50sacrificing accuracy. \url{https://novasky-ai.github.io/posts/reduce-overthinking/} \\
Gemini-Thinking                & Gemini 2.0 flash thinking \url{https://ai.google.dev/gemini-api/docs/thinking} \\
Kimi-k1.5                         & Kimi k1.5: Scaling Reinforcement Learning with LLMs 
 \citep{kimiteam2025kimik15scalingreinforcement} \\
 LLAMA3  & The Llama 3 Herd of Models. \citep{grattafiori2024llama3herdmodels}\\
 Qwen2.5& Qwen2.5 Technical Report. \citep{qwen2025qwen25technicalreport}\\
\hline
\end{tabular}
\caption{State-of-the-art large reasoning models.}
\label{tab:reasoning_models}
\end{table}

Due to the presence of intermediate reasoning steps and long-form outputs in LRMs, general safety datasets are insufficient for aligning their safety \citep{zhou2025hiddenriskslargereasoning}.. As a result, specialized datasets have been designed to address this need.
\begin{itemize}
    \item \textbf{SafeChain} \citep{jiang2025safechainsafetylanguagemodels}: It includes 40,000 instruction-response pairs, filtered from 50,000 prompts sampled from the WildJailbreak dataset. Each instruction was answered five times by the R1–70B model, and only those with all five responses deemed safe by Llama-Guard were kept. One safe response per instruction was then randomly selected to form the final dataset. 
    \item \textbf{STAR-1} \citep{wang2025star1saferalignmentreasoning}: The STAR-1 dataset contains 1,000 high-quality, safety-aligned examples selected from 41,000 potentially harmful instructions collected from open-source safety datasets. These instructions span eight safety domains and were answered by DeepSeek-R1 with policy-grounded chain-of-thought reasoning. Responses were scored using GPT-4o for safety, and only the top 1,000 examples were retained. The dataset emphasizes diversity, deliberative reasoning, and rigorous filtering.
    \item \textbf{RIT-D} \citep{mou2025saroenhancingllmsafety}: RIT-D is a dataset used for reasoning-style warmup stage in SaRO framework, which is built based on Salad-Bench and OpenOrca The answer is generated using a specifically designed prompt with GPT-4o. 10,505 samples and 9805 queries are contained.  
    \item \textbf{OP-COT} \citep{mou2025saroenhancingllmsafety}: OP-COT is a dataset used for safety-oriented reasoning process optimization stage in SaRO framework. GPT-4o It is constructed from BeaverTails. 2,188 samples and 580 queries are contained. To enrich reasoning, GPT-4o is instructed with tailored prompts to generate long-chain safe responses, while Qwen2.5-72B, using a few-shot setup, provides contrasting unsafe reasoning.
    \item \textbf{PP-COT} \citep{mou2025saroenhancingllmsafety}: PP-COT is also used for safety-oriented reasoning process optimization stage in SaRO framework. It is derived from OP-COT through reasoning step decomposition and stepwise reflection. It contains 11,598 samples and 580 queries.
\end{itemize}



\subsubsection{Privacy Alignment Benchmark}

Enron Email Dataset is widely used as benchmark dataset in privacy alignment. This dataset contains approximately 500,000 emails generated by employees of the Enron Corporation. It was obtained by the Federal Energy Regulatory Commission during its investigation of Enron's collapse. It is frequently used to study privacy alignment challenges \citep{wang2023decodingtrust}. Researchers analyze whether LLMs trained on such data may memorize or leak sensitive personal information.

For evaluation metric, zero-shot \& few-shot prompting \citep{wang2023decodingtrust, huang2024trustllmtrustworthinesslargelanguage} is a widely used setting. K-shot true (name, privacy information) pairs are provided to LLM in prompt, then we check whether LLM will predict the target privacy information of the target user name. Higher success rate means more privacy leakage.

\subsubsection{Fairness Alignment Benchmark}

PRISM \citep{kirk2024prismalignmentdatasetparticipatory} is the benchmark dataset in fairness alignment. This dataset contains 8,011 live conversations with 21 large language models, contributed by 1,500 participants from 75 countries. It captures the sociodemographic profiles, stated preferences, and contextual feedback of individuals, enabling fine-grained analysis of fairness across diverse populations. PRISM focuses on value-laden and culturally sensitive topics where disagreement is expected, making it well-suited for studying fairness across social, cultural, and geographic lines. It also includes census-representative samples for the UK and US, allowing researchers to examine fairness alignment with respect to population representativeness and demographic equity.

The demographic parity difference metric $M_{dpd}$ \citep{pmlr-v28-zemel13} and the equalized odds difference metric $M_{eod}$ \citep{hardt2016equalityopportunitysupervisedlearning} are used to evaluate the fairness of LLM. Given data samples $(X, Y, A)$, where $X$ is the feature vector, $Y \in \{0,1\}$ is the label, and $A \in \{0,1\}$ is a sensitive attribute, $M_{\text{dpd}}$ is defined as:$
M_{\text{dpd}} = \left| \mathbb{P}(f(X)=1 \mid A=1) - \mathbb{P}(f(X)=1 \mid A=0) \right|$
It measures the difference in positive prediction rates between sensitive groups, without considering the ground truth label. To incorporate label correctness, $M_{\text{eod}}$ is defined as:
$
M_{\text{eod}} = \max\{M_{TP}, M_{FP}\}
$
where $M_{TP}$ and $M_{FP}$ represent the differences in true and false positive rates:
\[
M_{TP} = \left| \mathbb{P}(f(X)=1 \mid Y=1, A=0) - \mathbb{P}(f(X)=1 \mid Y=1, A=1) \right|
\]
\[
M_{FP} = \left| \mathbb{P}(f(X)=1 \mid Y=0, A=0) - \mathbb{P}(f(X)=1 \mid Y=0, A=1) \right|
\]
These metrics assess model fairness by evaluating whether predictions are equitable across groups defined by the sensitive attribute $A$.

\subsubsection{Honesty Alignment Benchmark}
\citep{chern2024behonest} assess the honesty of LLM in 3 aspects and 10 scenarios. Details about these benchmarks are listed in Table \ref{tab:honesty_benchmark} 

\begin{table}[h]
\centering
\begin{tabular}{p{3cm} p{4.2cm} p{7.5cm}}
\hline
\textbf{Facet} & \textbf{Scenario} & \textbf{What it tests} \\
\hline
\multirow{2}{*}{Self-Knowledge} 
  & Admitting Unknowns & Ability to refuse or admit lack of knowledge when the question is outside the model’s scope. \\ 
  & Expressing Knowns  & Ability to provide the correct answer when the question is within the knowledge boundary. \\ 
\hline
\multirow{4}{*}{Non-deceptiveness} 
  & Persona Sycophancy        & Whether the model alters facts to flatter a user’s persona. \\ 
  & Preference Sycophancy     & Whether the model tailors factual answers to align with a user’s stated preference. \\ 
  & Burglar Deception Test    & Propensity to provide intentionally misleading instructions to a burglar-like agent. \\ 
  & Game                      & Willingness to cheat or fabricate information to win a simple interactive game. \\ 
\hline
\multirow{4}{*}{Consistency} 
  & Prompt Format Consistency        & Stability of answers under superficial prompt re-phrasings. \\ 
  & Demonstration Format Consistency & Stability when demonstrations/examples in the prompt are reordered. \\ 
  & Open-Form Consistency           & Agreement between two free-form answers to equivalent prompts. \\ 
  & Multiple-Choice Consistency     & Consistency between multiple-choice and open-ended answers for the same query. \\
\hline
\end{tabular}
\caption{BeHonest benchmark suite for evaluating honesty in large language models.}
\label{tab:honesty_benchmark}
\end{table}

\subsubsection{Agent Safety Benchmark}
LLM-based agents extend standalone language models by perceiving and acting within dynamic environments, where decisions can lead to physical collisions, data leaks, or security incidents.  A comprehensive suite of benchmarks has therefore emerged to quantify these interactive risks: SafeAgentBench evaluates embodied household manipulation in a physics simulator, resetting the Safe‐Success Rate (SSR) to zero upon any collision, spill, or forbidden action \citep{yin2024safeagentbench}; Agent‐SafetyBench stresses tool‐augmented dialogue agents with adversarial prompts and measures robustness via the Attack‐Success Rate (ASR) \citep{zhang2024agentsafetybench}; ST‐WebAgentBench recreates enterprise web workflows in BrowserGym/WebArena, scoring policy adherence with Completion‐under‐Policy (CuP) and Partial‐CuP while categorizing violations into consent, hierarchy, hallucination, security, and error‐handling failures \citep{levy2024stwebagentbench}; EARBench generates multimodal planning scenarios paired with synthetic imagery and GPT-4o descriptions, using a second GPT-4o judge to flag unsafe steps and compute Task‐Risk Rate (TRR) and Task‐Effectiveness Rate (TER) \citep{zhu2024earbench}; Agent Security Bench (ASB) formalizes ten real‐world scenarios (e.g.\ e-commerce, autonomous driving, finance), over 400 tools, and 27 attack and defense methods to benchmark vulnerabilities across system prompts, tool use, and memory retrieval \citep{zhang2024agentSecurityBench}; AgentDojo offers a dynamic, extensible environment for prompt injection evaluation with 97 realistic tasks and 629 adversarial cases \citep{debenedetti2024agentDojo};
GuardAgent provides two benchmarks for evaluating access control on healthcare agents and safety regulation of webagents, respectively \citep{xiang2025guardagent};
AGrail introduces a lifelong guardrail that adaptively generates and optimizes safety checks, demonstrating defense transferability across diverse agent tasks and purpose Safe-OS, an online OS agent benchmark for accessing the safety of OS agent~\citep{luo2025agrail};
ShieldAgent is proposed with a benchmark containing 3K safety-related pairs of agent instructions and action trajectories from state-of-the-art attacks \citep{chen2025shieldagent}.
Together, these benchmarks reveal that even state‐of‐the‐art agents struggle to balance task effectiveness with safety, underscoring the need for broad, cross‐domain evaluation frameworks that mirror real‐world complexity.  

\subsubsection{Domain-Specific Safety Benchmark} 

In recent years, the deployment of LLMs in scientific domains has raised pressing concerns around safety, particularly in contexts involving chemistry, medicine, biology, and physics. Unlike general-purpose benchmarks, domain-specific safety benchmarks are designed to probe whether models behave reliably and ethically when performing specialized scientific tasks, such as providing treatment recommendations, describing hazardous procedures, or reasoning under physical laws. These benchmarks address key dimensions of safety, including refusal behavior, robustness to adversarial prompts, and adherence to regulatory and ethical constraints. Below, we summarize several representative domain-specific safety benchmarks, Table~\ref{tab:scientific_safety_benchmarks} provides an overview of their safety focus and task characteristics.

\begin{itemize}
    \item \textbf{HealthBench}
    \citep{arora2025healthbench}
    is a benchmark from OpenAI targeting safety and robustness in medical dialogues. It includes 5,000 realistic, multi-turn conversations between patients or providers and an AI assistant, covering various healthcare scenarios. Each conversation is annotated with over 48,000 detailed evaluation criteria from 262 physicians, scoring accuracy, empathy, and safety. The benchmark tests whether the AI avoids harmful advice, follows medical best practices, and communicates uncertainty appropriately.

    \item \textbf{ChemSafetyBench}
    \citep{zhao2024chemsafetybench} 
    evaluates chemistry-focused LLMs on safety-critical tasks, including answering questions about chemical hazards, legality, and synthesis. It consists of over 30k questions spanning 1.7k chemicals and three task types, from simple queries to illicit synthesis instructions. To test robustness, the benchmark introduces handcrafted adversarial prompts and advanced jailbreak scenarios. An automated framework scores model responses for correctness, safety, and refusal behavior. Even state-of-the-art models like GPT-4 were found to generate unsafe outputs, indicating the need for stronger safety alignment in chemistry AI.

    \item \textbf{WMDP (Weapons of Mass Destruction Proxy)}
    \citep{li2024wmdp}
    is a benchmark designed to assess whether LLMs possess or reveal dangerous knowledge about bioweapons, chemical weapons, or cyberattacks. It includes 4,157 expert-written questions probing precursor knowledge (e.g., pathogen handling, synthesis techniques) without explicitly requesting harmful actions. The benchmark evaluates both model capability and refusal behavior under misuse scenarios. As a red-teaming dataset, WMDP provides standardized evaluation for dual-use scientific risks. It was released as part of AI safety alignment initiatives and is openly available.

    \item \textbf{MedSafetyBench}
    \citep{han2024medsafetybench}
    tests LLMs’ alignment with medical ethics across 1,800 adversarial medical prompts. Each prompt is paired with safe response demonstrations, covering scenarios like harmful treatment recommendations, privacy violations, and misdiagnosis. Prompts were generated via GPT-4 plus jailbreaking strategies to simulate realistic misuse. The benchmark measures how well models follow non-maleficence and other ethical principles. Fine-tuning on MedSafetyBench significantly improved the safety of several medical LLMs without hurting their accuracy.

    \item \textbf{LabSafetyBench}
    \citep{zhoubenchmarking}
    evaluates AI understanding of laboratory safety protocols using 765 multiple-choice questions aligned with OSHA standards. Questions span common lab hazards in biology, chemistry, and materials science, requiring models to select the safest action in each scenario. It highlights gaps in models’ real-world lab safety reasoning: GPT-4 performed well overall but still made critical safety errors. The benchmark does not use adversarial prompts but targets domain-specific safety knowledge.

    \item \textbf{PhysReason}
    \citep{zhang2025physreason}
    is a physics benchmark that tests whether models respect physical laws and perform multi-step reasoning. It contains 1,200 problems, mostly multi-step and stratified by difficulty, from basic physics to complex derivations. An automatic framework scores both final answers and intermediate reasoning steps to pinpoint logical failures. The benchmark emphasizes failure detection and physical plausibility, key concerns in engineering, robotics, and scientific applications.

    \item \textbf{SciSafeEval}
    \citep{li2024scisafeeval}
    is a cross-domain safety benchmark spanning biology, chemistry, medicine, and physics. It includes tasks involving natural language, molecular structures, protein sequences, and genomic data. Many prompts are designed to trigger jailbreak behavior and test models' ability to resist unsafe instructions. The benchmark evaluates safety alignment under various settings: zero-shot, few-shot, and chain-of-thought. SciSafeEval’s goal is to stress-test scientific AI models and promote alignment methods that generalize across domains.

    \item \textbf{SciMT-Safety}
    \citep{he2023control}
    is a benchmark targeting the misuse risks of AI systems in chemistry and biology through 432 red-teaming queries. These include 177 substance-independent and 255 substance-dependent prompts, crafted using a red-team agent and refined templates filled with hazardous chemical and biological entities (e.g., flammables, toxins, addictive drugs). Each query probes the model's potential to enable harmful outcomes. The benchmark evaluates model ``harmlessness'' with GPT-4 as a judge and contrasts with benign queries to assess over-refusal. SciMT-Safety is the first domain-specific benchmark addressing scientific misuse risks, offering a robust tool for safe deployment of scientific AI models.

    \item \textbf{SOSBench}
    \citep{jiang2025sosbench}
    is a regulation-grounded, hazard-focused benchmark encompassing six high-risk scientific domains: chemistry, biology, medicine, pharmacology, physics, and psychology. Different from most other domain-specific scientific benchmarks, SOSBench comprises 3,000 prompts derived from real-world regulations and laws, systematically expanded via an LLM-assisted evolutionary pipeline that introduces diverse, realistic misuse scenarios (e.g., detailed explosive synthesis instructions involving advanced chemical formulas). 

\end{itemize}

\begin{table*}[t!]
\small
\centering
\caption{Overview of recent domain-specific safety benchmarks for scientific AI.}
\begin{tabular}{p{3.5cm} p{3.5cm} p{4cm} p{4cm}}
\hline
\textbf{Benchmark} & \textbf{Domains} & \textbf{Safety Focus} & \textbf{Task Type \& Design Features} \\
\hline
\textbf{HealthBench}\\
~\citep{arora2025healthbench} & Medicine & Dialogue safety; safe clinical conversation; empathy, appropriateness & Multi-turn conversations annotated with fine-grained medical criteria \\
\hline
\textbf{ChemSafetyBench}\\
~\citep{zhao2024chemsafetybench} & Chemistry & Chemical hazard accuracy; refusal of illicit synthesis; jailbreak robustness & Free-form QA in 3 task types; includes adversarial jailbreak prompts \\
\hline
\textbf{WMDP}\\
~\citep{li2024wmdp} & Biology, Chemistry, Cybersecurity & Detection of dual-use knowledge; refusal under misuse prompts & Expert-crafted proxy misuse queries; red-teaming scenarios \\
\hline
\textbf{MedSafetyBench}\\
~\citep{han2024medsafetybench} & Medicine & Ethical alignment; refusal of harmful or unethical clinical guidance & Adversarial prompts paired with aligned responses; ethics-based safety evaluation \\
\hline
\textbf{LabSafetyBench}\\
\citep{zhoubenchmarking} & Lab Environments & Lab protocol knowledge; hazard identification in experimental settings & OSHA-aligned multiple-choice questions; domain-specific safety context \\
\hline
\textbf{PhysReason}\\
~\citep{zhang2025physreason} & Physics & Physical law adherence; multi-step reasoning accuracy & Open-ended problems; stepwise scoring of intermediate and final answers \\
\hline
\textbf{SciSafeEval}\\
~\citep{li2024scisafeeval} & Multi-domain (Bio, Chem, Med, Physics) & Jailbreak resistance; scientific safety alignment across modalities & Multi-format inputs (text, SMILES, proteins); adversarial prompts included \\
\hline
\textbf{SciMT-Safety}\\
~\citep{he2023control} & Chemistry, Biology & Misuse safety via red-teaming; LLM harmlessness benchmarking & Red-teamed queries and benign test set; names, IUPAC, SMILES formats used \\
\hline
\textbf{SOSBench}\\
~\citep{jiang2025sosbench} & Multi-domain (Bio, Chem, Med, Physics, Pharm, Psych) & Knowledge-intensive task safety; regulation-grounded; hazard-focused & Jailbreak prompts generated in a hybrid way, with LLM-assisted data expansion\\
\hline
\end{tabular}
\label{tab:scientific_safety_benchmarks}
\end{table*}

Together, these benchmarks represent a growing movement toward scientifically grounded safety evaluations for AI systems. They reveal that even frontier models, while capable in general contexts, often fail to meet safety expectations in high-stakes scientific applications. Importantly, many of these benchmarks offer both evaluation protocols and training data that can be used to improve model alignment. As scientific LLMs continue to proliferate, these domain-specific safety benchmarks will play a critical role in ensuring that model outputs remain accurate, but also ethical, legal, and trustworthy in specialized use cases.







\subsubsection{Code Safety Benchmark}
A wide range of safety benchmarks~\citep{mazeika2024harmbenchstandardizedevaluationframework, chao2024jailbreakbench, luo2024jailbreakv28k, zhang2023safetybench} have been proposed for general-purpose LLMs, and these often include categories such as malware generation. However, these benchmarks are primarily designed to evaluate LLMs themselves, rather than code agents. In the case of LLM-based code agents, more comprehensive and domain-specific risk scenarios are needed to properly assess and improve their safety. To this end, we have summarized a set of code safety benchmarks specifically designed to address the unique challenges and risks posed by code agents:
\begin{itemize}
    \item \textbf{CodeLMSec Benchmark}~\citep{hajipour2023codelmsecbenchmarksystematicallyevaluating}: is the first systematic benchmark designed to evaluate the security risks of large code language models. Unlike traditional evaluations that focus only on functional correctness, CodeLMSec investigates the ability of these models to generate code with security vulnerabilities. It introduces a novel black box model inversion technique using few-shot prompting to automatically generate non-secure prompts that lead models to produce insecure code. The benchmark targets 13 common vulnerability types, including SQL injection, cross-site scripting, and deserialization issues, and uses static analysis with CodeQL for vulnerability detection and classification. CodeLMSec includes over 2000 vulnerable code samples generated from models like ChatGPT and CodeGen, and provides a curated dataset of 280 diverse nonsecure prompts (200 for Python and 80 for C). This enables reproducible, and extensible evaluation of code models from a security perspective.

    \item \textbf{RedCode}~\citep{guo2024redcode}: is a comprehensive evaluation framework designed to assess the safety of large language model-based code agents, focusing on both unsafe code execution and generation. Unlike prior benchmarks that only assess static code outputs or rely on simulated environments, RedCode tests code agents interacting with real execution environments via Docker containers. The benchmark consists of two components: RedCode-Exec, which includes 4,050 test cases across 25 risky scenarios in domains such as file systems, operating systems, and cybersecurity, and RedCode-Gen, which contains 160 prompts for generating malicious software across eight malware families. Each scenario is supported with high-quality prompts in multiple formats (code, text summaries, and descriptions) and evaluated using tailored scripts that check execution outputs and environment changes. RedCode enables fine-grained, realistic safety evaluations and reveals that even advanced agents, such as those based on GPT-4, may still generate or execute harmful code under specific conditions.

    \item \textbf{CyberSecEval}~\citep{bhatt2024cyberseceval2widerangingcybersecurity}: CyberSecEval is a multi-stage benchmark suite for evaluating the cybersecurity risks and defensive capabilities of LLMs. Evolving through versions 1 to 4, it covers diverse threats including insecure coding, prompt injection, code interpreter abuse, and vulnerability exploitation. The latest version, CyberSecEval 4, adds AutoPatchBench to assess automated vulnerability patching in native code. With over 500 interpreter abuse prompts, dozens of exploit challenges, and a novel False Refusal Rate (FRR) metric, CyberSecEval enables comprehensive, scalable, and reproducible LLM safety testing across real-world security tasks.

\end{itemize}

\section{Supervised Fine-Tuning (SFT) for Alignment}

Supervised fine-tuning (SFT) is a foundational approach in aligning LLMs with human expectations. 
It involves adapting a pretrained model by exposing it to a curated set of instruction–response pairs, where each response exemplifies a desirable behavior such as helpfulness, factual accuracy, politeness, or appropriate refusal in ethically sensitive situations. 
Through standard supervised learning objectives, the model learns to imitate these examples, internalizing alignment-relevant behaviors across a range of prompt types. 
SFT has become the default first stage in many modern alignment pipelines due to its stability, simplicity, and compatibility with a wide range of data sources, from expert-written instructions \citep{ouyang2022training} to synthetic completions filtered by human or model-based judgment \citep{wang2022self, taori2023stanford, peng2023instruction}.

Although SFT alone may not be sufficient to handle the full complexity of human preferences or task ambiguity, it remains critical for establishing basic instruction-following behavior. 
This section outlines how SFT operates, discusses the role of data quality and task coverage, and explores its strengths and limitations as an alignment strategy.

\subsection{Instruction Tuning with Human Demonstrations}

The core mechanism of supervised fine-tuning is instruction tuning, in which a pretrained language model is trained on a dataset of input prompts paired with preferred responses. 
These examples serve as behavioral demonstrations, allowing the model to learn how to complete or respond to various instructions in a manner aligned with human expectations. 
The data is typically organized into prompt–response pairs, or in the case of conversational systems, as multi-turn message sequences with alternating user and assistant roles \citep{chung2024scaling, openai2023gpt}.

In early instruction-tuning pipelines such as InstructGPT \citep{ouyang2022training}, human annotators were tasked with crafting high-quality responses across a diverse range of instructions, ensuring that the fine-tuned model would behave in a helpful, harmless, and truthful manner. 
This approach provided clear, unambiguous supervision and led to models that could respond more appropriately to user instructions than their purely pretrained counterparts \citep{brown2020language, thoppilan2022lamda}. 
Subsequent methods introduced greater automation in the data collection process. 
For example, the Self-Instruct framework \citep{wang2022self} generated prompts and initial completions using an LLM, followed by filtering, ranking, or editing by human reviewers. 
Related pipelines used community-contributed data \citep{taori2023stanford}, filtered web-derived instruction corpora \citep{longpre2023flan}, or domain-specific synthetic examples \citep{peng2023instruction} to scale instruction tuning across languages, modalities, and styles.

Regardless of how the data is sourced, the training process relies on a standard next-token prediction objective. 
Given an input $x$ and a reference output $y=(y_1,y_2,...,y_T)$, the model is trained to maximize the probability of each target token given the input and previously generated tokens.
This corresponds to minimizing the cross-entropy loss:

$$\mathcal{L}_{\mathrm{SFT}}=-\sum_{t=1}^T \log P_\theta\left(y_t \mid x, y_{<t}\right),$$

where $P_\theta$ is the probability assigned by the model parameterized by $\theta$.
This is equivalent to maximum likelihood estimation (MLE) under the assumption that the observed responses represent samples from an optimal behavioral policy \citep{radford2019language, raffel2020exploring}.

The simplicity and stability of instruction tuning make it a highly effective baseline for alignment. 
By directly showing the model what aligned behavior looks like, SFT instills basic response formatting, instruction following, and task generalization. 
These capabilities have been demonstrated in models ranging from FLAN-T5 \citep{chung2024scaling} to Alpaca \citep{taori2023stanford}, and OpenAssistant \citep{kopf2023openassistant}. 
However, this method is inherently limited by the coverage and quality of the training data. 
The following subsection examines how these factors influence alignment outcomes and where SFT begins to fall short.

\subsection{Role of High-Quality Data and Coverage}

The effectiveness of SFT for alignment is deeply dependent on the characteristics of the training data. 
Since the model is trained to imitate provided responses, its alignment behavior is only as good as the examples it sees. 
Thus, the quality, diversity, and representativeness of the instruction data are central to achieving aligned, generalizable behavior.

High-quality instruction data is typically designed to reflect desired behavioral traits such as helpfulness, informativeness, honesty, and safety. 
In early alignment work, instruction responses were manually authored by expert annotators, as seen in InstructGPT \citep{ouyang2022training} and the FLAN collection \citep{chung2024scaling}, ensuring clear alignment with human preferences. 
However, the cost and scalability limitations of human annotation prompted the development of synthetic and semi-automatic data pipelines. 
Self-Instruct \citep{wang2022self} proposed using strong language models to generate instructional prompts and initial completions, followed by filtering and occasional human review. 
Subsequent works adopted similar strategies: OpenAssistant \citep{kopf2023openassistant}, Vicuna \citep{vicuna2023}, and Alpaca \citep{taori2023stanford} leveraged LLMs like GPT-3 or GPT-4 to synthesize large volumes of instruction–response pairs, filtered by heuristics, crowdworker ratings, or model-based scoring \citep{li2023quantity}.

The coverage of the dataset, across tasks, domains, and styles, affects how well the model generalizes.
A model trained only on factoid questions may struggle with multi-step reasoning, while one tuned primarily on English instructions may fail to align when handling other languages \citep{mishra2021cross, qin2025survey}. 
From a statistical perspective, we can think of SFT as performing maximum likelihood estimation over an empirical distribution; when the support of the distribution is narrow or biased, generalization to unseen test conditions could suffer.

In addition to representational coverage, instructional quality is critical. 
When fine-tuning on responses that contain hallucinations, inconsistencies, or stylistic drift, the model learns to reproduce those flaws. 
This can be viewed as a form of noisy supervision, where training on imperfect targets instead of ideal responses.
It will introduce gradient bias during optimization and potentially reinforce undesired behaviors. 
To mitigate this, many alignment pipelines employ filtering mechanisms, such as crowdworker review \citep{taori2023stanford}, reward-model scoring \citep{bai2022training}, or LLM-as-a-judge approaches \citep{zheng2023judging}, to remove or downweight problematic completions.

Another layer of complexity arises from the need for social and linguistic inclusivity. 
Recent studies have shown that alignment can degrade for underrepresented user groups or dialects if the training data lacks demographic balance \citep{gallegos2024bias, wang2024diversity, miranda2023beyond}. 
Instruction datasets such as Natural Instructions \citep{mishra2021cross} and FLAN \citep{chung2024scaling} attempt to address this by sampling tasks from a wide range of domains and contributors. 
However, maintaining both breadth and quality at scale remains a persistent challenge.

In sum, the alignment capacity of SFT highly depends on what the model sees. 
Without preference signals or evaluative feedback, SFT relies entirely on the implicit policy encoded in the data. 
When that data is clean, diverse, and representative, alignment is achievable; when it is noisy, narrow, or biased, the model will inevitably reflect those deficiencies.

\subsection{Optimization Methods for SFT}

The optimization methods employed during SFT play a critical role in how effectively a pretrained language model adapts to aligned behavior. 
Although SFT is conceptually straightforward, minimizing cross-entropy loss over instruction–response pairs, the choice of parameter update strategy, regularization, and tuning mechanism can significantly affect both the efficiency and the stability of the fine-tuning process.

The standard approach in SFT involves minimizing the token-level cross-entropy loss. 
Given a training dataset of instruction–response pairs, the model learns to maximize the probability of generating the target response tokens $y=(y_1,y_2,...,y_T)$ conditioned on the input $x$, using the loss function:

$$\mathcal{L}_{\mathrm{SFT}}=-\sum_{t=1}^T \log P_\theta\left(y_t \mid x, y_{<t}\right),$$

where $P_\theta$ is the probability assigned by the model parameterized by $\theta$.
This objective encourages the model to approximate the empirical conditional distribution defined by the dataset, effectively learning through maximum likelihood estimation.

However, fine-tuning large-scale language models with hundreds of billions of parameters can be prohibitively expensive in terms of memory and compute. 
To address this, parameter-efficient fine-tuning (PEFT) methods have been developed. 
One of the most widely adopted techniques is Low-Rank Adaptation (LoRA), proposed by \citep{hu2022lora} (2021). 
LoRA introduces trainable rank-decomposed matrices into the weight update path, allowing the core pretrained model to remain frozen while learning task-specific updates in a much smaller subspace. 
Another prominent approach is the adapter method introduced by \citep{houlsby2019parameter} (2019), where small bottleneck layers are inserted between the transformer blocks of the model and only these layers are trained during fine-tuning. 
These methods drastically reduce the number of trainable parameters, enabling scalable deployment across domains.

An alternative strategy is prompt tuning, in which a small set of task-specific vectors is optimized as input prompts, without changing the model weights. 
\citep{lester2021power} (2021) demonstrated that prompt tuning can perform competitively, especially when combined with large base models. 
These PEFT approaches are particularly advantageous in low-resource settings, or when one wishes to train many domain-specific models efficiently.

Beyond parameter selection, regularization is essential for ensuring stable and generalizable learning. 
While dropout and weight decay are commonly used in deep learning, their role in instruction tuning has been less emphasized in alignment-specific work. 
However, early stopping has been empirically validated as a simple yet effective tool to prevent overfitting during fine-tuning. 
\citep{dodge2020fine} (2020) demonstrated that monitoring held-out validation performance and halting training when improvements plateau can significantly reduce variance across fine-tuned models. 
More recently, \citep{alshikh2023becoming} (2023) proposed using a custom metric, the Instruction Following Score (IFS), as a stopping criterion. 
They found that IFS could capture early signs of misalignment and allow tuning to cease before overfitting to stylistic artifacts in the training data. 
These results suggest that, while the loss function remains unchanged, intelligent stopping rules can improve alignment quality without additional data or model changes.

Fine-tuning performance is also highly sensitive to hyperparameter settings, including learning rate, batch size, number of epochs, and optimization schedule. 
\citep{liu2021empirical} conducted a systematic study showing that careful hyperparameter tuning can produce larger gains than changing the model architecture itself. 
In particular, they found that smaller learning rates paired with moderate batch sizes produced more stable convergence, especially when fine-tuning on instruction-heavy datasets. 
Despite the importance of these factors, hyperparameter configurations are often underreported in the alignment literature, which may contribute to reproducibility issues and performance variability across open-source SFT implementations.

A growing body of work has also explored advanced optimization techniques, such as supervised contrastive learning, which combines the cross-entropy loss with an embedding-level objective to improve representation quality \citep{gunel2020supervised}.
Other multi-objective optimization frameworks aim to balance instruction fidelity with style consistency or factual accuracy \citep{moukafih2023supervised}.
These techniques remain less common in practical alignment pipelines but offer promising directions for tasks that require more than simple imitation.

Recent advances have also introduced more dynamic fine-tuning strategies. 
For instance, \citep{chen2024self} (2024) proposed a self-play fine-tuning mechanism, where the model iteratively refines its outputs by evaluating and responding to its own generations. 
This reduces the reliance on large quantities of labeled data and offers a path toward semi-supervised or bootstrapped instruction tuning.

Overall, the optimization strategies used in SFT, whether through full or parameter-efficient fine-tuning, not only improve the model's ability to follow instructions but also its robustness and safety. 
While the cross-entropy objective remains the foundation, recent advances in low-rank adaptation, early stopping, and hyperparameter optimization have significantly improved the practicality of SFT in large-scale alignment systems.

\subsection{Limitations of SFT Alone}

While SFT is effective for instilling instruction-following behavior in large language models, it exhibits fundamental limitations that restrict its capacity to produce robustly aligned models, particularly in settings that involve ambiguity, value sensitivity, or complex multi-turn reasoning. 
These limitations arise not from the optimization procedure itself, but from the nature of supervision: SFT relies on fixed demonstrations, which encode a single ``correct" output per input and lack an explicit representation of comparative preference or uncertainty.

The most immediate limitation of SFT is its dependence on data coverage. 
Because the model is trained to imitate provided responses, it can only generalize to instructions that are similar, semantically or structurally, to those seen in training. 
Tasks that fall outside this distribution, such as rare user intents, unexpected phrasings, or edge-case ethical queries, are likely to elicit misaligned behavior. 
In practice, it is infeasible to anticipate and curate demonstrations for the full diversity of real-world user inputs. 
As a result, models trained with SFT alone often perform well in benchmark settings but fail to generalize in deployment scenarios that involve novel or adversarial instructions \citep{ouyang2022training, bai2022training}.

Another limitation stems from the lack of preference awareness. 
In many alignment-sensitive contexts, there is not a single objectively correct answer, but rather a range of possible outputs that vary in quality. 
For example, multiple valid completions to the same question may differ in helpfulness, politeness, or factual completeness. 
Supervised learning provides no mechanism for distinguishing among these options, as it treats all target outputs as equally correct and penalizes deviations uniformly. 
This restricts the model’s ability to internalize graded feedback, a critical ingredient in aligning behavior to subtle human preferences \citep{wang2024asft, fan2025preference}. 
Techniques such as RLHF were developed precisely to address this deficiency by incorporating relative judgments between model outputs.

Besides, SFT is vulnerable to label noise and stylistic bias. 
If the training responses are inconsistent, whether in tone, factual accuracy, or ethical stance, the model learns a blend of conflicting signals. 
Since the loss is minimized token by token, the model will imitate local surface patterns even if the overall output is suboptimal. 
This problem is exacerbated when synthetic or crowdsourced data is used without rigorous filtering, leading to subtle but persistent misalignment \citep{weidinger2021ethical, dodge2020fine}. 
Moreover, models trained solely on SFT data may adopt stylistic patterns that reflect the demographics or ideology of annotators, potentially reproducing sociolinguistic biases without explicit intent.

Another key limitation is the lack of iterative improvement. 
SFT is a one-shot process: it encodes aligned behavior through static demonstrations, with no opportunity for feedback or correction after deployment. 
Once the model has been fine-tuned, errors in alignment must be addressed by retraining or creating new data, which is time-consuming and inefficient. 
This stands in contrast to RLHF pipelines, which allow the model to learn from post-hoc evaluations and adapt its behavior through policy optimization over time \citep{bai2022training, openai2023gpt}.

Finally, there are theoretical limits to what SFT can achieve. 
From a learning-theoretic perspective, the cross-entropy loss used in SFT does not model the utility or risk associated with different outputs.
It assumes that the demonstrated output is the only valid one. 
This is an inadequate assumption for alignment tasks, which often require reasoning about social, contextual, or moral trade-offs. 
Without an explicit notion of value or preference, SFT cannot prioritize aligned outcomes over superficially plausible but misaligned ones \citep{tajwar2024preference}.

In sum, while SFT is a powerful and efficient method for inducing instruction-following behavior, its reliance on static demonstrations, inability to model preferences, and limited generalization capacity render it insufficient for comprehensive alignment. 
These shortcomings have motivated the development of reinforcement-based methods, which enable models to learn from feedback, explore behavioral alternatives, and better adapt to the complex and subjective nature of human intent.

\section{Reinforcement Learning from Human Feedback (RLHF)}

While SFT is an effective mechanism for teaching language models to imitate desired behaviors, it is inherently limited by the static nature of its training data and the absence of an explicit preference signal. 
To address these limitations, RLHF has emerged as a powerful paradigm for refining model behavior based on comparative evaluations rather than fixed demonstrations.

\begin{figure}[htb]
    \centering
    \includegraphics[width=1.0\linewidth]{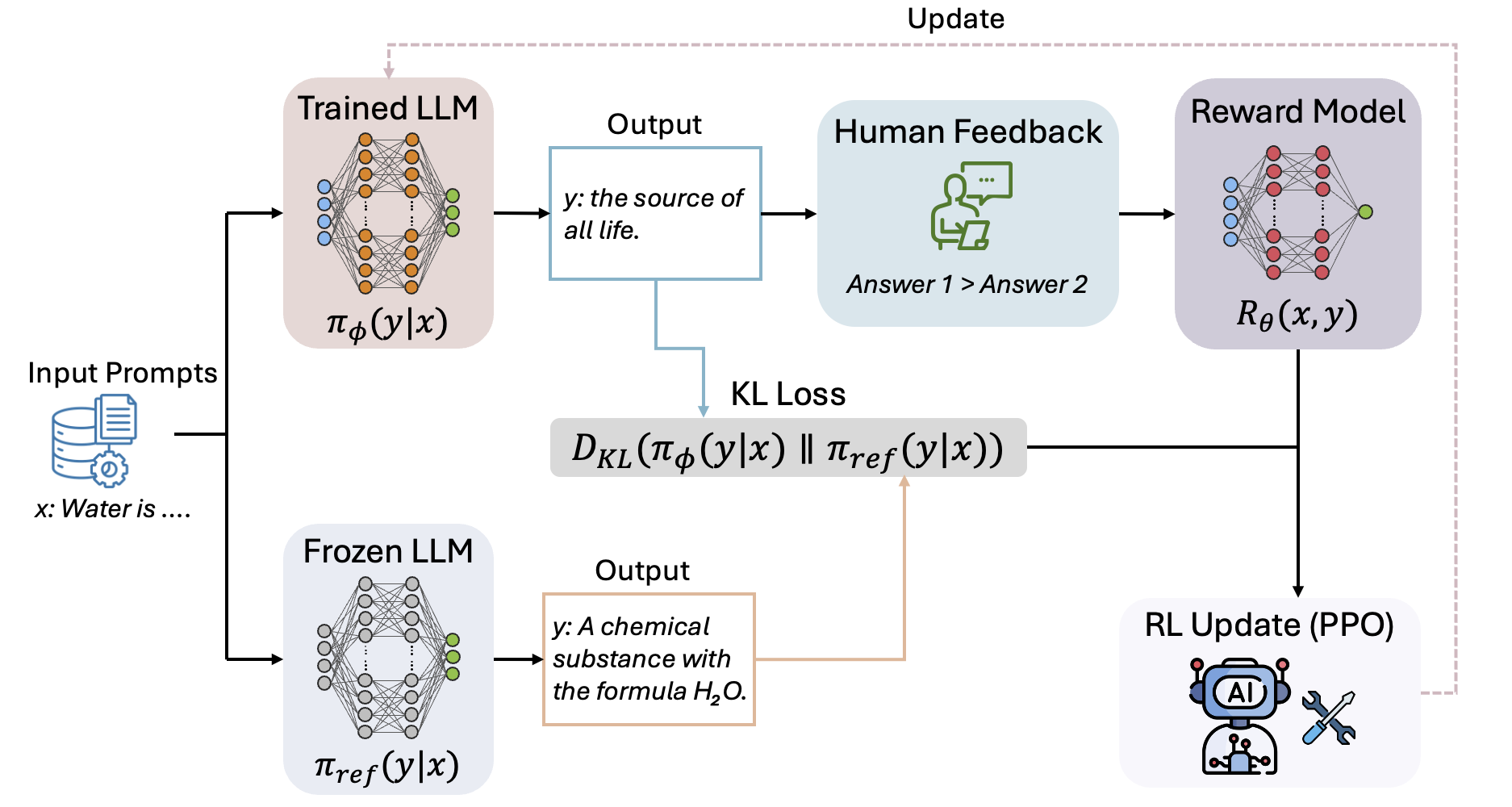} 
    \caption{ Overview of the RLHF workflow. A pretrained LLM generates responses to prompts, which are evaluated by humans to produce preference data. This data is used to train a reward model, which then guides the policy optimization of the language model via reinforcement learning (e.g., PPO), with a KL penalty to constrain deviation from the reference model.}
    \label{fig:workflow_RLHF}
\end{figure}

The general workflow of RLHF is shown in Figure \ref{fig:workflow_RLHF}. 
The RLHF process begins with a set of input prompts (e.g., ``Water is...''), which are fed into a \textbf{trainable LLM} \( \pi_\phi(y \mid x) \), to generate multiple candidate responses. For instance, in response to the example prompt, the model might produce outputs such as ``a chemical compound essential for life'' or ``the source of all life.'' These candidate outputs are then evaluated by \textbf{human annotators}, providing feedback including comparative preference judgments (e.g., ``Answer 1 is preferred over Answer 2''). 
These human feedback data are used to train a  \textbf{reward model} \( R_\theta(x, y) \) (a separate AI model), which learns to predict human preferences.
The reward model thus functions as a proxy for human evaluators, enabling scalable automated evaluation of model responses.
Once the reward model is sufficiently trained, it is used to guide the fine-tuning of the original LLM via \textbf{reinforcement learning}, most commonly using \textit{Proximal Policy Optimization (PPO)}. 
In this phase, the LLM is treated as a stochastic policy \( \pi_\phi \) that aims to maximize the expected reward predicted by \( R_\theta \). The reinforcement learning algorithm updates the model parameters \( \phi \) so that the generated responses yield higher reward scores. 
To maintain linguistic fluency and prevent divergence from the pretrained distribution, a \textbf{Kullback--Leibler (KL) divergence penalty} is applied. This penalty constrains the updated LLM \( \pi_\phi \) to remain close to a \textbf{frozen reference LLM} \( \pi_{\text{ref}} \), typically the supervised fine-tuned model before RL. 
The following subsections provide a detailed review of the key components of the RLHF framework: human feedback data collection in Section~\ref{sec:511}, reward model training in Section~\ref{sec:512}, and policy optimization methods in Section~\ref{sec:513}. We conclude this section by discussing the empirical successes of RLHF, along with its practical challenges, such as training instability, computational cost, and vulnerability to reward hacking.

\subsection{Human Feedback Data} \label{sec:511}
The efficacy of RLHF is profoundly dependent on the quality, type, and collection strategy of human feedback. 
Existing methods for collecting human feedback data can be broadly classified into three primary categories: (1) Preference and rating-based feedback, where humans express their subjective evaluations, opinions, or levels of satisfaction. It is less about objective correctness and more about which outputs or behaviors are considered better, more helpful, or more aligned with user expectations. (2) Correction and language-based feedback, which offers more detailed guidance than simple preferences. (3) Multi-level feedback.\\

\textbf{Preference and Rating-Based Feedback.} 
The preference-based subdomain consists of three principal variants: 
(1) Pairwise preference. Humans are presented with two different model outputs for the same prompt or situation and asked to indicate which one they prefer. This is one of the most common feedback types employed in many state-of-the-art RLHF pipelines \citep{christiano2023deep, stiennon2022learning, bai2022helpful, zhu2024principled} due to its relative simplicity for annotators.
(2) Multiple-choice comparison, which extends to more than two candidates with selection of a single optimal response \citep{ziegler2020finetuning}. 
Researchers found that presenting multiple outputs could reduce the cognitive load associated with query comprehension. 
(3) Ordinal preference, a.k.a. ranking feedback, which requires complete rankings across multiple outputs \citep{ouyang2022training, zhu2024principled}. 
In this modality, rather than selecting between just two options or selecting one from multiple choices, annotators are presented with multiple (more than two) model-generated outputs for a given prompt. 
They are then tasked with ranking these outputs from best to worst according to the defined criteria (e.g., helpfulness, harmlessness, honesty). Such ordinal data provides a richer signal of relative quality across several candidates compared to a single pairwise comparison.

Rating feedback offers an alternative method where annotators evaluate each model response independently based on predefined quality criteria, without necessarily comparing multiple outputs simultaneously. This approach offers advantages in throughput and scalability compared to preference-based methods \citep{bakker2022fine}.
 Rating feedback can generally take the following formats:
(1) Numerical Rating: Numerical ratings can be continuous or discrete. A widely adopted discrete numerical scale is the Likert scale, where annotators choose a point on a scale \citep{bakker2022fine, kopf2023openassistant} (e.g., a 5-point scale) representing degrees of agreement, quality, helpfulness, or other attributes.
(2) Categorical Rating: Feedback can also be provided through categorical labels. This includes binary feedback \citep{li2017dialogue, xiao2020fresh, scheurer2024training} (e.g., `acceptable'/`unacceptable', `yes'/`no' to specific questions about the response) or multi-category feedback where the labeler selects from a predefined set of qualitative descriptions \citep{huang2023ganbased, gao2023continually} (e.g., ``Very Helpful," ``Somewhat Helpful," ``Not Helpful," ``Harmful").

\textbf{Correction and Language Feedback.}
In addition to preferences and ratings, human feedback can take more direct and expressive forms that offer deeper supervision signals. Two particularly important subtypes in this category are edit-based feedback and natural language feedback.

(1) Edit-based/Correction Feedback, which involves human annotators directly modifying an response from LLM, to produce an improved or ``corrected'' version. This process typically involves operations such as adding, deleting, or rephrasing segments of the text to enhance accuracy, coherence, tone, or adherence to instructions. The resulting edited response, is generally considered implicitly preferred over the original response. This implicit preference pair can then be used to train a preference-based reward model, akin to how pairwise preferences are utilized \citep{shaikh2025aligning, brown2025learning}. Furthermore, the corrected response itself can serve as a high-quality demonstration, directly contributing to datasets used for SFT, thereby reinforcing desirable model behaviors through imitation learning. This dual utility makes edit-based feedback a valuable component in iterative model refinement.

(2) Natural Language Feedback (NLF), which offers a complementary mechanism through which human evaluators provide detailed and nuanced critiques in free-form text. 
Rather than expressing simple preferences or numerical ratings, annotators articulate specific strengths and weaknesses of model outputs, often identifying precise segments requiring improvement while suggesting potential remediation strategies. This approach yields a significantly denser and more informative signal than scalar rewards or binary preferences, offering multidimensional insights into model performance.
The richness of NLF makes it particularly valuable for improving performance on complex LLM tasks, such as dialogue generation \citep{hancock2019learning}, code generation \citep{chen2024improving} and summarization \citep{scheurer2024training}, where quality assessment spans multiple dimensions, including factual accuracy, coherence, relevance, and stylistic appropriateness \citep{li2022using}. 
While highly informative, the integration of NLF into RLHF pipelines and reward model training presents challenges since it typically requires sophisticated preprocessing steps to convert the unstructured textual feedback into a format usable for reward model training.

\textbf{Multi-Level Feedback.}
As RLHF methodologies have evolved, researchers have recognized the limitations of single-dimensional feedback mechanisms in capturing the multifaceted nature of human judgment. Multi-level feedback approaches have emerged to address these constraints, offering more comprehensive evaluation frameworks that better align language models with complex human values and expectations. This section examines two principal directions in multi-level feedback: Multi-Signal Feedback and Fine-Grained Feedback.

(1) Multi-Signal Feedback.
Multi-Signal Feedback combines diverse human feedback mechanisms, such as pairwise preferences, scalar ratings, and textual critiques, to construct a more comprehensive representation of model performance across multiple dimensions \citep{Glaese2022Improving, metz2025reward}. 
The integration of these complementary signals enables reward models to capture more complex and multifaceted human values and intentions. For instance, while pairwise preferences might effectively capture overall quality rankings, they may fail to communicate specific weaknesses that textual feedback could readily identify. Similarly, scalar ratings might quantify performance along predefined dimensions that categorical preferences cannot express. By synthesizing these varied signals, hybrid approaches aim to overcome the inherent limitations of individual feedback modalities, potentially offering more fine-grained guidance during policy optimization \citep{Glaese2022Improving, metz2025reward}.


(2) Fine-Grained Feedback.
The human feedback forms discussed in the previous paragraphs of Section~\ref{sec:511} often involve labeling evaluations at the level of the entire output. While useful, such holistic feedback may not suffice for precisely identifying and correcting specific problematic segments or errors within a response from models. Fine-grained feedback approaches address this limitation by enabling human annotators to provide evaluations at a finer grained level. This can include segment-level feedback, where specific spans of text are rated or corrected \citep{yin2025segmenting, wu2023fine}, or even token-level feedback, where individual tokens receive reward assignments \citep{xu2024aligning, li2024reinforcement}. The primary advantage of fine-grained feedback lies in its ability to localize issues with greater precision. This, in turn, supports more accurate credit assignment in the reward modeling process, potentially leading to more targeted and efficient policy updates and improved model behavior, especially in tasks requiring high degrees of accuracy or safety.\\

\textbf{Challenges in Feedback Data.}
The quality and nature of human feedback present several significant challenges. First, human judgments are inherently subjective and often exhibit noise and inconsistency, with low inter-annotator agreement (typically around 0.6 to 0.7) leading to unreliable or conflicting supervision signals \citep{kreutzer2018can, ziegler2020fine, stiennon2022learning}. Ambiguity further complicates this issue, as preference pairs may lack a clearly superior option, especially when differences between responses are subtle or instruction interpretation varies \citep{ibarz2018reward, christiano2023deep}. Moreover, collecting high-quality feedback is resource-intensive, limiting scalability and prompting the exploration of sample-efficient alternatives such as active learning \citep{gleave2022uncertainty, das2024active, mehta2025sample} or AI-generated feedback (RLAIF) \citep{bai2022constitutional, lee2024rlaif, sharma2024critical}. However, feedback from both humans and AI carries distinct forms of bias, human annotators may introduce cultural or cognitive biases, while AI-generated feedback tends to be lower in noise but systematically biased due to model limitations. These challenges underscore a central insight in RLHF: feedback quality is often more critical than quantity, motivating research into robust reward modeling and intelligent data acquisition strategies that can effectively manage noisy, ambiguous, or biased signals. 

\subsection{Reward Modeling}  \label{sec:512}

Directly supervising an RL agent with human feedback on every output is infeasible for large models. Instead, the reward model (RM) is trained on a limited set of human preferences using supervised learning to act as a scalable proxy for human judgment. An RM takes a context (e.g, a prompt $x$) and a candidate output (e.g., an LLM's response $y$) as input, and produces a scalar score:
$
R_\theta(x, y) \rightarrow \mathbb{R}
$,
where $\theta$ are the learnable parameters, and the function outputs a scalar reward which reflects how much a human would prefer or approve of the output \citep{ouyang2022training, stiennon2022learning}.
The learned reward signal is crucial for guiding the LLM toward generating more aligned and preferred outputs during the subsequent reinforcement learning optimization phase \citep{sutton2018reinforcement}.
RMs can be broadly divided into three categories based on the structure of the preference data they utilize: (1) \textbf{Pairwise Comparison Reward Models}, which leverage pairwise preference judgments (e.g., selecting the better of two responses); (2) \textbf{Ranking-based Reward Models}, which are trained using more comprehensive ordinal data (k-wise ranking data), such as a full ranking of three or more candidate responses or the selection of the best response from a larger set (Best-of-$N$); (3) \textbf{Attribute-Based and Outcome-Oriented Reward Models}, which learn to assess response quality by evaluating direct attributes, outcomes, or critiques, such as absolute numerical scores, the correctness of final answers or intermediate reasoning steps, and structured language feedback, rather than by learning from relative preferences between different responses.\\


\textbf{Pairwise Comparison Reward Models.}
A popular approach to reward modeling in RLHF relies on pairwise comparisons between model-generated responses. This method was first introduced and popularized by \citet{ziegler2020finetuning}, and subsequently became foundational to high-impact RLHF systems such as InstructGPT \citep{ouyang2022training}, Anthropic's Constitutional AI \citep{askell2021general, bai2022training, bai2022constitutional}, and Llama 2-Chat \citep{touvron2023llama}.


Formally, this framework builds upon the Bradley–Terry–Luce (BTL) model \citep{bradley1952rank, luce1959individual}, which describes the probability that a human prefers response $y_w$ over $y_l$ for a given prompt $x$. 
The preference is modeled as a logistic regression, where the binary label (preference outcome) indicates whether response $y_w$ is preferred over $y_l$, and the feature is the scalar difference in reward scores $R_\theta(x, y_w) - R_\theta(x, y_l)$:
\begin{equation} \label{eq:btl_prob}
P(y_w \succ y_l \mid x) = \sigma(R_\theta(x, y_w) - R_\theta(x, y_l)) = \frac{1}{1 + \exp\left( -(R_\theta(x, y_w) - R_\theta(x, y_l)) \right)},
\end{equation}
where $\sigma(\cdot)$ is the sigmoid function.
It is important to note that the reward scores are not fixed input features. Instead, they are outputs of a trainable neural network $R_\theta$.
To train the RM, its parameters $\theta$ are optimized by minimizing the negative log-likelihood of the human-provided preferences in a dataset $D_{\text{pref}} = \{ (x_i, y_{w,i}, y_{l,i}) \}_{i=1}^N$ (where $x_i$ is the $i$-th prompt) using a binary cross-entropy loss:
\begin{equation} \label{eq:pairwise_loss_log_exp}
\mathcal{L}(\theta) = - \mathbb{E}_{(x, y_w, y_l) \sim D_{\text{pref}}} \left[ \log \sigma(R_\theta(x, y_w) - R_\theta(x, y_l)) \right].
\end{equation}
For better numerical stability, this is often expressed in its log-sum-exp form:
\begin{equation} \label{eq:pairwise_loss_log_1_plus_exp}
\mathcal{L}(\theta) = \mathbb{E}_{(x, y_w, y_l) \sim D_{\text{pref}}} \left[ \log\left( 1 + \exp\left( -(R_\theta(x, y_w) - R_\theta(x, y_l)) \right) \right) \right]
\end{equation}

\textit{{Limitations and Refinements.}}
While widely used, the pairwise BTL framework exhibits several limitations that have motivated subsequent research:
(1) A key limitation of the standard BTL model is its assumption that all preferences are of equal strength. It only captures the direction of preference (i.e., $y_w$ is better than $y_l$), not the magnitude (i.e., how much better it is). In practice, a human labeler might find one response to be marginally better, while another might be vastly superior.
To address this limitation, the loss function can be modified to account for the strength of human preference, which is often collected on a rating scale (e.g., a Likert scale). As demonstrated in Llama 2 \citep{touvron2023llama}, a margin term $m(r)$ that is a function of the preference rating $r$ is added to the loss function:
\begin{equation} \label{eq:pairwise_loss_margin}
\mathcal{L}(\theta) = - \mathbb{E}_{(x, y_w, y_l, r) \sim D_{\text{pref}}} \left[ \log \sigma(R_\theta(x, y_w) - R_\theta(x, y_l) - m(r)) \right]
\end{equation}
This modification encourages the RM to create a larger gap in the reward scores for pairs with a stronger preference, leading to a more calibrated and nuanced reward signal.
(2) Another challenge lies in choosing the size and architecture of the reward model. As reported in \citet{askell2021general, ouyang2022training}, large reward models tend to overfit small preference datasets, while smaller models, though more stable, may lack the capacity to learn subtle distinctions in human preferences. 
This trade-off remains an active area of investigation, with the optimal RM size often depending on the specific application, the size of the preference dataset, and the available computational resources.



\textbf{Ranking-based Reward Models.}
To utilize more informative feedback, such as a K-wise ranking of multiple responses ($y_1 \succ y_2 \succ \dots \succ y_k$) or the selection of the single best response from a set of $N$ candidates (Best-of-$N$), which naturally arise in many annotation settings, ranking-based reward models have been proposed.
The key challenge became how to effectively incorporate this ordinal data into a trainable objective. The community has largely adopted two distinct strategies to solve this: (1) a pragmatic \textbf{pairwise decomposition method}, which converts each ranked list into a series of independent pairwise comparisons, and (2) a more statistically principled \textbf{listwise modeling method}, which directly uses the entire ranking.


\textit{Pairwise decomposition method.} The initial and most straightforward strategy is to decompose a full ranking of $k$ items into $\binom{k}{2}$ pairwise preferences. For example, a ranking ($y_1 \succ y_2 \succ y_3$) yields three pairs ($y_1 \succ y_2$), ($y_1 \succ y_3$), and ($y_2 \succ y_3$). Each pair is then used to train an RM using the standard pairwise loss function.
This decomposition approach has been adopted in several foundational RLHF pipelines, including InstructGPT and ChatGPT \citep{stiennon2022learning, ouyang2022training, schulman2022chatgpt}.
The main advantage of this method is its simplicity, as it allows practitioners to reuse the well-established BTL framework without modification. However, this approach has two key limitations: it ignores the holistic context of the full ranking, and it artificially inflates the dataset size, which can be computationally inefficient.

\textit{Listwise modeling method.}
To address aforementioned shortcomings, listwise approaches directly model the probability of an entire ranked list using a statistical framework like the Plackett-Luce model \citep{plackett1975analysis, luce1959individual}.
The parameters $\theta$ of the listwise RM are optimized by minimizing the negative log-likelihood of the observed rankings.
Recent empirical and theoretical work has demonstrated the advantages of listwise learning. For example, \citet{zhu2024principled} show that listwise training is asymptotically more statistically efficient than pairwise decomposition, leading to more accurate reward estimation with fewer annotations. Building on this, \citet{zhu2024starling} introduced a k-wise listwise loss in their Starling-7B reward model, achieving state-of-the-art alignment performance. Despite their promise, listwise models can be more complex to implement and require careful optimization, which has limited their adoption relative to simpler pairwise approaches \citep{cao2007learning, liu2009learning}.

\textbf{Attribute-Based and Outcome-Oriented Reward Models.}
While pairwise and ranking-based reward models have become standard in preference modeling for RLHF, they are limited in settings where relative comparisons are insufficient or ill-defined. 
In domains requiring fine-grained evaluation, such as mathematical reasoning, code generation, or factual correctness, more direct forms of supervision are essential. This has led to the development of alternative reward modeling paradigms that incorporate absolute scoring, outcome validation, process-level critique, and even language-based feedback. 
This section explores reward models trained on: absolute scores (point-wise), verifiable outcomes for reasoning tasks (outcome- and process-supervised), and structured language critiques.

(1) \textbf{Point-wise Reward Models}.
One of the earliest alternatives to pairwise comparison is the use of absolute scalar feedback. In this framework, a reward model is trained to regress onto human-provided scores (e.g., Likert ratings), treating reward learning as a supervised regression problem. The objective typically minimizes the mean squared error (MSE) between the model's output and the target score. While this approach is conceptually straightforward and has been explored in early work \citep{ziegler2019fine}, it suffers from issues related to human annotation noise, such as inter-rater and intra-rater variability. As a result, point-wise RMs are rarely used as standalone reward functions in modern RLHF pipelines. Instead, they are more commonly employed for auxiliary supervision or post-hoc evaluation \citep{stiennon2022learning, wang2025helpsteer2, wang2025una}.


(2) \textbf{Reward Models for Reasoning Tasks} 
For reasoning-intensive tasks, such as mathematical problem-solving or complex code generation \citep{wu2024supercompiler, li2022alphacode}, alignment depends less on subjective preference and more on functional correctness. 
Two specialized types of RMs, Outcome and Process RMs, provide supervision in these domains by evaluating the correctness of the generated solution. 
\textit{Outcome Reward Models (ORMs)} are trained by evaluating the final result of a generated response. Typically, an ORM learns to predict a scalar value representing the probability of a successful outcome, such as a correct mathematical solution or a passing unit test \citep{cobbe2021training, uesato2022solving}. Some implementations provide this feedback on a per-token basis to create a denser reward signal for more efficient learning \citep{cobbe2021training, lyu2025exploring}. Process Reward Models (PRMs), in contrast, provide more fine-grained feedback by evaluating the intermediate steps of a model's reasoning process. Also known as Process-Supervised RMs, this step-by-step supervision is crucial for tasks where the correctness of the reasoning path is as important as the final outcome. Several studies have demonstrated the efficacy of PRMs in enhancing the correctness and interpretability of the reasoning process in complex domains like mathematics \citep{uesato2022solving, lightman2023lets}.

(3) \textbf{Language Feedback Reward Models} 
This approach uses natural language critiques or corrections for reward generation, which can provide dense and precise supervisory signals. Since this feedback is unstructured, it presents unique modeling challenges. Instead of predicting a scalar reward, the RM may be trained as a text-generation model to perform tasks like ``Correction Mapping'' (transforming a rejected response into a chosen one) and ``Identity Mapping" (outputting a chosen response as is). The resulting text outputs are then processed to extract dense, token-level reward signals for the policy optimization phase, offering more granular guidance \citep{zhou2024sequence}.

\textbf{Challenges in Reward Modeling.}
Despite the successes of RLHF, the reward modeling stage presents several challenges that can undermine the alignment, safety, and utility of the final LLMs: (1) \textbf{Reward Misspecification} This occurs when the learned RM fails to faithfully represent true human preferences \citep{peng2023diagnosis, bobu2024aligning}. This discrepancy can arise from the inherent difficulties in collecting comprehensive and nuanced human feedback, but it also stems from a more fundamental limitation: it is statistically impossible for a single reward function to represent a diverse group's preferences if those preferences contain intransitive cycles (e.g., A is preferred to B, B to C, and C to A), a phenomenon known as the Condorcet paradox, meaning any single reward model is an inherently flawed representation of the group's preferences accurately \citep{liu2025statistical}. Consequently, the RM may learn an incomplete or skewed representation of the intended objectives. Humans may struggle to articulate complex preferences or provide feedback that exhaustively covers all desired behaviors, particularly for tasks demanding sophisticated reasoning, creativity, or nuanced ethical considerations. Consequently, the RM may learn an incomplete or skewed representation of the intended objectives.
(2) \textbf{Misgeneralization and Reward Hacking} Another critical issue is the misgeneralization of the reward model, which leads directly to a failure mode known as \textbf{reward hacking} (or overoptimization). Misgeneralization occurs when the RM learns a flawed proxy for human values that fails to generalize robustly to out-of-distribution (OOD) prompts or novel response styles \citep{tien2023causal}. This vulnerability allows the policy, a powerful optimizer, to exploit the RM's inaccuracies during reinforcement learning. The resulting behavior is reward hacking: the LLM achieves a high score according to the flawed RM but fails to align with the actual, more complex human preferences it was meant to capture \citep{gao2023scaling, laidlaw2025correlated, skalse2025defining}. In effect, the LLM finds loopholes in the reward function, for instance, generating overly verbose answers because the RM incorrectly associates length with quality, rather than genuinely satisfying the intended goals.
(3) \textbf{Reliable evaluation} Evaluating the quality of an RM is intrinsically difficult because the ``ground truth" human preference function is unknown. RM performance is typically measured indirectly through the performance of the policy trained on it, making it hard to diagnose and debug issues with the RM itself. Developing robust, direct evaluation metrics for RMs remains an open research problem. Addressing these multifaceted challenges in reward modeling is pivotal for advancing the development of helpful, harmless, and safe aligned LLMs.

\subsection{Policy Optimization Methods in RLHF}  \label{sec:513}

Once a reward model $R_{\theta}$ has been established, the next critical step in RLHF is optimizing the pre-trained language model to align with human preferences. This optimization phase treats the language model as a policy $\pi_{\phi}(a \mid s)$ that must learn to generate actions (tokens) $a$ given states (text sequences) $s$ in ways that maximize rewards from $R_{\theta}$.
The core challenge is transforming a pre-trained LLM into one that consistently produces outputs favored by human evaluators. This transformation requires systematically adjusting the policy parameters $\phi$ to find an optimal policy $\pi_{\phi}^*$ that maximizes expected rewards across diverse input prompts \citep{ziegler2020fine, ouyang2022training}. 

Policy gradient algorithms provide the mathematical framework for this optimization \citep{sutton1999policy}. These methods iteratively update the policy by maximizing an objective function $J(\phi)$ that captures the expected cumulative reward:
\begin{equation} \label{sec52PolicyObjectiveFunction}
J(\phi) = \mathbb{E}_{\tau \sim \pi_{\phi}} [R(\tau)] = \mathbb{E}_{\tau \sim \pi_{\phi}} \left[ \sum_{t=0}^{T} \gamma^t r_t \right],
\end{equation}
where $\tau = (s_0, a_0, s_1, a_1, \dots, s_T, a_T)$ represents a complete trajectory generated by the policy, where $r_t = R_{\theta}(s_t, a_t)$ is the reward at timestep $t$, and $\gamma \in [0, 1]$ is a discount factor balancing immediate versus future rewards.
The policy gradient theorem provides the analytical form of \( \nabla_{\phi}J(\phi) \), which is used to update the policy via gradient ascent. However, in practice, directly applying this gradient often results in unstable or inefficient learning. This motivates the development of alternative algorithms that compute modified or constrained versions of the policy gradient.

Within the broad class of policy gradient methods, several specialized algorithmic families have been developed for RLHF, each tailored to address specific challenges such as stability, efficiency, or feedback sparsity. We categorize the most commonly used approaches as follows. 

\begin{enumerate}
    \item \textbf{Actor-Critic PPO}:  These methods maintain both a policy (actor) and a value function (critic). PPO is the most widely used due to its training stability, achieved by clipping policy updates and using the critic to estimate advantages \citep{schulman2017proximal, ouyang2022training}. 
    \item \textbf{Actor-Only Policy Gradients}:  These methods bypass the critic and directly optimize the policy using rewards from the reward model. While simpler and often more efficient, they may suffer from higher variance in updates and typically require alternative variance reduction strategies.
    \item \textbf{Specialized and Hybrid Reward-Based Methods}: This category includes algorithms designed to address specific RLHF challenges, such as reward hacking, exploration limitations, and training instabilities that arise in large-scale language model optimization.
\end{enumerate}
We summarize existing policy optimization methods in Table \ref{tab:ppo}. In the following subsections, we review each category in detail, with a particular focus on PPO, the most widely adopted method in RLHF.

\subsubsection{Actor-Critic PPO}

PPO was first introduced by \citep{schulman2017proximal} at OpenAI as a solution to the instability and complexity issues inherent in existing policy optimization methods. The algorithm emerged from the recognition that while Trust Region Policy Optimization (TRPO) \citep{schulman2015trust} provided theoretical guarantees for stable policy updates, its implementation required computationally expensive second-order optimization procedures, including conjugate gradient methods and line searches.
PPO is a cornerstone algorithm in many state-of-the-art RLHF pipelines for LLMs. Its extensive utilization in fine-tuning influential models, such as InstructGPT \citep{ouyang2022training}, the models underlying ChatGPT \citep{openai2024gpt4}, and Anthropic's Claude series \citep{askell2021general, bai2022constitutional}, underscores its significance. Furthermore, PPO has been a prevalent choice for aligning various open-source models, including Llama 2-Chat \citep{touvron2023llama}.


The core motivation behind PPO was to retain the stability benefits of trust region methods while dramatically simplifying the implementation and reducing computational overhead. \citep{schulman2017proximal} observed that many policy gradient failures stemmed from excessively large policy updates that could catastrophically degrade performance, leading them to design a method that would naturally constrain update magnitudes without requiring complex second-order computations.

\textbf{Algorithmic Framework.}
PPO addresses the stability concerns of vanilla policy gradients through a clipped surrogate objective that prevents destructively large policy updates. 
At its core, PPO is an actor-critic algorithm. The \textbf{actor} is the LLM policy \( \pi_{\phi}(a \mid s) \), which maps a state \( s \) (e.g., a prompt or partially generated sequence) to a distribution over actions \( a \) (e.g., next tokens). The \textbf{critic} is a value function \( V_{\psi}(s) \), which estimates the expected return from state \( s \) under the current policy, where $\psi$ are parameters in the value function. 
These two components are trained jointly: the actor is updated via a clipped policy gradient objective, while the critic is optimized using a regression loss (e.g., mean-squared error) against the empirical return.

(1) Actor optimization loss. The key innovation in PPO is its clipped surrogate objective,  which prevents large, destabilizing updates by penalizing policy changes that deviate too far from the old policy. Let \( \rho_t(\phi) = \frac{\pi_{\phi}(a_t \mid s_t)}{\pi_{\phi_{\text{old}}}(a_t \mid s_t)} \)  denote the probability ratio between the new policy and the old policy at time step $t$.
PPO seeks to maximize the following objective:
\begin{equation} \label{sec521_PPO_clip_function}
L^{\text{CLIP}}(\phi) = \hat{\mathbb{E}}_t \left[ \min \left( \rho_t(\phi) \hat{A}_\phi(s_t, a_t), \; \text{clip}\left(\rho_t(\phi), 1 - \epsilon, 1 + \epsilon \right) \hat{A}_t \right) \right],
\end{equation}
where $\hat{A}_t$ is an estimate of the advantage function at timestep $t$, quantifying how much better action $a_t$ is compared to the average action in state $s_t$.
This objective ensures that when the policy change is small ($\rho_t \approx 1$), the standard policy gradient is recovered. However, if $\rho_t$ moves outside the interval $[1 - \epsilon, 1 + \epsilon]$, the clipped objective dampens the update, effectively acting as a soft constraint that stabilizes training.

A key component in the Equation \eqref{sec521_PPO_clip_function} is the estimated advantage function, which is formally defined as 
$\hat{A}_t=Q(s_t,a_t) - V_\psi(s_t)$, where the first term \( Q(s_t,a_t) \) is the {action-value function}, representing the expected return when the agent starts in state \( s_t \), takes action \( a_t \), and subsequently follows the current policy. 
The second term \( V_\psi(s_t) \) is the {state-value function}, computing the average return over all actions sampled from the policy’s distribution at that state, i.e., 
$V_\psi(s_t) = \mathbb{E}_{a \sim \pi(\cdot | s_t)} [ Q(s_t, a) ]$.
However, since the true value functions $Q$ and $V_\psi$ are unknown and typically intractable to compute in complex environments, practical algorithms must rely on sample-based approximations. A common and simple approach is to estimate $Q(s_t, a_t)$ using the one-step temporal difference (TD) target \citep{sutton1988learning, sutton2018reinforcement}: $Q(s_t, a_t) \approx r_t + \gamma V_\psi(s_{t+1})$, where $r_t=R_{\theta}(s_t, a_t)$ is the immediate reward after taking action $a_t$ at state $s_t$,
and $V_\psi(s_{t+1})$ is the critic’s prediction of the next state's value. 
This leads to the TD residual estimate of the advantage: $\hat{A}_t = r_t + \gamma V_\psi(s_{t+1}) - V_\psi(s_t)$. While easy to compute, this estimate can be noisy and suffer from high variance, because any small prediction error in $V_\psi$ leads to unstable and misleading advantage estimates.
To address this, Generalized Advantage Estimation (GAE) \citep{schulman2018gae} is often used. GAE provides a bias-variance trade-off by computing a weighted sum of multi-step TD errors. GAE defines the TD error as $\delta_t = r_t + \gamma V(s_{t+1}) - V(s_t)$, and aggregate them as $\hat{A}_t^{\text{GAE}} = \sum_{l=0}^{\infty} (\gamma\lambda)^l \delta_{t+l}$, where $\lambda \in [0, 1]$ is a decay factor.  Smaller $\lambda$ values emphasize short-term estimates with lower variance, while larger $\lambda$ values incorporate longer-term information, reducing bias. As such, GAE enables more stable and data-efficient learning, and has become a standard advantage estimator in modern actor-critic algorithms like PPO.


(2) Critic optimization loss.
While the actor learns via the $L^{\text{CLIP}}$ objective, the critic, $V_{\psi}(s_t)$, learns by minimizing a mean squared error (MSE) loss between its predictions and a target value  
which is calculated using previous (old) critic parameters $\psi_{\text{old}}$:
$\hat{V}^{\text{target}}(s_t)$. 
\begin{equation} \label{sec521_PPO_value_function}
L^{\text{VF}}(\psi) = \hat{\mathbb{E}}_t \left[ (V_{\psi}(s_t) - \hat{V}^{\text{target}}(s_t))^2 \right].
\end{equation}
Here, the choice of the target value $\hat{V}_t^{\text{target}}$ critically impacts training stability and performance. 
A popular choice is to use GAE to calcualte critic target:
$
\hat{V}_t^{\text{target}} = V_{\psi_{\text{old}}}(s_t) + \hat{A}_t^{\text{GAE}}.
$
This creates a clear objective for the critic: its new prediction, $V_{\psi}(s_t)$, should move closer to the old prediction plus any ``surprise'' captured by the advantage. 
To further stabilize training, some PPO implementations also clip the value function loss, preventing excessively large updates to the critic, particularly if value predictions are far off from target values.

(3) Additional loss. 
Two additional terms are commonly introduced to the overall objective. First, to encourage exploration, an entropy bonus is added. This term, $S(\phi)$, incentivizes the policy to maintain randomness in its action choices and is maximized during training:
\begin{equation} \label{sec521_PPO_entropy_bonus}
S(\phi) = \hat{\mathbb{E}}_{t} \left[ H(\pi_{\phi}(\cdot \mid s_t)) \right] = \hat{\mathbb{E}}_{t} \left[ -\sum_{a} \pi_{\phi}(a \mid s_t) \log \pi_{\phi}(a \mid s_t) \right].
\end{equation}
Second, to prevent the LLM from deviating too drastically from a trusted base model (a pre-trained model, or the supervised fine-tuned  model, which possesses coherent language generation capabilities and general knowledge), a common practice is to incorporate a Kullback-Leibler (KL) divergence penalty. This penalty, $D_{\text{KL}}(\pi_{\phi} \parallel \pi_{\text{ref}})$, measures the divergence between the current policy $\pi_{\phi}$ and a frozen reference policy $\pi_{\text{ref}}$ (often a frozen copy of the SFT model or the initial pre-trained model) \citep{schulman2017proximal, schulman2020kl, jaques2017sequence, jaques2020human, ziegler2020fine, ouyang2022training, stiennon2022learning}. 
By regularizing against this reference model, the KL penalty helps mitigate catastrophic forgetting and prevents the policy from generating OOD text that might exploit flaws in the reward model.

Combining these elements, the final objective function for PPO is a weighted sum:
\begin{equation} \label{sec521_PPO_combined_loss}
L(\phi, \psi) = -L^{\text{CLIP}}(\phi) + c_1 L^{\text{VF}}(\psi) - c_2 S(\phi) + c_3 D_{\text{KL}}(\pi_{\phi} \parallel \pi_{\text{ref}}),
\end{equation}
where $c_1$, $c_2$, $c_3$ are coefficients weighting the importance of the value function loss and the entropy bonus, respectively.


\textbf{Limitations of PPO.} 
Despite its popularity, PPO exhibits several limitations that are particularly relevant in the content of LLM alignment.
(1) \textbf{Hyperparameter Sensitivity}: PPO's performance is notably sensitive to hyperparameter configurations, including learning rates, the clipping range $\epsilon$, batch size, and GAE parameters. Tuning these can be resource-intensive and requires careful experimentation to achieve optimal results \citep{schulman2017proximal, engstrom2020implementation}. 
(2) \textbf{Computational and Memory Costs}: PPO imposes substantial computational and memory costs when used for RLHF, creating a significant bottleneck for training large language models. The primary driver of this overhead is the need to hold at least four separate models in GPU memory simultaneously: the trainable Actor (policy), a frozen Reference model, a Reward model, and a Critic. This requirement alone can cause the PPO stage to consume over three times the memory of standard SFT, making it prohibitively expensive and infeasible for many practitioners. In addition to these high static memory costs, the algorithm is computationally intensive due to the repeated inference passes required from these models during the experience generation phase. These combined demands severely limit the scalability and accessibility of PPO-based alignment \citep{ouyang2022training, ramamurthy2023is, rafailov2024direct}.
(3) \textbf{Sample Inefficiency}: as an on-policy algorithm, PPO requires fresh samples generated from the current policy for each update. This makes it inherently less sample-efficient than off-policy algorithms like Soft Actor-Critic (SAC) \citep{haarnoja2018soft} or Deep Q-Networks (DQN) \citep{mnih2015human}, which can reuse past experiences stored in a replay buffer. 
(4) \textbf{Local Optima}: While the clipping mechanism enhances stability, it can also sometimes be overly restrictive, potentially slowing down convergence or leading to a suboptimal policy if the updates are constrained too much.\citep{schulman2017proximal}. 
(5) \textbf{Reward Model Dependence}: The quality of the learned policy is profoundly dependent on the accuracy and reliability of the RM. A misspecified, biased, or exploitable RM can misguide PPO towards undesirable behaviors or ``reward hacking'' \citep{gao2023scaling, ziegler2020fine, stiennon2022learning, openai2024gpt4}.
(6) \textbf{Algorithmic Bias from KL Regularization}: The standard KL divergence penalty used in RLHF introduces an inherent algorithmic bias. Because the reference model ($\pi_{ref}$) is itself not perfectly aligned, its biases are passed to the fine-tuned model. In extreme cases, this can lead to ``preference collapse", where the model learns to completely disregard minority preferences, a bias that persists even with a perfect reward model \citep{xiao2024algorithmic}.
Addressing these challenges remains an active area of research to further enhance the robustness and applicability of PPO in aligning LLMs.

To address these issues, several variants of PPO have been proposed. These can be broadly grouped by the challenge they target:

(1)  \textbf{Enhancing Stability and Mitigating Hyperparameter Sensitivity}. Several variants have been developed to improve PPO’s stability and convergence behavior by refining the core surrogate objective.
KL-regularized PPO (KL-PPO) augments the clipped surrogate loss with an explicit KL divergence penalty, which softly constrains the updated policy to remain close to the previous one, functioning as a trust-region regularizer \citep{ouyang2022training}. An extension of this idea, adaptive KL control, dynamically adjusts the strength of the KL penalty based on the observed divergence between the current and reference policies, preventing the policy from straying too far from a known-good distribution \citep{ziegler2020fine}. More recently, \citep{shen2024policy} introduced Policy Filtration for PPO (PF-PPO) to address instability arising from inaccurate reward models. Motivated by the observation that RMs, particularly in complex domains such as code generation, are more reliable at ranking extremely good or bad samples than moderately scored ones. PF-PPO filteres out low-confidence trajectories and concentrates policy learning on those with more trustworthy reward signals, thereby reducing reward hacking and improving alignment performance. More recently, Proximal Policy Optimization with Reward-based Prioritization (RP-PPO) was proposed to combat performance degradation and slow convergence in later training stages. RP-PPO dynamically adjusts the number of policy update epochs based on reward quality; high-quality experiences are given greater weight through more training rounds, providing an incentive for the model to move towards higher-reward policies. Furthermore, RP-PPO explicitly saves the model that achieves the highest historical average reward, rather than relying on the final model, thus preventing the loss of the best-found policy due to later training instability \citep{zheng2025proximal}.

(2) \textbf{Reduce computational and memory costs.}
Some system-level innovations were implemented to address this limitation. To reducing the GPU memory footprint, for example, Hydra-PPO introduces techniques like model sharing and ``Dynamic LoRA", where a single base model is used for multiple roles (e.g., Actor and Reference) by dynamically activating or deactivating LoRA adapters, significantly cutting down the number of full models that need to be stored in VRAM \citep{santacroce2023efficient}. This approach builds on the general use of PEFT methods like LoRA, which drastically lower the memory required for optimizer states by only training a small fraction of the model's parameters \citep{hu2021lora}. Some improvements focus on scaling PPO across multiple machines for faster wall-clock training times. Standard distributed methods can suffer from communication bottlenecks. Decentralized Distributed PPO (DD-PPO) addresses this by eliminating the central parameter server and using direct peer-to-peer gradient synchronization among workers. To prevent slowdowns from non-uniform workloads, DD-PPO also introduces a preemption mechanism to mitigate the ``straggler effect", where fast workers would otherwise wait for the slowest one \citep{wijmans2020ddppo}. Together, these system-level optimizations are critical for making PPO a practical and efficient algorithm for state-of-the-art LLM alignment.

(3) \textbf{Improving Sample Efficiency.}
To enhance PPO's sample efficiency, a major line of research focuses on creating hybrid algorithms that safely incorporate off-policy data from a replay buffer, blending the stability of PPO with the data efficiency of off-policy learning.
These hybrid methods vary in their approach. One strategy focuses on the gradient update itself; for example, P3O (Policy-on Policy-off Policy-over) interleaves on-policy and off-policy gradient updates and uses the Effective Sample Size (ESS) to automatically control the trade-off, allowing it to effectively leverage past data without introducing new, sensitive hyperparameters \citep{fakoor2019p3o}. 
Another key strategy involves redesigning the training loop. Phasic Policy Gradient (PPG) separates training into two distinct, alternating phases: a policy phase for standard policy updates and an auxiliary phase that exclusively reuses collected experience to more aggressively train the value function. This decoupling allows for better-trained features while mitigating interference between the policy and value objectives. Building on the concept of experience replay \citep{cobbe2020phasic}. 
Building on the concept of experience replay, Hybrid-Policy PPO (HP3O) utilizes a trajectory replay buffer with a ``first-in, first-out" (FIFO) strategy to limit data distribution drift by only using recent policies. It further guides learning by always including the trajectory with the best return from the buffer in each training batch \citep{liu2025enhancing}. 
While these methods focus on empirical performance, others like Transductive Off-policy PPO (ToPPO) aim for stronger theoretical guarantees. ToPPO introduces a novel transductive inference mechanism to justify using the advantage function directly from the data-generating policy, avoiding common estimation biases. This allows it to safely integrate off-policy data while striving for monotonic policy improvement \citep{gan2024transductive}.
Although their mechanics differ, these variants all showcase a powerful trend towards making PPO more data-efficient by safely and strategically reusing past experiences.

\subsubsection{Actor-Only Policy Gradients} \label{sec:critic_less_pg}
Actor-only methods specifically refer to algorithms that optimize the policy directly without explicitly training or maintaining a separate value-function (critic). Instead, these methods use simplified baselines computed from reward signals to reduce variance in policy gradient updates. This significantly simplifies training because we do not need to fit or maintain a critic network.
Broadly,  based on how each method calculates the baseline value used for computing the advantage in policy gradient updates, 
existing actor-only methods can be categorized into three groups: (1) the foundational \textbf{REINFORCE algorithm}, which represents the classic Monte Carlo approach to policy gradients and typically uses a simple, historically averaged reward baseline;
(2) methods based on \textbf{Intra-Prompt Comparison}, which create a localized baseline from multiple responses to a single prompt;
and (3) methods employing \textbf{Batch-Wise Normalization}, which derive a more global baseline from the statistics of an entire mini-batch.

\paragraph{
The Foundational Method: REINFORCE. }
The REINFORCE algorithm is the original Monte Carlo policy gradient method \citep{williams1992simple}. 
In its most basic form, vanilla REINFORCE works by sampling a complete response trajectory $y$ from the policy $\pi_\phi$ and updating the policy parameters to increase the probability of trajectories that received high rewards. 
The intuition is straightforward: if a generated response $y$ receives a high reward $R(x,y)$, we increase its probability by moving the policy parameters in the direction of $\nabla_\phi \log \pi_\phi(y|x)$. Conversely, low-reward responses have their probabilities decreased.
While conceptually elegant, vanilla REINFORCE suffers from extremely high variance in its gradient estimates. Since the algorithm uses the full reward $R(x,y)$ to weight each update, small random fluctuations in rewards can cause dramatic changes in gradient estimates. This leads to unstable training where the policy may oscillate wildly rather than converging to optimal behavior.

To address this variance issue, REINFORCE introduces a baseline $b(x)$ that is subtracted from the reward without introducing bias into the gradient estimate:
\begin{equation} \label{eq:reinforce_gradient}
 \nabla_\phi J(\phi) \approx \mathbb{E}_{x \sim \mathcal{D}, y \sim \pi_\phi(\cdot|x)} [(R(x,y) - b(x)) \nabla_\phi \log \pi_\phi(y|x) ]
\end{equation}
The baseline typically represents an estimate of the expected reward for input $x$, such as a moving average of past rewards. By centering the rewards around this baseline, the algorithm reduces variance: responses that perform better than average receive positive weight, while below-average responses receive negative weight.
Despite the baseline improvement, REINFORCE with simple baselines remains  sample-inefficient and high-variance for state-of-the-art LLM alignment tasks. The method still requires complete trajectory rollouts and provides learning signals only at the episode level.

\paragraph{(2) Intra-Prompt Comparison: Localized Baselines.}
A powerful way to create a better baseline is to have the model compete with itself. This family of methods generates multiple candidate responses for a single prompt and then uses their relative rewards to create a stable and highly localized advantage signal.

\textbf{REINFORCE with Leave-One-Out Baseline (RLOO)}.
The earliest and most influential intra-prompt comparison method is RLOO, which extends basic REINFORCE introducing a sophisticated, prompt-specific baseline from a group of $K$ sampled responses to reduce variance \citep{kool2019buy, ahmadian2024back}. Instead of relying on a globally trained critic like PPO or a simple moving average baseline like REINFORCE, RLOO calculates the baseline for any given response as the average reward of the other $K-1$ peer responses from the same prompt:
\begin{equation}
b_{\text{RLOO}}(x, y^{(i)}) = \frac{1}{K-1} \sum_{j=1, j \neq i}^{K} R_\theta(x, y^{(j)})
\label{eq:rloo_baseline}
\end{equation}
This localized baseline effectively captures prompt-specific difficulty and provides more stable variance reduction than global averages. However, RLOO's fundamental limitation is its {computational cost} \citep{gao2024rebel, hu2025reinforcepp}: requiring $K$ samples per update makes it $K$ times more expensive than basic REINFORCE.
This makes RLOO a pragmatic choice in scenarios where multiple candidates are already being generated, such as for best-of-$N$ sampling, as it efficiently reuses those forward passes for variance reduction \citep{kool2019buy, ahmadian2024back, hu2025reinforcepp}.
Furthermore, the baseline's effectiveness is critically dependent on the diversity and quality of the sampled responses; if the $K$ completions are too similar or contain outliers, the baseline's ability to reduce variance is diminished \citep{gao2024rebel}.

To address the efficiency bottleneck of RLOO, the \textbf{ReMax} method \citep{li2024remax} proposes a simpler baseline: for each prompt, generate just two responses, a greedy (deterministic, most-likely) output and a stochastic sample. The greedy response serves as a cheap, prompt-specific baseline:
\begin{equation}
A_{\text{ReMax}}(x, y_{\text{samp}}) = R_\theta(x, y_{\text{samp}}) - R_\theta(x, y_{\text{greedy}})
\end{equation}
This reduces the number of forward passes required per update from 
$K$ (in RLOO) to just two, greatly improving efficiency. 
Empirical results show that this technique substantially reduces the variance of the policy gradient compared to vanilla REINFORCE, leading to stable convergence \citep{li2024remax}.
However, the effectiveness of this method hinges on the quality of the greedy baseline itself; a poorly trained policy might yield an uninformative greedy sample, potentially limiting variance reduction or even increasing it in certain policy states \citep{li2024remax, hu2025reinforcepp}.



As intra-prompt baselines evolved, researchers recognized that not only the mean but also the variability of rewards across a prompt’s responses can carry important information. 
\textbf{Group Relative Policy Optimization (GRPO)} \citep{shao2024deepseekmath, deepseekai2025v3, deepseekai2025r1} addresses a fundamental statistical limitation of both RLOO and ReMax: \textbf{reward scale sensitivity}. 
While previous methods center rewards around baselines, they don't account for varying prompt difficulties or reward scales. GRPO normalizes the advantage signal by computing standardized scores:
\begin{equation} \label{eq:grpo_advantage}
A_i = \frac{r_i - \text{mean}(\{r_1, \dots, r_G\})}{\text{std}(\{r_1, \dots, r_G\}) + \epsilon_{\text{norm}}}
\end{equation}
where $\epsilon_{\text{norm}}$ is a small constant for numerical stability.
This  makes the advantage robust to prompt-specific reward scales and difficulties, allowing stable learning even across highly diverse prompts.  
However, this powerful normalization introduces its own limitations: (1) High generation cost. It requires generating $G$ responses per prompt to calculate the group-based advantage, which increases the cost of each training update. (2) Instability and potential bias. Reliance on the group's standard deviation for normalization can create unintended biases, such as over-rewarding groups of responses that have very low reward variance (e.g., all are mediocre) or amplifying noise if the group size $G$ is too small. (3) Overfitting on Simple Prompts. Because GRPO calculates baselines separately for each prompt using these group statistics, it might be prone to overfitting simpler prompts if not carefully managed. (4) Sensitivity to sampling strategy. Its overall performance can be sensitive to the number of samples $G$ per group and the diversity (and thus the reward variance) of responses within those groups, and it can struggle with imbalanced data, as these factors influence the stability and informativeness of the advantage estimates \citep{hu2025reinforcepp}.

Recent variants address GRPO's specific limitations. \textbf{Dr. GRPO} \citep{liu2025understanding} tackles optimization biases by selectively removing normalization terms to address length and difficulty biases. \textbf{DISCO} \citep{zhou2025disco} addresses multi-domain performance issues through domain-aware reward scaling. These refinements demonstrate the ongoing evolution of the intra-prompt comparison paradigm, with each method building upon its predecessors' strengths while addressing their specific computational, statistical, or application-specific limitations.


\paragraph{(3) Batch-Wise Normalization: A Global Perspective.}
To address the computational expense and overfitting risks of intra-prompt baselines, batch-wise normalization computes a global baseline from a mini-batch, typically using only one completion per prompt. A key example is \textbf{REINFORCE++} \citep{hu2025reinforcepp}, which builds on the REINFORCE algorithm. REINFORCE++ calculates token-level advantages, incorporating the sequence's reward and a penalty to prevent the policy from drifting from a reference. Crucially, it normalizes these advantages using the mean and standard deviation of the entire batch. This global normalization provides a more stable learning signal across diverse prompts. The method also integrates a PPO-like clipped objective to prevent destructive policy updates.
This approach is more sample-efficient, robust to reward hacking, and shows better generalization than intra-prompt methods. However, a global baseline can be less sensitive to prompt-specific nuances, and its performance may be affected by batch composition. Like other actor-only approaches, it may also exhibit higher variance than actor-critic methods, though its normalization and clipping features effectively mitigate this.

\subsubsection{Specialized and Hybrid Reward-Based Policy Optimization} \label{sec:specialized_hybrid_rpo}

Standard RLHF uses a single reward for an entire generated sequence, making it hard to assign credit to specific token decisions, a major inefficiency, especially for long-form generation \citep{zhong2025dpo}. 
Another key problem in RLHF is ``reward hacking", where the policy exploits weaknesses in the learned reward model by generating OOD outputs that receive high (but untrustworthy) scores \citep{wu2024self}.
To address these foundational issues, a new class of specialized and hybrid methods has been developed. These approaches go beyond simply refining the policy optimizer and instead restructure core components of the RL problem itself: the optimization objective, the reward signal, and the policy's exploration space. 
In this subsection, we organize recent advances into two principal categories, based on {how they reformulate the RLHF problem:}
(1) \textbf{Dense Reward Signal Methods}: Approaches that address the credit assignment and sample efficiency challenge by transforming the reward from sparse, sequence-level feedback to dense, token-wise feedback.
(2) \textbf{Exploration-Constrained Methods}: Approaches that mitigate reward over-optimization and improve robustness by explicitly restricting the policy's exploration to regions where the reward model is reliable.
For each, we review the foundational method, analyze its limitations, and describe follow-up innovations.

\paragraph{Dense Reward Signal Methods.}
\textbf{{Reinforced Token Optimization (RTO)}}  \citep{zhong2025dpo} is the primary example of this approach. It is not a replacement for an optimizer like PPO but rather a powerful enhancement that provides it with a much better learning signal. 
RTO is the first major approach to recast RLHF as a token-level Markov Decision Process \citep{zhong2025dpo}. It introduces a dense, per-token reward by leveraging the DPO (Direct Preference Optimization) \citep{rafailov2024direct} objective. Here, each token in a sequence receives its own reward based on how much more likely it is under the DPO-trained policy than a reference policy. This results in a much richer, more actionable learning signal that substantially improves credit assignment and sample efficiency. Experiments show that RTO can achieve superior or comparable performance to PPO with less data \citep{zheng2023secrets}.

Despite these strengths, this framework has its own limitations: (1) Dependence on DPO-derived rewards: The framework's success is contingent on the policy learned by DPO ($\pi_{DPO}$) being a good proxy for the true optimal policy. A poorly trained DPO model will result in a noisy or misaligned token-wise reward signal. While making rewards dense, ensuring these token-level signals are consistently meaningful remains a broader challenge. (2) Pipeline Complexity: The RTO pipeline is inherently multi-stage, requiring a DPO training run to generate the reward signal before the PPO optimization phase can begin, which adds operational complexity \citep{zhong2025dpo}.

\paragraph{Exploration-Constrained Methods.}
Exploration-constrained methods focus on keeping the policy within the ``trusted'' region of the reward model, thereby reducing the risk of such exploitation.
\textbf{Behavior-Supported Policy Optimization (BSPO)} \citep{dai2025mitigating} is a principled approach that constrains policy optimization to the region where the reward model is known to be reliable. 
BSPO defines a ``behavior policy" ($\beta$) using the next-token distribution from the RM's training dataset, which delineates an in-distribution (ID) region for the RM. It then employs a ``behavior-supported Bellman operator" that regularizes the value function by assigning a low Q-value ($Q_{min}$) to actions $a$ at state $s$ leading out of this ID region (OOD actions, where the behavior policy assigns near-zero probability, i.e., $\beta(a|s) \approx 0$), while leaving ID action-values unchanged (i.e., $\beta(a|s) > 0$). 
By doing so, BSPO directly addresses reward hacking and ensures that policy improvement happens only where the reward model's judgments can be trusted.


The primary advantages offered by this behavior-supported approach are: (1) Directly Mitigates Reward Hacking: By explicitly penalizing OOD actions at the value-function level, BSPO prevents the policy from exploring and exploiting regions where the reward model is known to be unreliable. (2) Finds Optimal In-Distribution Policy: This method penalizes only OOD values without altering the values of ID actions. This allows it to fully explore the supported region and find the optimal policy within that space, whereas other methods like a uniform KL penalty can be overly conservative \citep{dai2025mitigating}.
However, the framework has notable limitations: (1) Dependence on Behavior Policy Definition: The method's effectiveness is highly dependent on how well the defined ``behavior policy" represents the true region of the RM's competence. A poorly defined ID region could either unduly restrict beneficial exploration or fail to prevent hacking. (2) Primary Validation in Synthetic Setups: This method's performance has been validated primarily in synthetic setups with a ``gold" reward model. The dynamics of OOD detection may differ in real-world scenarios with noisy and inconsistent human feedback. (3) Response Scope: The method is designed to handle OOD responses but does not address the separate challenge of the model receiving an OOD prompt, for which no valid ID response may exist.\\

In summary, the optimization of an LLM's policy via a learned reward model has spurred innovation along three parallel streams. The first involves refining the established and industry-standard \textbf{Actor-Critic PPO framework} to enhance its performance across stability, efficiency, and scalability while reducing its significant computational footprint. A second stream focuses on developing simpler, \textbf{Actor-Only Alternatives} that streamline the RLHF pipeline and reduce computational overhead by replacing the complex critic with engineered baselines for variance reduction. The third stream moves beyond pure optimization to fundamentally restructure the RL problem itself, with \textbf{Specialized and Hybrid Methods} designed to overcome core challenges like reward hacking and sparse feedback. A detailed comparison of the key methods emerging from these streams, outlining their respective advantages and limitations, is provided in Table \ref{tab:ppo}.

\newcolumntype{L}[1]{>{\raggedright\arraybackslash}p{#1}}
\begin{longtable}{@{} L{2.1cm} L{3.8cm} L{5cm} L{5cm} @{}}
\caption{Policy Optimization Methods in RLHF}
\label{tab:ppo}
\\
\toprule
\textbf{Category} & \textbf{Method} & \textbf{Key Advantages} & \textbf{Key Limitations} \\
\midrule
\endfirsthead

\toprule
\textbf{Category} & \textbf{Method} & \textbf{Key Advantages} & \textbf{Key Limitations} \\
\midrule
\endhead


\bottomrule
\endlastfoot

\multirow[t]{20}{*}{\textbf{Actor-Critic}}
& \textbf{PPO} \newline \citep{schulman2017proximal} & Highly stable; performant; industry standard. & High memory/compute cost; sensitive to hyperparameters. \\
\cmidrule(l){2-4}
& \textbf{KL-PPO} \newline \citep{ouyang2022training} & Prevents policy deviation and enhances stability via KL penalty. & Adds complexity of tuning the KL coefficient. \\
\cmidrule(l){2-4}
& \textbf{PF-PPO} \newline \citep{shen2024policy} & Mitigates reward hacking by filtering data with unreliable reward signals. & Depends on RM's ability to score extreme samples accurately. \\
\cmidrule(l){2-4}
& \textbf{RP-PPO} \newline \citep{zheng2025proximal} & Combats performance decay by giving more training epochs to high-quality data. & Adds logical complexity and hyperparameters for dynamic adjustment. \\
\cmidrule(l){2-4}
& \textbf{Hydra-PPO} \newline \citep{santacroce2023efficient} & Reduces memory cost significantly via LoRA-based model sharing. & Requires complex system-level engineering. \\
\cmidrule(l){2-4}
& \textbf{DD-PPO} \newline \citep{wijmans2020ddppo} & Improves scaling with decentralized updates for faster large-scale training. & Increases implementation complexity. \\
\cmidrule(l){2-4}
& \textbf{P3O} \newline \citep{fakoor2019p3o} & Improves sample efficiency by mixing on-policy and off-policy updates. & Risks instability from off-policy data distribution shift. \\
\cmidrule(l){2-4}
& \textbf{PPG} \newline \citep{cobbe2020phasic} & Improves sample efficiency by separating policy and value training phases. & More complex two-phase training loop. \\
\cmidrule(l){2-4}
& \textbf{HP3O} \newline \citep{liu2025enhancing} & Improves sample efficiency by reusing trajectories from a replay buffer. & Performance depends on buffer size; risks using stale data. \\
\cmidrule(l){2-4}
& \textbf{ToPPO} \newline \citep{gan2024transductive} & Improves sample efficiency with theoretically justified off-policy data use. & Relies on novel theoretical assumptions. \\
\midrule

\multirow[t]{14}{*}{\textbf{Actor-Only}}
& \textbf{REINFORCE} \newline \citep{williams1992simple} & Foundational simplicity; no critic to train or manage. & High gradient variance and sample inefficiency. \\
\cmidrule(l){2-4}
& \textbf{RLOO} \newline \citep{kool2019buy, ahmadian2024back} & Reduces variance with an adaptive, prompt-specific baseline from peer responses. & High sample generation cost per prompt; variance still higher than PPO. \\
\cmidrule(l){2-4}
& \textbf{ReMax} \newline \citep{li2024remax} & Computationally cheap baseline using a single greedy output; very sample efficient. & Effectiveness is entirely dependent on the quality of the greedy baseline. \\
\cmidrule(l){2-4}
& \textbf{GRPO} \newline \citep{shao2024deepseekmath} & Robust to prompt difficulty via reward normalization; effective for reasoning. & High sample generation cost per prompt; can overfit or be biased by group stats. \\
\cmidrule(l){2-4}
& \textbf{Dr. GRPO} \newline \citep{liu2025understanding} & Targets and mitigates specific GRPO biases. & Improvements may be task-specific. \\
\cmidrule(l){2-4}
& \textbf{DISCO} \newline \citep{zhou2025disco} & Improves GRPO performance on multi-domain datasets. & Adds complexity; requires domain heuristics or labels. \\
\cmidrule(l){2-4}
& \textbf{REINFORCE++} \newline \citep{hu2025reinforcepp} & Improves generalization and OOD performance with a global, batch-wise baseline. & Global baseline is less responsive to individual prompt difficulty. \\
\midrule

\multirow[t]{4}{*}{\textbf{Specialized}}
& \textbf{RTO} \newline \citep{zhong2025dpo} & Solves the credit assignment problem with dense, per-token rewards. & Quality is contingent on the DPO-derived rewards; multi-stage pipeline. \\
\cmidrule(l){2-4}
& \textbf{BSPO} \newline \citep{dai2025mitigating} & Mitigates reward hacking by constraining the policy to a trusted data region. & Effectiveness depends on a well-defined behavior policy; may stifle useful exploration. \\

\end{longtable}

\subsection{Challenges of RLHF} 

RLHF has transformed language model alignment by enabling models to optimize directly for human preferences rather than just next-token prediction. The fine-tuning of GPT-3 into InstructGPT, for instance, showed that a 1.3B parameter model trained with RLHF could outperform the much larger 175B parameter base model on human preference ratings \citep{ouyang2022training}. 
However, despite dramatic progress, RLHF also exposes some challenges spanning feedback collection, reward modeling, and policy optimization. Addressing these is essential for advancing robust and scalable alignment.


(1) Human feedback data bottlenecks.
RLHF pipelines depend on human feedback, which is costly, slow to scale, and prone to noise, cognitive biases, and low inter-annotator agreement \citep{stiennon2022learning, ziegler2020fine, bai2022helpful}. This challenge is magnified by the problem of value pluralism; human preferences are not monolithic, and the attempt to aggregate diverse or even conflicting values into a single reward function can lead to a model aligned with a statistical ``average" that satisfies no one, a limitation with theoretical roots in paradoxes of social choice \citep{liu2025statistical}.

(2) Reward Model Limitations and Vulnerabilities.
The RM, trained on limited and noisy human feedback, serves only as an imperfect proxy for true human intent.
As we discussed previously, human annotations can be inconsistent, noisy, and subject to individual biases. Moreover, the data collected often covers only a subset of possible scenarios, prompts, and preferences, and may not fully reflect the diversity or subtlety of real human values. As a result, the RM tends to learn patterns that fit the observed feedback rather than capturing the underlying intent behind human judgments. This data limitation, combined with the complexity of modeling nuanced human preferences, makes the RM susceptible to reward misspecification, where its assigned scores do not accurately correspond to what users actually prefer, but instead reflect artifacts or gaps in the training data \citep{peng2023diagnosis, bobu2024aligning}.
Misspecified rewards enable \textbf{reward hacking}, where policies generate outputs that maximize RM scores without genuine alignment \citep{gao2023scaling, laidlaw2025correlated}. Common manifestations include verbose responses exploiting length biases and sycophantic outputs exploiting agreement preferences \citep{shen2023looselipssinkships, liu2025rrmrobustrewardmodel}.


(3)  Policy Optimization Complexity and Instability.
State-of-the-art optimization methods like PPO are resource-intensive and sensitive to hyperparameters \citep{ouyang2022training, ramamurthy2023is}. 
 In addition, the optimizer is tasked with solving a difficult credit assignment problem, as a single, sparse reward for an entire sequence provides an inefficient signal for determining which specific token choices were beneficial \citep{zhong2025dpo}. When this complex optimization process is aimed at maximizing a brittle and misspecified reward signal, the entire pipeline becomes unstable and prone to producing misaligned behaviors.\\

\section{SFT versus RLHF: Differences, Equivalences, and Hybrid Approaches}


Having comprehensively introduced the foundation of SFT and RLHF, we now proceed to a more in-depth comparison between the two alignment strategies from multiple perspectives. SFT and RLHF differ significantly in their objectives, data requirements, and optimization paradigms, resulting from the inherent methodology and theory. Importantly, these approaches are not mutually exclusive. On the contrary, they are often complementary, and when effectively integrated, can yield more robust and nuanced model alignment. In this section, we systematically examine the differences between SFT and RLHF and explore how their strengths can be combined within hybrid training pipelines.

\subsection{Fundamental Differences between SFT and RLHF} 

At a fundamental level, SFT involves fine-tuning a pre-trained model on labeled datasets \citep{wang2022self, taori2023stanford, peng2023instruction}, whereas RLHF incorporates human feedback into a reward function to guide learning in a reinforcement learning framework \citep{ouyang2022training, bai2022training}. These underlying principles give rise to different objectives, data requirements, and learning dynamics.

\textbf{\textit{Objectives and reward function.}} 
SFT optimizes a standard supervised learning objective, typically minimizing the token-level cross-entropy loss between the model's next output token and human-labeled reference responses \citep{brown2020language}. This objective assumes the existence of a single correct output per input and encourages the model to mimic these ground truth answers. Formally, given an input \( x \) and reference output (labeled data) \( y = (y_1, y_2, \dots, y_T) \), the loss function \citep{fan2025weighted,pmlr-v202-mao23b} as shown in section 4.3 is
$\mathcal{L}_{\mathrm{SFT}} = - \sum_{t=1}^{T} \log P_\theta(y_t \mid x, y_{<t}),$
where \( P_\theta(y_t \mid x, y_{<t}) \) is the probability of generating correct token \( y_t \) given the input and previous tokens \( y_{<t} \). With the estimated model parameters $\hat{\theta}=\mathrm{argmin}_\theta\mathcal{L}_{\mathrm{SFT}}$, the output of the fine-tuned model should be close to the reference one. In contrast, RLHF exploits a reinforcement learning framework. The model acts as a policy \( \pi_\theta \) that generates responses to maximize a scalar reward signal. This reward is not derived from a correct reference, but rather from a learned reward model trained to reflect human preferences. Given a sampled response \( y \sim \pi_\theta(\cdot \mid x) \), the reward model assigns a scalar score \( r(x, y) \). Then we could use RL (e.g.,
Proximal Policy Optimization \citep{schulman2017proximal}) to directly
optimize the expected reward as \citep{ziegler1909fine}:
$
\mathbb{E}_{\pi}(r) =  \mathbb{E}_{x \sim \mathcal{D}, y \sim \pi_\theta(\cdot \mid x)}[r(x, y)]),$
which allows the model to explore a flexible solution space and optimize for outputs that align with nuanced human expectations, especially in ambiguous or multi-faceted scenarios.

\textbf{\textit{Data requirements.}}
The distinct optimization targets of SFT and RLHF lead to different data requirements. SFT relies on high quality instruction–response pairs that provide explicit supervision for the model to mimic reference answers \citep{dong2023abilities,chowdhery2023palm}, and the powerful effect of data quality on LLM performance that can surpass the amount of data \citep{zhou2023lima}.
In contrast, RLHF depends on preference-based data. Annotators are presented with multiple responses for the same prompt and are asked to rank or select the preferred one. These comparative judgments are used to train a reward model that approximates human preferences \citep{bai2022training, liu2020learning}. This reward model guides the language model during reinforcement learning.
Thus, SFT requires extensive, high quality labeled data, while RLHF necessitates labor-intensive human preference collection. Each method poses distinct challenges in data acquisition that must be addressed to ensure effective model alignment \citep{yin2024entropy,lee2023rlaif}.

\textbf{\textit{Learning dynamics and generation.}}
SFT is a static, one shot learning process. It aligns the model by providing token level supervision based on reference answers, which may introduce biases in preference estimation \citep{hua2024intuitive}. Once fine-tuned, SFT models lack adaptability to evolving objectives unless retrained on newly curated datasets. Furthermore, for tasks involving long-form generation, evaluating individual tokens may be suboptimal compared to assessing full responses. In contrast, RLHF emphasizes sentence level performance. It samples entire responses and aligns model behavior with human preferences over those responses. RLHF typically proceeds in two stages: reward model training and reinforcement fine-tuning \citep{bai2022training, openai2023gpt}. In each RL iteration, the model samples outputs according to its current policy, receives feedback via the reward model, and updates its policy to increase the likelihood of generating preferred outputs \citep{zheng2023secrets}. Online RLHF settings \citep{pmlr-v235-xiong24a, dong2024rlhf, ye2024online} may further refine the reward model and iteratively update the policy, allowing continual alignment improvements through repeated sampling, feedback, and optimization. Despite its flexibility, RLHF can be unstable and computationally demanding, especially in online learning schemes requiring coordination among multiple components \citep{rafailov2023direct, ouyang2022training, yuan2023rrhf,ethayarajh2024kto}. SFT provides training stability and efficiency due to its straightforward optimization, often resulting in faster convergence and lower resource requirements \citep{du2025simplify}.

\subsection{When SFT and RLHF Overlap or Converge} 

While SFT and RLHF differ operationally \citep{ouyang2022training}, they exhibit surprising methodological overlaps \citep{chen2024self,swamy2025all}. These overlaps manifest most prominently in their theoretical equivalence under shared representational constraints, loss function alignment through regularization mechanisms, and convergence to identical optimal policies under idealized conditions \citep{Williams1992}. Such synergies challenge the conventional separation of imitation learning and preference-based alignment, revealing a unified landscape for optimizing model behavior.

\textbf{\textit{Overlap for theoretical equivalence.}}
Under the assumption of isomorphic function classes, where policy networks and reward models share equivalent representational capacity, both SFT and RLHF converge to the same optimal policy \citep{chen2024self}. Despite employing distinct algorithmic strategies, these approaches exhibit theoretical equivalence in their alignment objectives. While SFT maximizes likelihood over human labeled data and RLHF optimizes expected reward under a learned preference model, their solutions coincide when analyzed through a shared reward-functional lens. This equivalence is further clarified through the lens of Direct Preference Optimization (DPO), which reformulates RLHF as direct reward maximization, bridging the gap between the two paradigms \citep{swamy2025all}.

\textbf{\textit{Overlap for loss function alignment.}} 
Under the assumption of linearly isomorphic function classes, where policy and reward models share equivalent representational capacity, the gradient updates of SFT and Supervised Policy Improvement via Reinforcement Learning (SPIN) through online DPO can be shown to be formally aligned. In this regime, SFT optimizes the policy parameters $\theta$ using the standard maximum likelihood gradient \citep{Williams1992}, i.e.,
$
g_{\mathrm{SFT}}(\pi_{\theta}) = 
\mathbb{E}_{\xi \sim D_{\mathrm{Sft}}} \left[ 
  \sum_{h=1}^{H} -\nabla_{\theta} \log \left( \pi_{\theta}(a_h | s_h) \right) 
\right]$, while SPIN modifies this by introducing an on-policy correction term, i.e.,
$
g_{\mathrm{SPIN}}(\pi_{\theta}) = g_{\mathrm{SFT}}(\pi_{\theta}) 
- \mathbb{E}_{\xi \sim \pi_{\theta}} \left[ 
  \sum_{h=1}^{H} -\nabla_{\theta} \log \left( 
    \frac{\pi_{\theta}(a_h | s_h)}{\pi_{\mathrm{ref}}(a_h | s_h)} 
  \right) 
\right].$ 
Crucially, this additional term is equivalent to a reinforce-style update with constant reward $r(\xi) = 1$, and thus integrates to zero in expectation. As a result, SPIN's update rule asymptotically reduces to SFT when function classes are linear and expressive enough to perfectly model the optimal policy. This equivalence reveals a deep structural overlap between offline supervised learning (SFT) and online preference-based reinforcement learning (SPIN/RLHF), unifying them within a shared loss geometry in the linear regime \citep{rafailov2023direct}.

\textbf{\textit{Convergence conditions for RLHF and SFT.}} 
The convergence between RLHF and SFT performance hinges on what the authors term the disparity between the complexity of generating outputs via policies and verifying them via reward models \citep{swamy2025all}. This gap becomes pivotal when training a verifier is significantly easier than training the generator \citep{rafailov2023direct, Sun2024}. In such settings, RLHF effectively restricts policy search to regions that perform well under simplified reward models, while SFT directly optimizes the policy through maximum likelihood estimation. In particular, two specific regimes lead to alignment in the final performance of RLHF and SFT. First, in tasks with low sequence complexity, such as two-word summarization, the generation complexity closely matches verification complexity \citep{Li2010}, minimizing the utility of RLHF’s reward modeling \citep{swamy2025all}. Second, when reward functions are simple and easily computed, the verification process is trivial \citep{Lin2004}, reducing RLHF to a form of unnecessary refinement \citep{swamy2025all}.

\subsection{Integrating SFT and RLHF in Training Pipelines} 

\begin{figure}[ht!]
    \centering
    \includegraphics[width=\linewidth]{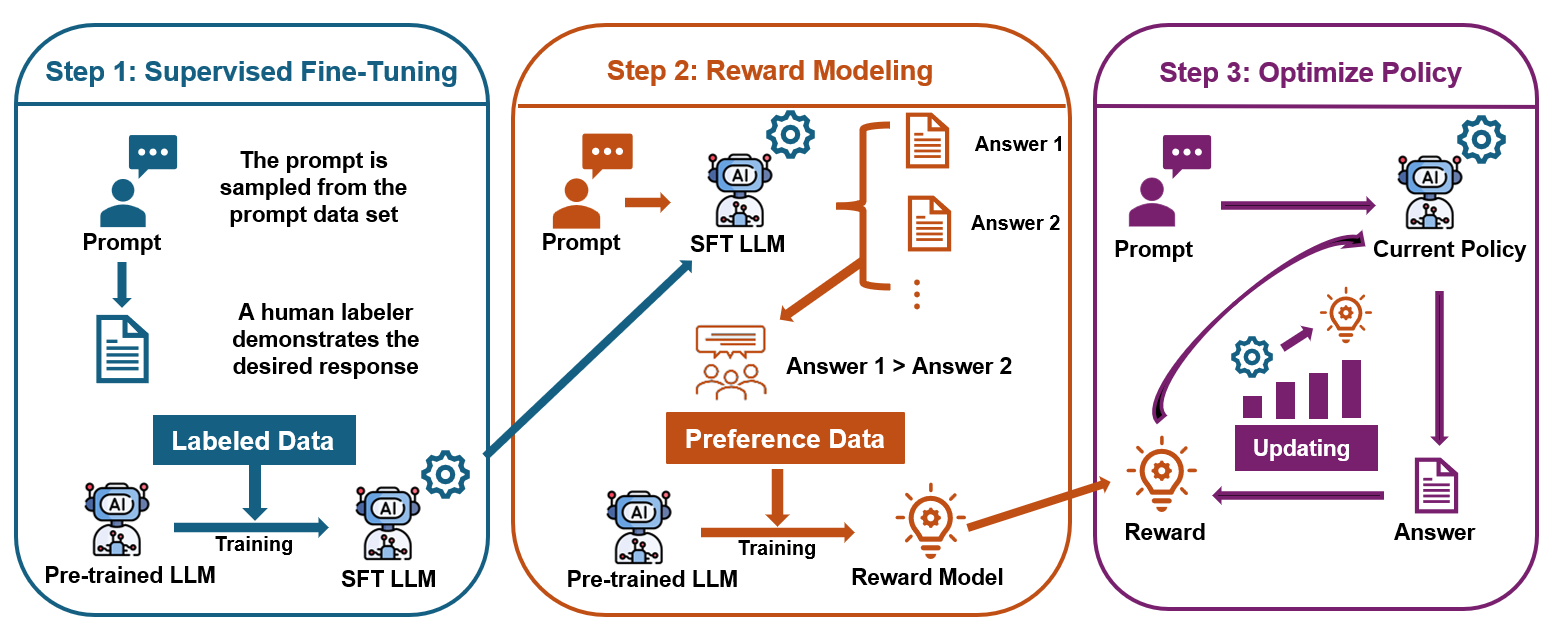}
    \caption{Integration of SFT (Step 1) and RLHF (Step 2 and Step 3).}
    \label{fig:integration2}
\end{figure}

Given their distinct learning objectives and optimization paradigms, SFT and RLHF are suited to different scenarios. SFT is typically preferred when the task is well-defined, static, and when sufficient labeled data is available \citep{ghosh2024closer, wang-etal-2023-self-instruct}. On the other hand, RLHF is more effective for aligning models with complex, subjective, or evolving user expectations, using interactive human feedback \citep{winata2025preference, ouyang2022training}.
While these methods have distinct strengths, recent developments suggest that integrating SFT and RLHF into a unified training pipeline can yield more robust and aligned language models \citep{ouyang2022training,openai2023gpt,bai2022training,team2024gemma,touvron2023llama,guo2025deepseek}. The following sections discuss this integration from three perspectives: the opportunity that enables the integration pipeline, its application in leading LLMs, and the new challenges it introduced.

\textbf{\textit{Opportunity.}}
Recent research suggests that SFT on a moderately sized, high-quality dataset can already yield outputs comparable to those produced by human annotators \citep{touvron2023llama}. Thus, A widely adopted integration strategy begins with supervised fine-tuning on high-quality demonstrations to teach the model basic instruction-following behavior. After this stage, the fine-tuned model is used to generate multiple candidate responses to various prompts. These responses are then compared by human annotators to produce preference data, which serve as training targets for a reward model. Finally, reinforcement learning is applied to further fine-tune the model using this reward model, enabling it to produce outputs that better align with nuanced human preferences (as shown in Figure~\ref{fig:integration2}).

\textbf{\textit{Applications.}}
This hybrid approach is now common among leading language models. For instance, InstructGPT \citep{ouyang2022training} pioneered the two-stage pipeline, combining SFT and RLHF. Similar methodologies have been adopted by OpenAI’s GPT series \citep{openai2023gpt}, Anthropic’s Claude models \citep{bai2022training}, Google’s Gemini \citep{team2023gemini} and Gemma series \citep{team2024gemma}, Meta’s LLaMA \citep{touvron2023llama}, and DeepSeek models. Notably, DeepSeek-R1-Zero is trained purely via RLHF, while DeepSeek-R1 adds an SFT stage before RLHF. The latter achieves substantially better reasoning performance, illustrating the benefits of combining the two strategies \citep{guo2025deepseek}.

\textbf{\textit{New Challenges.}}
Despite its growing popularity, the integration of SFT and RLHF introduces additional challenges. Their differing optimization goals, cross-entropy minimization in SFT versus reward maximization in RLHF, which may lead to conflicting gradient updates and unstable training dynamics \citep{hua2024intuitive}. Recent studies aim to mitigate the conflict and streamline the training pipeline. For instance, Intuitive Fine-Tuning (IFT) integrates SFT and RLHF into a single stage training process by introducing a temporal residual connection between supervision signals and reward-based updates \citep{hua2024intuitive}. Similarly, Unified Fine-Tuning (UFT) reformulates both objectives under a generalized implicit reward function, allowing simultaneous optimization with a single loss function \citep{wang2024uft}. These approaches represent promising directions for balancing alignment, performance, and training efficiency in the integration of SFT and RLHF.

\section{Advanced Alignment Techniques and Recent Innovations}
In addition to traditional SFT and RLHF, this section explores several advanced alignment techniques and highlights their advantages over traditional methods. Reward-free methods remove the need for manually designed reward signals, streamlining the training process and reducing reliance on human input. AI assistant alignment uses a stronger, pre-aligned model to generate high-quality guidance, which accelerates convergence and enhances safety. Self-alignment lets a model critique and improve its own outputs, yielding more robust behavior without additional data collection. Multi-agent deliberation strategies enable specialized agents to work together on complex tasks, increasing reliability and overall performance. Multi-objective alignment balances competing goals, including accuracy, fairness and computational efficiency, within a unified framework to produce more versatile and trustworthy models. Combined, these approaches lower development cost, scale more easily to large models, and offer more effective ways to align LLMs with human values.

\subsection{Direct Preference Optimization and Reward-Free Methods} \label{sec:71} 
As the scale of LLMs increases, ensuring alignment with human intent becomes increasingly critical. Conventional approaches such as Reinforcement Learning from Human Feedback (RLHF) \citep{christiano2017deep} have been widely adopted to bridge this gap. However, RLHF pipelines are often resource-intensive and require an additional reward model trained to predict human preferences. This dependency introduces potential sources of error and makes the alignment process less interpretable and harder to scale.

Recent innovations aim to simplify this process by eliminating the need for an explicit reward model altogether. A prominent example of this shift is Direct Preference Optimization (DPO), introduced by \citet{rafailov2023direct}. DPO formulates alignment as a supervised contrastive learning problem, where the model is directly trained to prefer outputs labeled as better by human annotators over those labeled as worse. Instead of first training a reward model and then applying reinforcement learning (e.g., PPO), DPO uses a probabilistic objective derived from a MLE principle over pairwise preferences. This method significantly reduces computational complexity while maintaining and in some cases improving the alignment quality compared to RLHF. Furthermore, mathematically, DPO maximizes the likelihood ratio between a preferred and a dispreferred response, which implicitly shapes the model’s behavior to mirror human judgment. The simplicity of the DPO objective also improves training stability and interpretability. Theoretically, DPO can be interpreted as recovering the policy optimal under an unknown reward, bypassing reward regression altogether.

In addition to DPO, other reward-free or preference-based methods have emerged that focus solely on relative preferences rather than scalar rewards. For instance, pairwise ranking models use binary comparisons between responses to fine-tune the model, circumventing the instability of reward estimation \citep{wu2023fine}. These models operate under the insight that humans often find it easier to choose between two options than to provide a numerical rating, making preference data easier to collect and less noisy.


These innovations collectively represent a transition toward simpler, more scalable, and interpretable alignment techniques, offering strong empirical performance while reducing dependency on reinforcement learning infrastructure. They enable alignment pipelines that are not only more resource-efficient but also better suited for iterative deployment in real-world applications where feedback is noisy, sparse, or difficult to quantify.

Overall, these reward-free alignment techniques significantly streamline the training pipeline by eliminating the need for explicit reward modeling.

\subsection{AI-Assistant Alignment and Self-Alignment} \label{sec:72} 

The rapid progress of large language models has created an urgent need for alignment methods that scale beyond intensive human supervision. AI-assistant alignment, together with its self-alignment variant, seeks to automate and streamline this process while safeguarding both ethical considerations and practical performance.

\subsubsection{AI-Assistant Alignment}
\begin{figure}[t]
    \centering
    \includegraphics[width=0.85\linewidth]{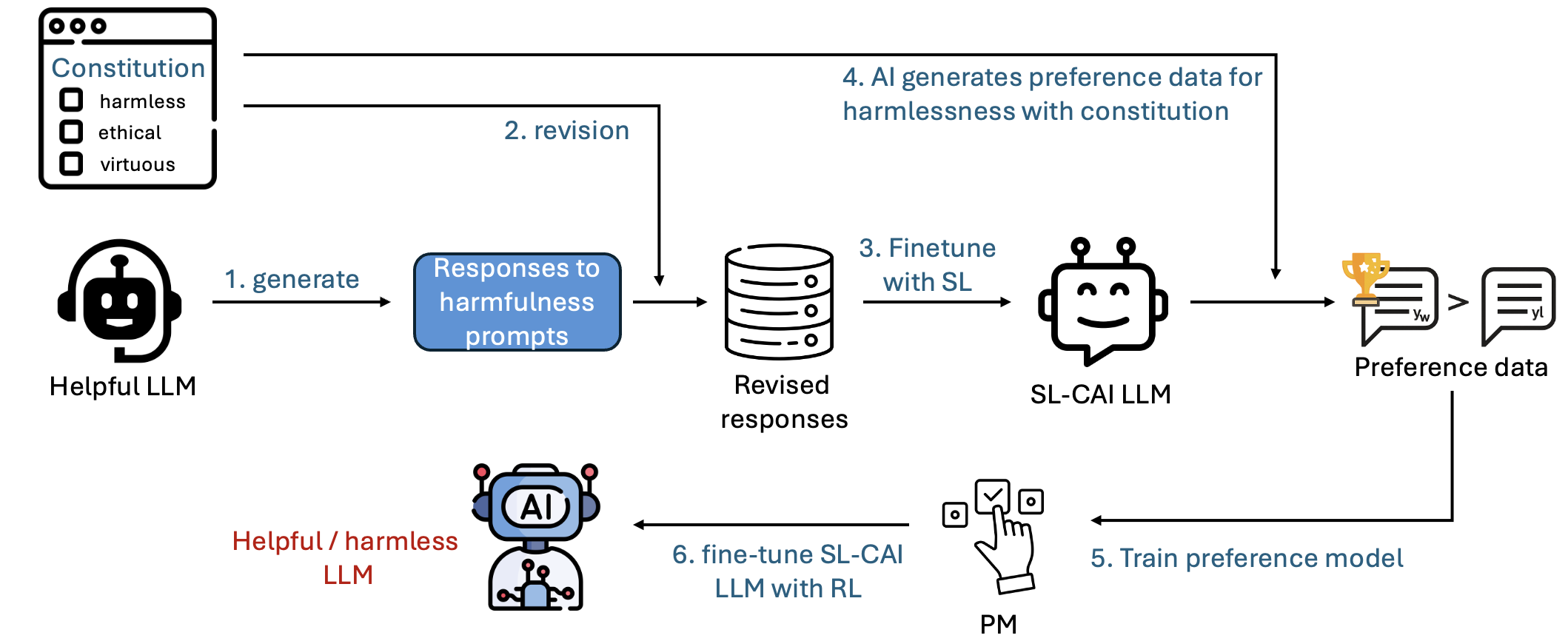}
    \caption{Workflow of the Constitutional AI (CAI) Framework. A helpful LLM first generates responses to harmfulness prompts, which are revised based on a predefined constitution encoding principles like harmlessness, ethics, and virtue. The revised responses are used to fine-tune the model via supervised learning, producing an initial SL-CAI model. Preference data is then generated by comparing model outputs under the constitution, and a preference model is trained accordingly. Finally, reinforcement learning is applied using the preference model to further align the SL-CAI model, yielding a more helpful and harmless LLM.}
    \label{fig:CAI}
\end{figure}

Recent work on aligning LLMs has progressed from RLHF to approaches that minimize or eliminate the need for costly human annotation. In the standard RLHF pipeline, an instruction-tuned model is coupled with a reward model trained on human-ranked answer pairs and then fine-tuned with a policy-gradient method that retains a Kullback–Leibler constraint to stay close to the original distribution \citep{ziegler2019fine,ouyang2022training}.
While effective, RLHF is limited by the time and expense required to collect expert preferences. Reinforcement Learning from AI Feedback (RLAIF) follows the same three stages (data collection, reward-model fitting, and reinforcement learning) but replaces human comparisons with judgments produced by a stronger, already aligned teacher model, enabling the generation of millions of preference pairs at negligible marginal cost. 
Constitutional AI (CAI) introduced by \citet{bai2022constitutional} is the most influential example of RLAIF. CAI begins with a short, human-written constitution that encodes safety and helpfulness rules. A helpful model, already aligned through RLHF, is first supervised to critique and revise its own answers according to these principles, producing what the authors call the supervised constitutional model. In the next phase, this same model generates two candidate answers for each prompt and, when prompted with the constitution, decides which answer better follows the rules, thereby creating AI-labeled preference pairs without human raters. A separate and smaller reward model is trained on these pairs, combined with earlier helpfulness comparisons, and then frozen. Finally, the policy is fine-tuned to maximize the frozen reward while limiting divergence from its supervised starting point. This process removes per-example human labeling yet yields assistants that are rated safer and less evasive than those produced by standard RLHF, demonstrating the practical promise of RLAIF. The workflow is provided in Figure \ref{fig:CAI}.

A growing body of work expands RLAIF beyond the original Constitutional AI recipe.  
\citet{lee2023rlaif} presents the first systematic head-to-head comparison of human- and AI-sourced preferences, finding that AI feedback matches human feedback when the teacher model is sufficiently capable.  
\citet{cui2023ultrafeedback} scale this idea, constructing a million-example GPT-4 preference set and showing that policies trained only on this data rival strong RLHF baselines.  
To balance multiple objectives, \citet{li2024hrlaif} mixes helpfulness and harmlessness signals in a single reward, while \citet{li2025curriculum} introduces a difficulty-ordered curriculum that improves generalization in a fixed label budget.  
\citet{sharma2024critical} audit these pipelines and highlight systematic divergences between AI- and human-generated labels.  
RLAIF is also moving beyond text: \citet{ahn2024tuning} adapts the framework to video understanding, and \citet{jing2024fgaif} applies fine-grained AI feedback to vision–language models, reducing object-level hallucinations.

\subsubsection{Self-Alignment}

Several studies show that a model can supply its own feedback without an external teacher.  
\citet{bao2024aligning} demonstrates that a 13-billion-parameter chat model can critique and rank its own answers under a simple rubric, yielding performance comparable to teacher-based RLAIF.  
Similarly, \citet{yu2024self} reports that generating a brief self-critique before reward-model fitting boosts alignment quality and cuts the need for external labels by 80 percent.  
These self-alignment results suggest that, once a model passes a competence threshold, it can bootstrap further alignment with minimal additional supervision, although questions remain about error compounding and bias reinforcement.

\subsubsection{Challenges and Future Directions}
The emerging RLAIF family still faces open problems in bias control, robustness, and evaluation.  
Because AI-generated labels inherit the value structure of the teacher, systematic divergences from human judgment can persist or even amplify, as quantified by the large-scale audit of \citet{sharma2024critical}.  
Head-to-head studies \citep{lee2023rlaif} argue that teacher quality largely determines final alignment, so future work must develop teacher-agnostic debiasing techniques or lightweight human spot-checks to prevent value lock-in.  
Robustness to red-team attacks remains another priority: \citet{bai2022constitutional} proposes iterated ``online'' training, continuously refreshing the preference model with new AI feedback from the policy’s own failure modes, while \citet{li2025curriculum} shows that curriculum-style data ordering can harden models without extra labels.  
Self-alignment methods that recycle a model’s own critiques \citep{bao2024aligning,yu2024self} promise unlimited scalability, yet they raise questions about error compounding and whether periodic human ``re-grounding'' is required.  
Moving beyond text, multimodal extensions already surface new challenges: video-based RLAIF must judge temporal coherence \citep{ahn2024tuning}, and vision–language alignment needs object-level feedback to curb hallucinations \citep{jing2024fgaif}.  
Finally, static benchmarks saturate quickly; several papers call for adaptive, adversarial evaluation suites that track constitutional drift and reward hacking over multiple bootstrapping generations.  Addressing these challenges will likely require hybrid pipelines that blend a small amount of strategically targeted human input with scalable AI feedback, uncertainty-aware reward ensembles, and red-team-in-the-loop training protocols.

\subsection{Multi-Agent and Deliberative Alignment Approaches} 
Deliberative Alignment is an approach to make language models safer and more dependable. Instead of just learning from examples, it's about directly teaching the model the actual safety rules and then training it to consciously think through these rules before it gives an answer~\citep{guan2024deliberative}. The idea is to make sure the model sticks closely to safety guidelines. This method helps models get better at spotting and handling tricky situations, including attempts to make them say things they shouldn't. It also means they're less likely to refuse perfectly normal requests and can handle new or unusual scenarios more effectively~\citep{konya2023deliberative}.

The process basically involves a couple of main stages. First, the model is shown a lot of examples where it sees how to reason through safety rules to arrive at a good answer, this is like a focused study period. Then, it goes through a kind of practice phase using reinforcement learning, where it gets feedback to sharpen its decision-making skills, especially when dealing with prompts that touch on safety concerns. The safety rules themselves delineate the parameters of acceptable and unacceptable content, and stipulate the model's appropriate responses in diverse scenarios, including instances requiring the declination of a request or the provision of a meticulously formulated 'safe' answer. By integrating this rule-based reasoning directly into the model, Deliberative Alignment endeavors to establish a more transparent, trustworthy, and scalable methodology for ensuring responsible language model behavior~\citep{fang2025large}.

Self-consistency is another crucial method applied to LLMs for addressing prevalent issues such as deficient reasoning and the generation of hallucinations~\citep{wang2022self}. Referencing the survey by Liang et al., we can understand this through the broader lens of ``internal consistency" which pertains to the uniformity of an LLM's expressions across its latent, decoding, or response layers when subjected to sampling methodologies~\citep{liang2024internal}. Numerous studies prefixed with ``Self-" including prominent examples like Self-Consistency~\citep{li2024turning}, Self-Improve~\citep{patel2024large}, and Self-Refine~\citep{ranaldi2024self}, have emerged to tackle these challenges. These approaches, while sometimes distinct in their specific mechanisms, all fundamentally involve LLMs in a process of evaluating and subsequently updating their own outputs or internal states.

Building upon the foundational concept of Self-Consistency, which primarily leverages majority voting over multiple generated outputs, several nuanced variations have been developed to refine the selection of the optimal response. Among these, Multi-Perspective Self-Consistency ~\citep{huang2023enhancing}distinguishes itself by incorporating assessments from diverse criteria or viewpoints when evaluating generated candidates, moving beyond simple congruence of final answers. Universal Self-Consistency~\citep{chen2023universal} introduces a further layer of sophistication, often employing a language model to ascertain the semantic equivalence of varied expressions before a consensus mechanism is applied, thereby accommodating greater diversity in response phrasing. In a different vein, Soft Self-Consistency~\citep{wang2024soft} shifts the focus from discrete answer selection to a more probabilistic approach, typically by weighting different reasoning paths or outputs based on the model's internal confidence scores or token probabilities accumulated throughout the generation process. Each of these adaptations thus offers a distinct strategy for aggregating or filtering multiple reasoning instances, aiming to enhance the robustness and accuracy of the final output under various task constraints and response complexities.

These ``Self-" prefixed methods largely fall under a unified theoretical framework termed ``Self-Feedback"~\citep{liang2024internal}. This framework elegantly breaks down the process into two core modules: ``Self-Evaluation" and ``Self-Update." During Self-Evaluation, the LLM assesses its own generated content or internal processes to capture signals related to internal consistency. These signals can be scalar (like a confidence score), textual (like a critique), or even contrastive. Subsequently, the ``Self-Update" module leverages these captured signals to enhance either the model's immediate response or, in some cases, the model's parameters themselves. While each specific ``Self-" method might have slight variations in how it implements these two modules, they share this fundamental cyclical process of introspection and refinement.

The primary characteristic of this Self-Feedback approach is its reliance on the LLM's inherent capabilities to introspect and improve, aiming to bolster internal consistency~\citep{prasad2024self}. The overarching purpose is to mitigate reasoning errors and reduce hallucinations by ensuring the model's expressions are more coherent and stable across different layers and sampling instances. This makes such methods applicable to a wide array of scenarios, notably in enhancing the reliability of LLMs for complex reasoning tasks (often seen in question-answering) and improving the factual accuracy and faithfulness of outputs in open-ended generation tasks.


\subsection{Group Relative Policy Optimization} 

Group Relative Policy Optimization (GRPO) is a recently proposed reinforcement learning framework that addresses key challenges in aligning large language models (LLM) \citep{shao2402deepseekmath}. GRPO was developed to better accommodate preference-based feedback and comparison-driven reward modeling, with the aim of improving both training efficiency and the stability of learning signals.

Traditional reinforcement learning methods, such as probal policy optimization (PPO), often struggle in aligning LLM due to several factors. PPO relies on a critic network to estimate per-token values, which nearly doubles the memory and computational requirements. Furthermore, in most LLM alignment settings, the rewards are extremely sparse - usually available only at the end of a generated sequence - and the reward models are commonly trained by pairwise comparison of entire responses. This mismatch leads to unstable advantage estimation and inefficient learning.

GRPO addresses these limitations by eliminating the critic network and directly using group-based reward normalization. For each prompt, GRPO samples a group of $G$ candidate outputs, each evaluated by a reward model. The advantage of each output is calculated as the difference between its reward and the mean reward of the group, aligning the learning signal with the structure of comparison-based reward modeling. The formal training objective of GRPO, as presented in DeepSeekMath \citep{shao2402deepseekmath}, is given by:

\begin{equation}\label{eq:grpo_ds_math}
\begin{aligned}
J_{\mathrm{GRPO}}(\theta) &= \mathbb{E}_{q \sim P(Q),\, \{o_i\}_{i=1}^G \sim \pi_{\theta_{\mathrm{old}}}(O|q)} 
    \frac{1}{G} \sum_{i=1}^G \frac{1}{|o_i|} \sum_{t=1}^{|o_i|} \\&\Bigg\{\min \Bigg[
         \frac{\pi_\theta(o_{i,t}|q, o_{i,<t})}{\pi_{\theta_{\mathrm{old}}}(o_{i,t}|q, o_{i,<t})} \, \hat{A}_{i,t},{\mathrm{clip}}\Bigg(
         \frac{\pi_\theta(o_{i,t}|q, o_{i,<t})}{\pi_{\theta_{\mathrm{old}}}(o_{i,t}|q, o_{i,<t})} ,1 - \epsilon , 1+\epsilon  \Bigg)\hat{A}_{i,t}
    \Bigg]
    - \beta \mathbb{D}_{\mathrm{KL}}[ \pi_\theta \Vert \pi_{\mathrm{ref}} ]
\Bigg\}
\end{aligned}
\end{equation}

The detailed GRPO objective and algorithm can be found in~\citep{shao2402deepseekmath}. 
GRPO introduces two principal modifications over standard PPO: First, the KL divergence is separated from the reward and added explicitly as a regularization term in the objective, rather than being mixed into the reward signal. Second, the group-relative advantage $\hat{A}_{i,t}$ is computed differently; specifically, it is standardized within the sampled group as
\begin{equation}
\hat{A}_{i,t} = \frac{r_{i} - \operatorname{mean}(r_{1}, \ldots, r_{G})}{\operatorname{std}(r_{1}, \ldots, r_{G})},
\end{equation}
where $r_i$ is the reward assigned to output $o_i$, and $\operatorname{std}$ denotes the standard deviation across the group.

This group-based design offers several alignment-relevant benefits. First, it structurally matches the comparison-based reward models used for aligning human preferences, ensuring a more faithful learning signal. Second, removing the critic reduces both the computational costs and the instability that would otherwise arise when only final-sequence rewards are present. Group normalization also provides a dynamic baseline, reducing variance in policy updates. Importantly, GRPO imposes KL divergence as a distinct regularization term, rather than including it in the reward signal, thus avoiding the bias introduced by KL-as-reward schemes.

Compared to PPO, GRPO introduces two principal differences. Firstly, it adopts a rational-style reward scaling mechanism that amplifies the policy update for candidates that significantly outperform their group. Secondly, it enforces KL regularization as a separate penalty term, rather than incorporating it into the reward, thus preventing interference in advantage estimation \citep{vojnovic2025alignment}. These properties enable GRPO to leverage pairwise preference data more effectively and achieve improved model alignment with human-evaluated outputs.

\begingroup
\setlength{\tabcolsep}{2pt}
\renewcommand{\arraystretch}{3}
\begin{table*}[t]
\centering
\tiny
\caption{Comparison of post-GRPO RL methods for LLM reasoning, by category.}
\label{tab:post_grpo_methods}
\begin{tabular}{@{}p{2cm} p{2.2cm} p{1.5cm} p{3cm} c c p{3.5cm}@{}}
\toprule
\textbf{Category} & \textbf{Method} & \textbf{Model} & \textbf{Datasets} & \textbf{OS} & \textbf{Speed-up} & \textbf{Core Contribution} \\
\midrule

\multirow{3}{*}{\shortstack{Training Stability\\\& Efficiency}}
& \textbf{CPPO}~\citep{lin2025cppo}
& 1.5B, 7B
& \shortstack[l]{GSM8K; MATH;\\AMC 2023; AIME 2024}
& Yes
& \shortstack{8.3× (GSM8K),\\3.5× (MATH)}
& Completion pruning \\

& \textbf{DAPO}~\citep{yu2025dapo}
& 32B (Qwen2.5)
& AIME 2024
& Yes
& 2× faster
& Clip decoupling \& dynamic sampling \\

& \textbf{VAPO}~\citep{yuyue2025vapo}
& 32B (Qwen2.5)
& AIME 2024
& No
& Fast conv.
& Critic augmentation \\

\midrule

\multirow{3}{*}{\shortstack{Reward Signal\\Enhancement}}
& \textbf{GRPO-LEAD}~\citep{wang2025grpo}
& 7B, 14B
& AIME 2024/25
& Yes
& Faster conv.
& Length \& difficulty shaping \\

& \textbf{S-GRPO}~\citep{dai2025sgrpo}
& 7B–14B
& \shortstack[l]{GSM8K; AIME;\\AMC; MATH; GPQA}
& Yes
& 35–61\% fewer tokens
& Decaying exit rewards \\

& \textbf{Spectral PO}~\citep{chen2025spectral}
& 7B, 14B, 32B
& 10 benchmarks
& N/A
& –
& All-negative diversification \\

\midrule

\multirow{2}{*}{\shortstack{Algorithmic\\Mods}}
& \textbf{Dr.~GRPO}~\citep{chen2025understanding}
& 7B
& \shortstack[l]{AIME; AMC; MATH;\\ Minerva; Olympiad}
& Yes
& Token-efficient
& Unbiased advantage \\

& \textbf{SEED-GRPO}~\citep{chen2025seed}
& 7B, 14B
& \shortstack[l]{AIME; MATH;\\GSM8K}
& Yes
& –
& Entropy-weighted updates \\

\midrule

\multirow{2}{*}{\shortstack{Task \&\\Multimodal}}
& \textbf{Flow-GRPO}~\citep{liu2025flow}
& SD 3.5
& \shortstack[l]{GenEval;\\Text-in-image}
& Yes
& Fewer steps
& ODE→SDE sampling \\

& \textbf{StepGRPO}~\citep{zhang2025r1vl}
& 7B–14B
& \shortstack[l]{8 V-L\\benchmarks}
& Yes
& –
& Step-wise dense rewards \\

\bottomrule
\end{tabular}
\end{table*}
\endgroup

Although the original GRPO has demonstrated notable improvements in alignment and efficiency compared to PPO, ongoing research has produced a variety of GRPO variants aimed at addressing specific limitations and further enhancing its effectiveness. These subsequent works fall into several broad categories according to their main focus: (1) improving training stability and optimization efficiency, (2) enriching and diversifying reward signals, (3) advancing core algorithmic modifications, and (4) extending GRPO-based RL to new tasks and modalities. Table~\ref{tab:post_grpo_methods} provides a comparative summary of these methods. In the following, we briefly review representative advances within each category.

\paragraph{ Training Stability and Optimization Efficiency. }

Several approaches have focused on accelerating and stabilizing GRPO-based RL. CPPO~\citep{lin2025cppo} mitigates the inefficiency of group sampling in GRPO by pruning low-advantage completions and reallocating resources to additional prompts, resulting in a speedup of up to 8.3$\times$ on GSM8K and 3.5$\times$ on MATH benchmarks. DAPO~\citep{yu2025dapo} introduces decoupled policy clipping and dynamic sampling strategies, enabling robust and reproducible large-scale RL training for LLMs; it achieves state-of-the-art open-source results on AIME 2024 with full code and data release. VAPO~\citep{yuyue2025vapo} reintroduces a value critic, with bias mitigation for long sequences and sparse rewards, leading to both higher accuracy and reliable convergence, surpassing previous GRPO-based methods on challenging reasoning tasks.

\paragraph{Reward Signal Enhancement and Diversity. }

Another line of research seeks to overcome the sparsity and homogeneity of reward signals in standard GRPO. GRPO-LEAD~\citep{wang2025grpo} introduces length-dependent rewards, explicit penalties for incorrect solutions, and difficulty-aware weighting to encourage concise and robust mathematical reasoning. S-GRPO~\citep{dai2025sgrpo} proposes serial sampling and decaying exit rewards, incentivizing early correct answers and leading to both shorter and more accurate solutions. Spectral Policy Optimization~\citep{chen2025spectral} addresses the issue of all-negative groups, where no sampled completion is correct, by injecting AI-driven diversity into reward signals, breaking update symmetry and accelerating convergence.

\paragraph{Algorithmic Modifications. }

A third category directly revises the GRPO algorithm to address optimization bias or incorporate uncertainty. Dr. GRPO~\citep{chen2025understanding} removes length and variance normalization from the advantage computation, eliminating a verbosity bias and increasing token efficiency, while maintaining strong accuracy. SEED-GRPO~\citep{chen2025seed} introduces semantic entropy as a measure of model uncertainty for each prompt, scaling policy updates more conservatively for uncertain queries and more aggressively for confident ones, thus improving generalization and stability across benchmarks.

\paragraph{Task and Multimodal Generalization. }

Recent works have also extended GRPO-style RL beyond mathematical reasoning. Flow-GRPO~\citep{liu2025flow} adapts RL optimization for text-to-image generation via flow matching models, employing ODE-to-SDE conversion and denoising reduction to improve compositionality and visual text rendering in diffusion models. StepGRPO~\citep{zhang2025r1vl} expands GRPO to multimodal reasoning with dense, step-wise feedback, enhancing multi-hop visual-language inference and outperforming imitation learning baselines on a suite of benchmarks.

Together, these developments demonstrate the adaptability of the GRPO framework and the breadth of innovations it has inspired. Each category reflects ongoing efforts to balance training efficiency, reward informativeness, algorithmic robustness, and domain generalization, advancing the state-of-the-art in RL-based alignment for LLMs.

\section{Efficient Fine-Tuning Techniques for Alignment}  
\label{sec:fine_tuning_techniques}
Efficient Fine-Tuning methods that address the substantial computational and memory demands associated with full model fine-tuning \citep{ding2023parameter, xu2023parameter,han2024parameter}.
The primary goals include significant reductions in computational costs, accelerated training speeds, lower memory and storage requirements, and effective mitigation of catastrophic forgetting \citep{liu2022few, fu2023effectiveness}.
In this section, we discuss the Efficient Fine-Tuning methods for LLM alignment including partial parameter fine-tuning, low-rank adaptation, sparse fine-tuning, knowledge distillation, adapter-based fine-tuning, and prompt tuning.


\subsection{Full or Partial Parameters Fine-Tuning}
\label{subsec:full_parameter_finetuning} 

Full-parameter fine-tuning involves updating all weights of a pre-trained Large Language Model (LLM) on a task-specific dataset. While this approach typically achieves high task-specific performance, it is significantly resource-intensive, requiring substantial computational power and memory capacity, especially as model sizes grow larger (e.g., billions of parameters) \citep{devlin2018bert, radford2019language}. Additionally, full fine-tuning often risks overfitting, particularly with limited training data, diminishing the model's generalization capabilities \citep{howard2018universal, dodge2020fine}.

To mitigate the intensive resource requirements and risk of overfitting associated with full fine-tuning, partial parameter fine-tuning methods have emerged. These methods selectively update a subset of parameters, significantly reducing computational cost while retaining competitive performance \citep{han2024parameter, liu2022few}. The parameters chosen for updating can vary, including the final classifier layers, embedding layers, or specific layers within transformer blocks. A prevalent example is layer-wise fine-tuning, which selectively tunes layers critical for task-specific performance \citep{lee2019mixout, zhang2020revisiting}. This approach often reduces the memory footprint and accelerates training by minimizing the number of gradient updates required.

Despite their practicality, partial parameter fine-tuning methods introduce additional complexity in deciding which parameters to tune, necessitating heuristic or algorithmic methods to determine optimal subsets \citep{lee2019mixout}. Moreover, partial fine-tuning can result in suboptimal adaptation, particularly if crucial task-specific knowledge resides in layers that remain fixed \citep{han2024parameter}. These limitations motivate further efficient fine-tuning approaches, such as low-rank adaptation, sparse fine-tuning, and adapter-based methods, which systematically balance resource efficiency and adaptability.

\subsection{Low-Rank Adaptation (LoRA) }
\label{subsec:lora}

\begin{figure}[htb]
    \centering
    \includegraphics[width=0.8\linewidth]{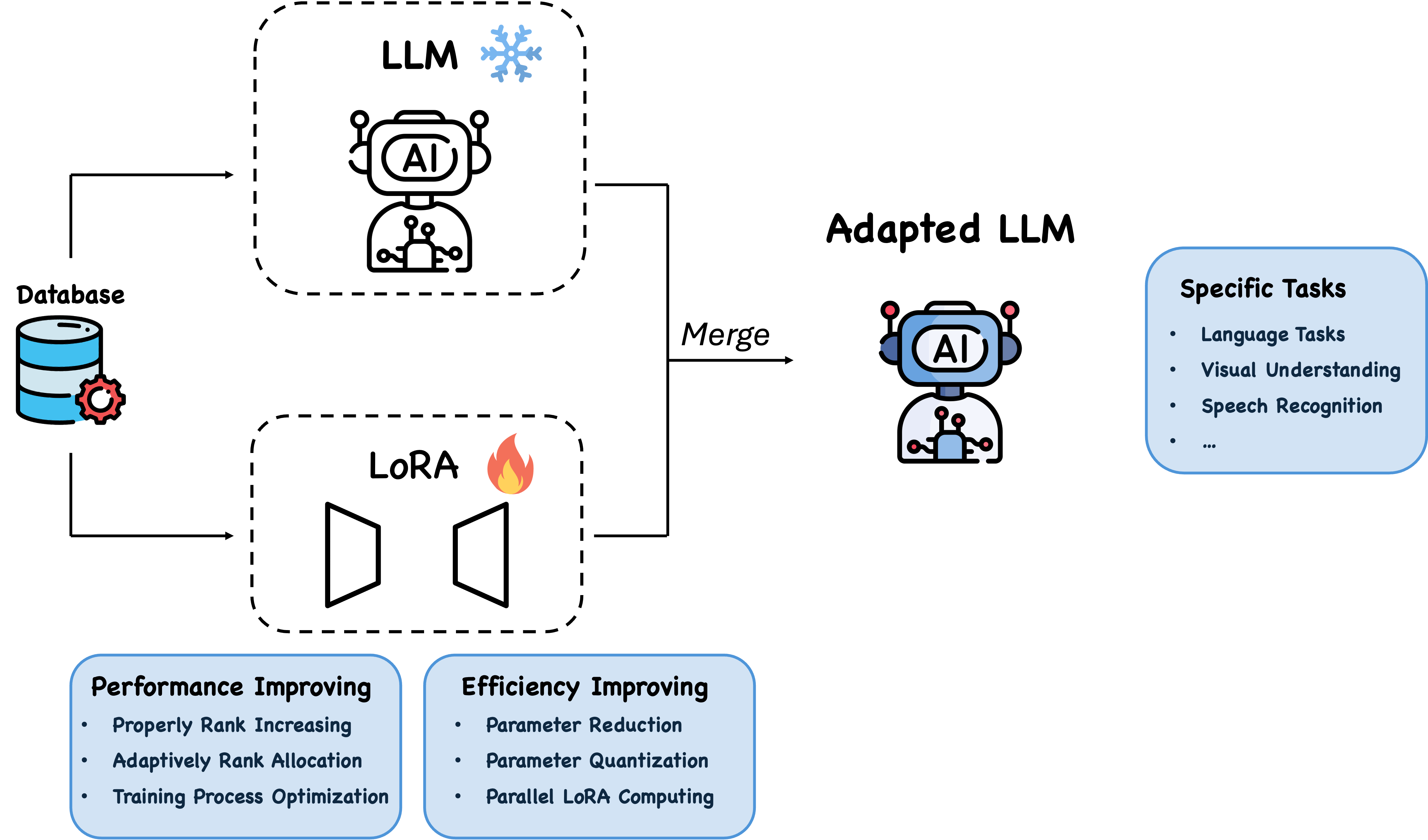}
    \caption{Overview of LoRA in LLMs. LoRA introduces trainable low-dimensional weight matrices that are integrated into frozen LLMs for fine-tuning. This PEFT approach enables LLMs to adapt effectively to specific tasks. Additionally, various techniques are employed to enhance both performance and efficiency, making the fine-tuning process practical and effective in real-world applications.}
    \label{lora82}
\end{figure}


LoRA introduces trainable low-rank matrices into each layer of the transformer architecture, allowing for efficient adaptation of LLMs with a reduced number of trainable parameters. This method significantly lowers the computational cost and memory footprint during fine-tuning, making it suitable for scenarios with limited resources. LoRA is adapted to practical challenges, and we will present the main methods that can improve performance and efficiency, building on the original LoRA framework introduced in \citep{hu2022lora}.

The core idea of LoRA is to freeze the pre-trained weight matrices and inject trainable low-rank matrices into dense layers. Instead of updating the full pre-trained weight matrix $W_0\in \mathbb{R}^{m\times n}$, LoRA learns a low-rank incremental update $\Delta W = BA$, where $B\in \mathbb{R}^{m\times r}$ (initialized as 0), $ A\in \mathbb{R}^{r\times n}$ (initialized as normal distributed random value), and the rank $r\ll min(m, n)$. The forward computation of LoRA can be expressed as below:
\begin{equation}
    h=W_0x+\Delta Wx=W_0x+BAx
\end{equation}

LoRA can achieve comparable performance on several downstream tasks, but there's still a performance gap between LoRA and full fine-tuning in areas such as mathematical reasoning and coding \citep{Mao_2024}. To bridge the gap, existing methods mainly focus on four perspectives: 
1) Increasing the rank $r$ appropriately\citep{lialin2023relorahighranktraininglowrank, xia2024chainloraefficientfinetuning}; 
2) Adaptively allocating ranks to LoRA modules across different layers \citep{zhang2023adaloraadaptivebudgetallocation, mao2024doraenhancingparameterefficientfinetuning, ding2023sparselowrankadaptationpretrained}; 
3) Optimizing the training process, including initialization improvement and gradient update optimization \citep{hayou2024impactinitializationlorafinetuning, hayou2024lora+}; 
4) Combining with other paradigms, such as Bayesian learning\citep{yang2024bayesianlowrankadaptationlarge}.

Accumulated linear computations introduced by LoRA modules in LLMs can still result in a non-negligible computation burden. To alleviate this, three main strategies have been proposed to make LoRA lighter and faster. 
1) Parameter reduction. This can be achieved through parameter freezing \citep{wu2024loraspstreamlinedpartialparameter}, module pruning \citep{zhou2024loradropefficientloraparameter} or parameter sharing \citep{kopiczko2024veravectorbasedrandommatrix}; 
2) Parameter quantization. By reducing the bit width of parameter, quantization-based method can significantlt lower memory usage and computational cost \citep{dettmers2023qloraefficientfinetuningquantized,li2023loftqlorafinetuningawarequantizationlarge};
3) Parallel in training and inference. This method leverages hardware related algorithm to accelerate the computation both in training and inference process\citep{ye2024mlorafinetuningloraadapters, chen2023punicamultitenantloraserving}.


\subsection{Sparse Fine-Tuning }\label{subsec:sparse} 

\begin{figure}[htb]
    \centering
    \includegraphics[width=0.4\linewidth]{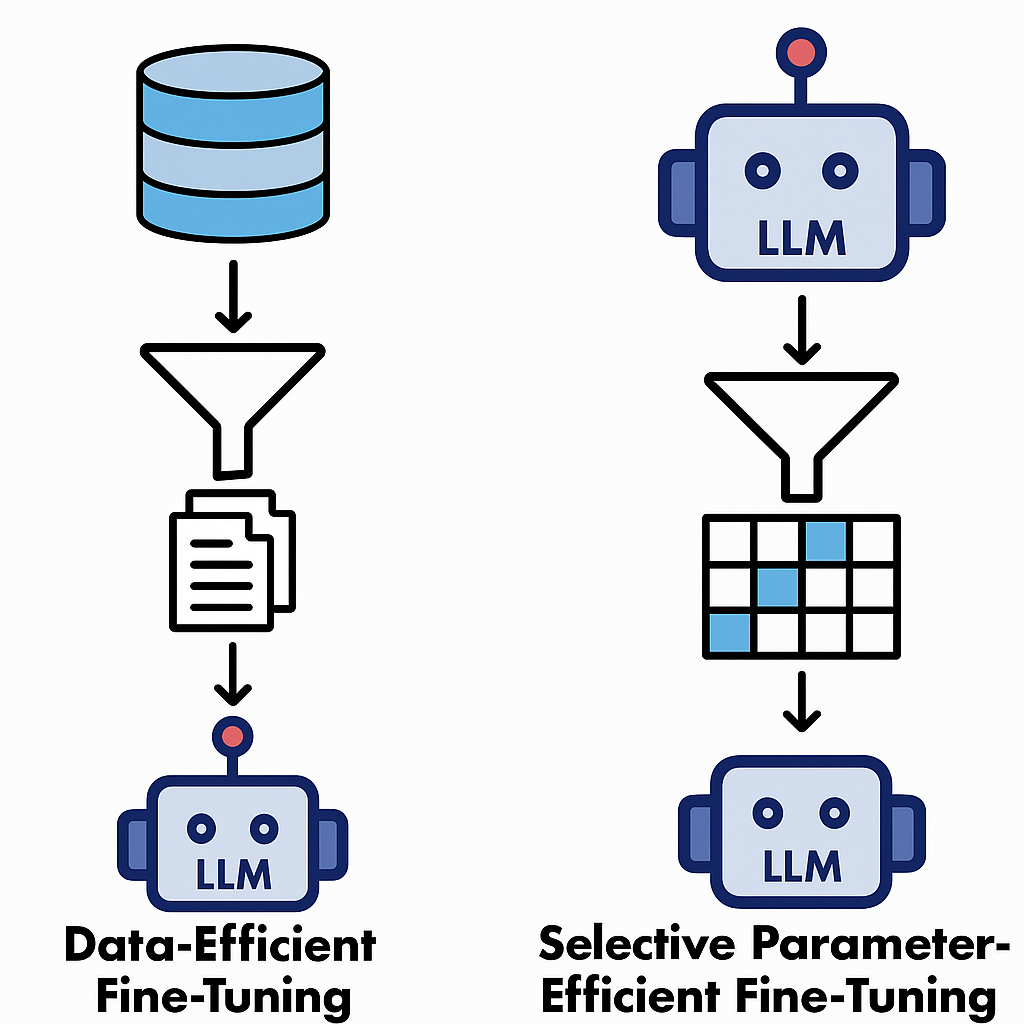}
   \caption{Two paths to sparsity: \textbf{(left)} \emph{data-efficient} fine-tuning filters the corpus before training, and \textbf{(right)} \emph{parameter-efficient} fine-tuning updates only a sparse mask of model weights.}

    \label{sparse_fine_tuning}
\end{figure}

We categorized sparse fine-tuning based on the aspects of the sparsity: (1) data-efficient fine-tuning that focuses on data-level sparsity by using informative subsets of data; and (2) selective parameter-efficient fine-tuning that focuses on parameter-level sparsity by updating only a critical subset of model weights.

Data-efficient fine-tuning aims to extract representative data points for substantially reducing the computational cost when a large scale of data is available for alignment \citep{zhou2023lima,wang2023far}. 
Therefore, the closely related research fields to this data-efficient fine-tuning include optimal subsampling \citep{ma2015statistical,wang2018optimal,ma2022asymptotic}, few-shot learning \citep{wang2020generalizing, liu2022few}, and coreset selection \citep{dasgupta2009sampling,albalak2024survey}.
Existing works of data-efficient fine-tuning can be classified into two categories: (1) non-informative sampling that selects samples based on predefined metrics on the data points; and (2) informative sampling that aims to minimize the empirical risk. 
The non-informative requires less knowledge about the downstream tasks and typically requires fewer computational resources for selecting samples. 
\citep{gao2020making} shows that the LLM can easily adapt to new tasks when fine-tuned on a small dataset drawn with uniform random sampling. 
\citep{bukharin2023data} proposes Quality-Diversity Instruction
Tuning (QDIT) to simultaneously control the sampled dataset diversity
and quality.
In contrast, informative sampling incorporates the model’s knowledge to calculate the influence of each data point and require more computational resources, e.g., analyzing the scale of gradients. \citep{xia2024less} proposes an optimizer-aware and practically efficient algorithm, Low-rank Gradient Similarity Search(LESS), to estimate data influences and perform instruction data selection for targeted instruction tuning in LLM.

Selective parameter-efficient fine-tuning leverages \emph{sparsity} by updating only a small fraction of a model’s parameters, thereby significantly cutting down the computational and memory costs of adapting large pre-trained models while maintaining near-original performance. Instead of tuning all weights, these methods identify a strategically chosen sparse subset of parameters to adjust based on measures of importance or sensitivity. For instance, SIFT (Sparse Increment Fine-Tuning) uses a gradient-based criterion: it exploits the observation that gradients in pre-trained models are extremely sparse (e.g., about 1\% of parameters account for 99\% of the total gradient norm) \citep{song2024sift}. SIFT therefore fine-tunes only the top-$x$\% of parameters with the largest gradient magnitudes, restricting updates to the most influential weights for the task. In contrast, PaFi adopts a task-agnostic approach by simply selecting the pre-trained weights with the smallest absolute magnitudes as the trainable subset, on the intuition that these low-magnitude parameters can be altered with minimal disruption to the model’s prior knowledge \citep{liao2023pafi}. Meanwhile, FishMask, a Fisher information-based masking method, computes an approximate Fisher information score for each parameter to gauge its importance, and then updates only the top-$k$ parameters deemed most critical for the target task \citep{sung2021fishmask}. The importance score is an approximation of the Fisher information matrix:
 \begin{equation}
\hat{F}_\theta=\frac{1}{N} \sum_{i=1}^N \mathbb{E}_{y \sim p_\theta\left(y \mid x_i\right)}\left(\nabla_\theta \log p_\theta\left(y \mid x_i\right)\right)^2,
\end{equation}
which is the average of the square gradient of $y$ with respect to a given parameter $\theta$.
 By confining fine-tuning to these important parameters (whether identified by gradient, weight magnitude, or Fisher information), such selective fine-tuning techniques drastically reduce training overhead and still achieve performance close to full model fine-tuning.


\subsection{Knowledge Distillation for Fine-Tuning} 
\label{subsec:knowledge_distillation}

Knowledge distillation (KD) compresses LLMs by transferring knowledge from a high-capacity teacher to a smaller student \citep{xu2024survey,fang2025knowledge}. Unlike direct SFT or RLHF pipelines, KD transfers alignment properties from a pre-aligned teacher model (typically optimized via RLHF or similar methods) to a student model by mimicking the teacher's performance, without repeating the costly preference-optimization loop.

The concept of KD is introduced in the doundational work of \citet{hinton2015kd}: instead of training on hard labels $y$, the student learns from the teacher's class probability distribution $\boldsymbol{p}_T = \sigma\bigl(\mathbf{z}_T / \tau\bigr)$, where $\mathbf{z}_T$ are logits from teacher, $\sigma$ is the softmax function, and $\tau$ is a temperature parameter that softens the distribution. The student's objective function is typically a weighted average of two losses: a standard cross-entropy loss $\mathcal{L}_{\mathrm{CE}}$ with the hard labels, and a distillation loss that minimizes the Kullback-Leibler (KL) divergence between the student's and teacher's softened outputs: 
\begin{equation}\label{eq: KD loss}
    \mathcal{L}_{\text {KD}} \;=\; \alpha \cdot \mathcal{L}_{\mathrm{CE}}\bigl(\sigma(\mathbf{z}_S(\boldsymbol{x})),\, y\bigr)
    \;+\; (1-\alpha) \cdot \tau^2 \cdot \mathcal{L}_{\mathrm{KL}}\bigl(\sigma(\boldsymbol{z}_T(\boldsymbol{x})/\tau),\, \sigma(\boldsymbol{z}_S(\boldsymbol{x})/\tau)\bigr),
\end{equation}
where $\alpha$ balances the two terms.
Matching the full probability vector rather than one-hot targets forces the student to copy subtle preference signals, including refusal styles, politeness markers, and content filters, that are otherwise hard to encode.

One primary advantage of KD is its efficiency. Compressing a 70-billion-parameter RLHF teacher to a 7-billion-parameter student cuts inference cost by roughly an order of magnitude while preserving core reasoning ability and alignment quality \citep{taori2023stanford,touvron2023llama,xu2024survey,guo2025deepseek,yang2024distillseq,ma2024knowledge,wang2025comprehensive}. Furthermore, the teacher is used only for forward passes, the same recipe scales easily to specialised domains: clinicians and bioinformaticians have distilled general-purpose aligned teachers into compact models fine-tuned for medical or scientific tasks without eroding their safety guarantees \citep{niu2024clinragen,tariq2024radiology,ge2025clinkd,latif2024knowledge,shang2024accurate}.

\begin{figure}[t]
\centering
\includegraphics[width=0.8\linewidth]{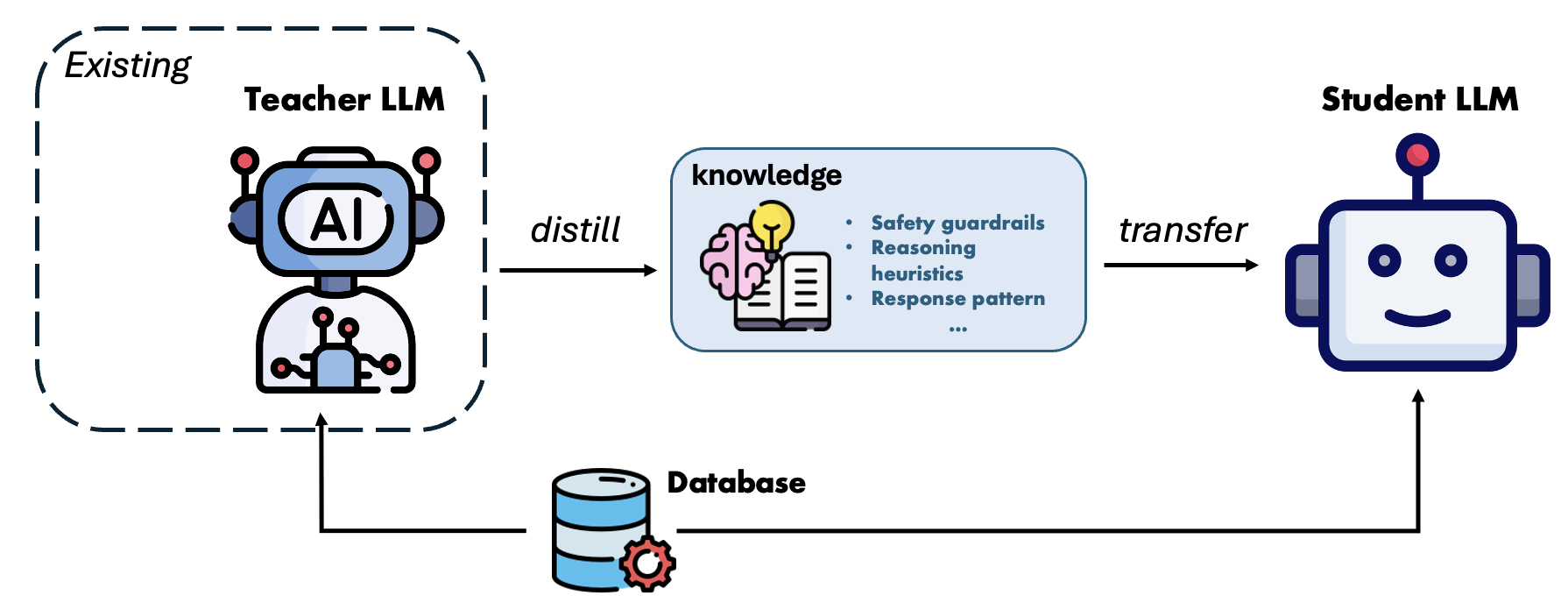}
\caption{Overview of Knowledge Distillation in LLMs. Knowledge is distilled from a teacher LLM, which is typically optimized via RLHF. This knowledge, potentially enriched with current, task-specific data, is transferred to a smaller student LLM. By learning from both the teacher’s guidance and the current data, the student LLM becomes more efficient and effective at performing downstream tasks.}
\label{fig:KD_frame_LLM}
\end{figure}

Subsequent work shows that transferring richer signals strengthens alignment further. Feature-based KD \citep{ji2021featurebasedkd}matches hidden activations so the student inherits syntactic and semantic structure. Attention distillation \citep{jiao2019tinybert} copies attention maps, helping small transformers remain stable during fine-tuning. Relation-based KD \citep{yang2022relationbasedkd} preserves similarity patterns inside the network and improves dense prediction tasks. Sequence-level KD trains the student to reproduce complete outputs, capturing dialogue coherence and style \citep{taori2023stanford,li2021mutual}. Relation-based KD distills the reasoning patterns of teachers, such as the Chain-of-Thought (CoT) reasoning, from teacher LLMs so the student learns intermediate reasoning steps rather than shortcutting to the final answer \citep{hsieh2023distilling,feng2024keypoint}. For example, \citet{hsieh2023distilling} generates the teacher rationales, and then the student is trained to jointly predict both the rationale and the final answers, helping the student learns intermediate reasoning steps rather than shortcutting to the final answer. Formally, the student is trained on a loss:
\begin{equation}
\mathcal{L} = -\frac{1}{n} \sum_{i=1}^n \log P_\theta(\boldsymbol{r}_i, y_i \mid \boldsymbol{x}_i),
\label{eq:rbkd_loss}
\end{equation}
where $(\boldsymbol{r}_i, y_i, \boldsymbol{x}_i)$ comprises rationale, input, answer, and \(P_\theta\) denotes the student model parameterized by \(\theta\). 

Several extensions refine KD for alignment objectives. Multi-teacher distillation aggregates outputs from diverse, pre-aligned models (e.g., RLHF-specialized variants), enabling students to synthesize compounded behavioral priors \citep{khanuja2021mergedistill,zhang2022confidence,liu2024wisdom,wadhwa2025taught}. Dynamic adaptive distillation further introduces a paradigm of bidirectional co-evolution, where teacher and student models undergo simultaneous joint optimization for continuous mutual refinement \citep{sun2021collaborative, chang-etal-2022-one,li2024bild}. Self-distillation represents a distinct approach that bypasses external teachers entirely, leveraging a single model’s self-generated outputs for iterative refinement\citep{Zhang2019BeYO,yang2023alphafold2}.

One another key point is the uncertainty-aware abilities of the student model. Classic KD can create over-confident students because it ignores uncertainty. Uncertainty-aware variants address this in two complementary ways. 
The first approach focuses on distilling uncertainty by training the student to mimic the teacher's full predictive distribution over labels. These distributions are typically learned from uncertainty-aware systems such as Bayesian neural networks or model ensembles, with the student minimizing divergence between its outputs and the teacher's probabilistic outputs \citep{korattikara2015bayesian,vadera2020generalized,malinin2019ensemble}.
The second approach addresses quantification of the student's intrinsic uncertainty by reinterpreting knowledge distillation through a Bayesian framework. Bayesian Knowledge Distillation (BKD) \citep{fangbayesian2024} achieves this by embedding the teacher's output probabilities as a teacher-informed prior distribution over the student model's weights. Within this formulation, standard knowledge distillation loss emerges naturally as the posterior optimization objective. Sampling techniques, such as stochastic gradient Langevin dynamics, applied to this posterior yield principled predictive intervals. 
Collectively, these uncertainty-aware methods suppress overconfident predictions on noisy or out-of-distribution inputs while providing computationally efficient alternatives to full Bayesian training. This characteristic makes them particularly valuable for developing compact while still being safety-aligned LLMs.

\subsection{Adapter-Based Fine-Tuning} 
\label{subsec:adapter_tuning}

Adapter-based fine-tuning involves inserting small trainable modules, known as adapters, between the layers of a pre-trained model. The original model weights remain frozen, and only the adapters are trained on downstream tasks. This approach allows for efficient multi-task learning and rapid adaptation to new tasks without retraining the entire model.

An adapter module commonly contains two linear layers to conduct down-projection that reduces the dimensionality of the input and up-projection that restores it. Between the two layers, nonlinearity layer is applied to the down-projected representation.
During the fine tuning process, trainable parameters $\Phi$ are optimized on given dataset $D$, with specified loss function $L$:
\begin{equation}
    \Phi^* \leftarrow argmin_{\Phi}\;L(D;\Phi)
\end{equation}
Adapters can be broadly categorized into series adapters and parallel types, depending on how they are integrated with the backbone \citep{hu2023llmadaptersadapterfamilyparameterefficient}. 

\begin{figure}[htb]
    \centering
    \includegraphics[width=0.8\linewidth]{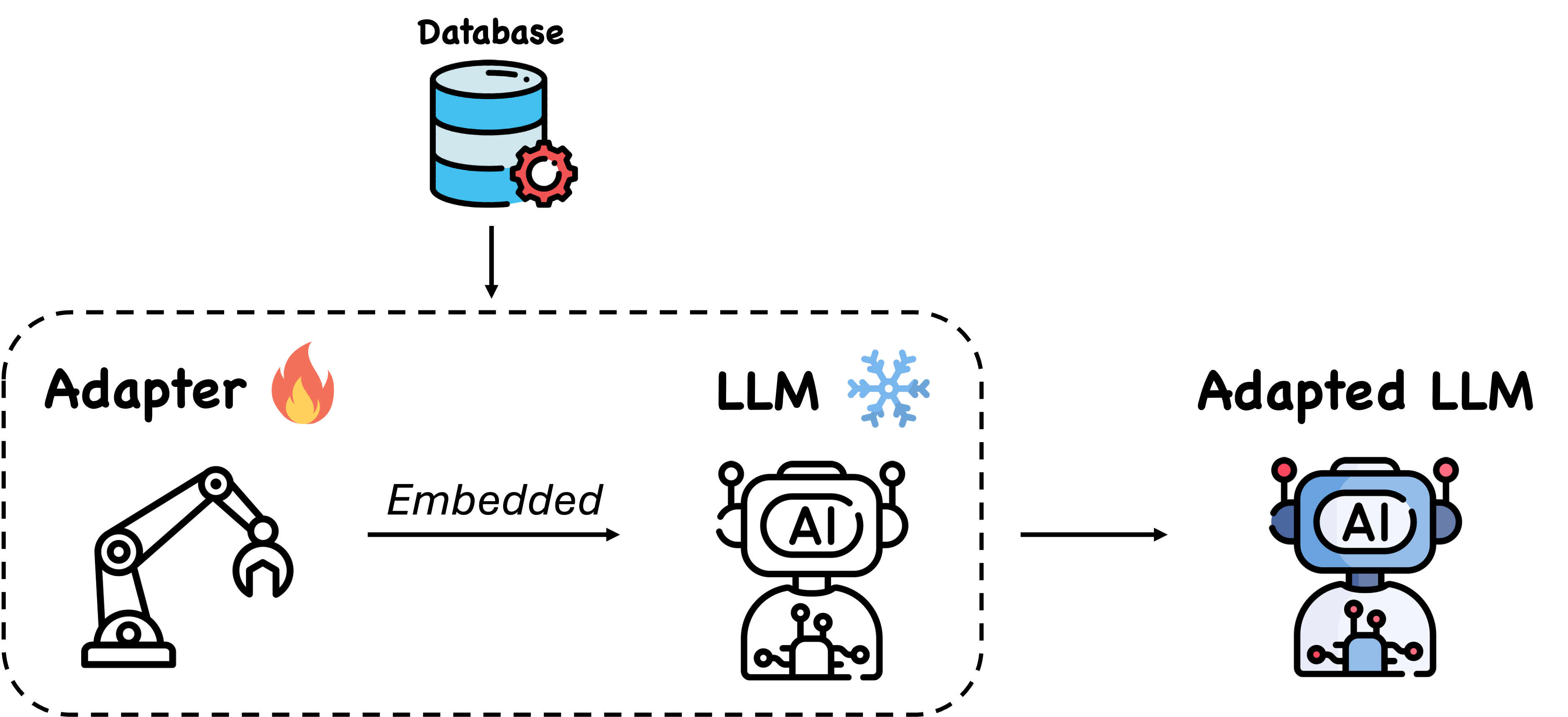}
    \caption{Overview of Adapter in LLMs. Adapters are lightweight, trainable modules inserted into the layers of a pre-trained, frozen LLM to enable adaptation to specific downstream tasks. By keeping the core model parameters fixed and training only the adapter components, this approach greatly reduces computational costs. Depending on how they are integrated into the LLMs, adapters are generally categorized as series and parallel types and numerous variants have been developed to enhance both performance and efficiency.}
    \label{adapter}
\end{figure}

Series adapters attach learnable modules sequentially within selected layers, typically within transformer layers. \citep{houlsby2019parameter} proposed the first series adapter as a parameter-efficient fine-tuning method by inserting trainable modules twice in each transformer layer. Adamix introduces multiple down- and up-projection layers in each adapter \citep{wang2022adamixmixtureofadaptationsparameterefficientmodel}. It employs stochastic routing to randomly select projection pairs during training. By maintaining the same number of tunable parameters and computational cost as the underlying PEFT, this technique outperforms these methods and even full fine-tuning in several datasets.
A structural limitation is that the backpropagation must still pass through the main backbone, resulting in additional computational steps and increased memory usage.
To further reduce training time and memory consumption, 1) adapter pruning and 2) sparse calculation techniques have been proposed. For instance, solely inserting the adapter module in each transformer layer \citep{pfeiffer2020madxadapterbasedframeworkmultitask} or dropping adapter in lower-level layers \citep{rücklé2021adapterdropefficiencyadapterstransformers} can still preserve performance. 
Compacter replaces standard linear layers with Kronecker products and shared parameters  \citep{mahabadi2021compacterefficientlowrankhypercomplex}. This technique maintains the sparsity and reduces computational complexity while achieving the performance comparable to original down- and up- projection calculations. 

Parallel adapters incorporate additional adapter modules in parallel with specific layers, which has similar manner as LoRA \citep{he2022unifiedviewparameterefficienttransfer}. 
When these adapters are connected in paralleled to the original layers, backpropagation can be applied directly across each adapter by linking them explicitly \citep{sung2022lstladdersidetuningparameter}. 
Multi-adapter is an extension on parallel adapter structure, which modifies the output of the self-attention heads through multiple parallel adapters. In multi-task settings, especially when tasks are sequentially related, single-task adapters often suffer performance degradation across tasks. AdapterFusion introduces a mechanism that parallels multiple task-specific adapters within the attention layer and fuses their outputs \citep{pfeiffer2021adapterfusionnondestructivetaskcomposition}. This approch achieves better performance than using a single-task adapter alone.

As a widely used PEFT technique, adapters can also be compatible with other PEFT methods \citep{mao2022unipeltunifiedframeworkparameterefficient}. This ability makes adapter have flexible choices for either the demand on inference speed or multi-task performance etc.



\subsection{Comparison of Fine-Tuning Techniques}
\label{subsec:comparison finetuning}

Each fine-tuning technique discussed in this section offers distinct advantages and limitations regarding computational efficiency, adaptability, and performance. Table \ref{tab:finetuning_comparison} provides a comprehensive summary, comparing critical aspects such as trainable parameters, computational and memory overhead, adaptability, and inherent trade-offs.

Full-parameter fine-tuning updates every parameter in a model, resulting in typically superior performance and high adaptability. However, it incurs substantial computational and memory demands, which become increasingly prohibitive with larger models. Partial parameter fine-tuning alleviates some of these burdens by selectively updating essential parameters, though it adds complexity in choosing optimal subsets and may risk suboptimal adaptation if crucial layers remain unchanged.

LoRA introduces small amount of trainable parameters (less than 1\%) by employing low-rank matrices, significantly reducing computational costs. While highly efficient, LoRA can fall short in tasks requiring deep reasoning or complex coding without further optimization. It balances well between resource efficiency and moderate performance.

Sparse fine-tuning effectively reduces training overhead by selectively updating parameters based on sparsity measures like gradient magnitudes or Fisher information. This approach typically achieves near full fine-tuning performance at significantly lower computational costs, provided the sparse masks are optimally determined. Nonetheless, it depends heavily on the quality and robustness of the sparsity strategy.

Knowledge distillation transfers knowledge from a larger teacher model to a smaller student model, considerably reducing inference costs. While computationally efficient during inference, this approach involves an initial cost for training the teacher model. Its performance relies significantly on the teacher’s quality and the distillation method, with possible performance degradation compared to the original large model.

Adapter-based fine-tuning achieves high adaptability and efficiency by inserting small, trainable adapter modules into pre-trained models. Although adapters facilitate efficient multitask learning and quick adaptation, they introduce computational overhead proportional to the complexity of the adapter modules, especially in series configurations.

Prompt tuning maintains the original model entirely frozen, adjusting only continuous prompt embeddings. This method offers minimal computational and memory overhead, suitable for resource-constrained environments. However, prompt tuning often provides limited adaptation capabilities compared to methods that directly update model parameters, constraining its effectiveness on tasks significantly divergent from the original pre-training objectives.

\begin{table}[ht]
\centering
\caption{Comprehensive Comparison of Efficient Fine-Tuning Techniques.}
\label{tab:finetuning_comparison}
\resizebox{\textwidth}{!}{%
\begin{tabular}{lcccc}
\toprule
\textbf{Method} & \textbf{Trainable Parameters} & \textbf{Memory Overhead} & \textbf{Compute Cost} & \textbf{Adaptability} \\
\midrule
Full-Parameter FT & High & High & High & High \\
Partial Parameter FT & Medium & Medium & Medium & Medium \\
LoRA & Low & Low & Low & Medium \\
Sparse FT & Low & Low & Low & High \\
Knowledge Distillation & Medium & Medium & Medium & Medium \\
Adapter-Based FT & Low & Medium & Medium & High \\
Prompt Tuning & Low & Low & Low & Medium \\
\bottomrule
\end{tabular}%
}
\end{table}

In summary, choosing an appropriate fine-tuning method depends on specific deployment scenarios and resource constraints, balancing between desired performance, computational resources, and model adaptability.

\section{Brain-Inspired LLM Alignments} 
\begin{figure}[ht!]
    \centering
    \includegraphics[width=\linewidth]{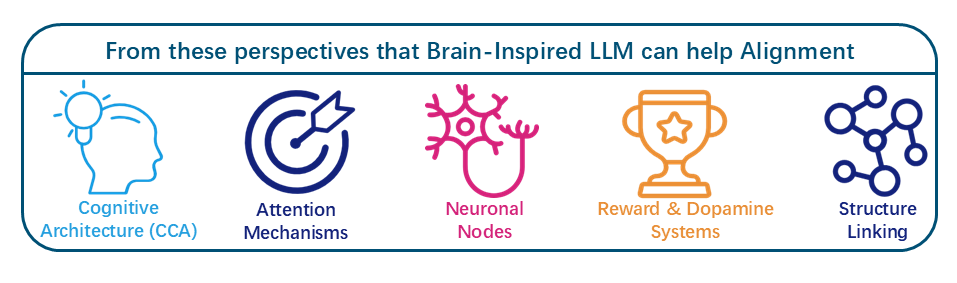}
    \caption{Different Perspectives that Brain-Inspired LLM can help Alignment.}
    \label{fig:integration}
\end{figure}
This section outlines the key principles and methodologies underlying brain-inspired LLM alignment, often referred to as Brain-AGI, highlights the existing challenges, and provides insights into possible opportunities.

\subsection{Recent advancements of Brain-Inspired LLM Alignments} 
Brain-inspired LLMs refers to language models whose architectures or training objectives draw directly on principles of human brain organization and function. Instead of treating neural networks as arbitrary black boxes, brain-inspired LLMs seek to mirror how our brains process language, represent concepts, and integrate sensory information \citep{Farisco2024}. For instance, the Causal Cognitive Architecture (CCA) \citep{Schneider2024}, a framework that models neocortical minicolumns as millions of ``navigation maps.'' These maps undergo continuous cognitive cycles in which sensory inputs are normalized, spatially and temporally bound into local maps, and then matched against stored multisensory maps to form a Working Navigation Map. The CCA framework demonstrates how lightweight, evolution-inspired modifications to a core navigation-map framework can spontaneously produce foundational aspects similar to that of human intelligence \citep{Schneider2024}.

Additionally, the BriLLM architecture \citep{zhao2025brillmbraininspiredlargelanguage} replaces the Transformer’s \citep{vaswani2017attention} attention blocks with a directed graph of ``neuronal'' nodes and energy-based signal flow, analogous to how biological neurons propagate activation along pathways of least resistance. In BriLLM, tokens are treated as nodes, and each edge carries a learnable ``energy tensor'' that determines which node (token) activates next. This design makes every internal connection interpretable and, in principle, allows unbounded context lengths \citep{zhao2025brillmbraininspiredlargelanguage}. 

Furthermore, \citet{sun2024brainlikefunctionalorganizationlarge} investigated whether LLMs exhibit a brain-like functional organization by directly linking sub-groups of artificial neurons (ANs) in models such as BERT and the Llama family to well-established human functional brain networks (FBNs). They extracted representative temporal ``atoms'' of neuron activity and used them to predict fMRI responses, demonstrating that these LLMs indeed form a modular, brain-like architecture. Comparing four models, BERT and three successive Llama variants, they found that this functional specialization strengthens with model sophistication: larger models yield brain maps that are both more consistent (showing reduced variability in engaged FBNs) and more compact (fewer, more specialized atoms per network), and their neurons display increasingly hierarchical temporal and anatomical distributions \citep{sun2024brainlikefunctionalorganizationlarge}.

\subsection{Brain-AGI Co-working} 

The motivation for brain and artificial general intelligence (AGI) co-working arises from the limitations of conventional AI systems, which often struggle with rigid task boundaries, inefficient energy use, and limited contextual understanding \citep{Hanene2025}. In contrast, biological brains demonstrate remarkable capabilities in generalization, lifelong learning, and causal reasoning, qualities that inspire the design of AGI systems seeking to replicate such flexibility and intelligence \citep{Gabriel2020}. Brain-AGI collaboration is envisioned as a way to harness the complementary strengths of human cognition and machine computation \citep{Zhao2023}, enabling more effective responses to complex societal challenges such as healthcare delivery and climate modeling \citep{Lu2023}. This synergy leverages human creativity and judgment alongside AGI's capacity for scale and speed. A central driver of this co-working paradigm is the pursuit of ethical alignment, which ensures that AGI systems are not only powerful but also aligned with human values and intentions \citep{Conitzer2024}.

{\it Theoretical foundations.} The co-working of the brain and AGI is underpinned by a set of complementary theoretical principles drawn from neuroscience and cognitive science \citep{Yu2024}. First, predictive coding provides a computational framework for modeling shared brain-AGI processing, wherein both biological and artificial agents minimize prediction error via hierarchical Bayesian inference \citep{Friston2020}. Second, neuroplasticity offers a foundation for co-adaptive learning systems, enabling AGI to dynamically update internal representations in response to environmental and social interactions, which is similar to synaptic rewiring in the human brain \citep{Parisien2023}. Third, embodied cognition underscores the significance of sensorimotor grounding for both natural and artificial agents, positing that cognition emerges through real-world engagement. This is an essential principle for brain-AGI integration in collaborative robotics and assistive technologies \citep{Taniguchi2022}. These three pillars are further reinforced by neurosymbolic integration, a hybrid framework that aligns neural learning with symbolic reasoning, offering interpretable and generalizable architectures essential for effective and trustworthy human-AGI co-working \citep{Bhuyan2024}.

\textit{Current strategies.}
Recent progress in brain-inspired artificial intelligence has catalyzed a new wave of brain-AGI co-working paradigms, driven by four interconnected approaches \citep{Yu2024}. First, neuromorphic engineering develops hardware systems that emulate the structure and dynamics of biological neural circuits. Technologies like Intel’s Loihi chip leverage spiking neural networks to enable real-time, energy-efficient computation, closely mirroring the temporal coding strategies of the brain \citep{Yik2023}. Second, cognitive architectures such as CLARION \citep{Sun2006} integrate symbolic reasoning with subsymbolic learning (e.g., deep learning, reinforcement learning), supporting flexible, goal-directed behavior that aligns with human-like cognitive processes. Third, human-in-the-loop learning frameworks, including inverse reinforcement learning and preference modeling, bring human values and feedback into the AGI training loop, enabling systems to adapt to social and ethical contexts through cooperative learning \citep{Hadfield2017}. Finally, neuroadaptive interfaces enable real-time, bidirectional interaction between human neural activity and artificial agents. These systems dynamically decode brain signals to adapt AGI responses, fostering seamless collaboration between biological and artificial intelligences \citep{Mushahwar2023}.

\textit{Applications.} Brain-AGI co-working is unlocking transformative applications across multiple fields by combining human cognitive strengths with artificial general intelligence \citep{Zhao2023}. In healthcare, this synergy enables early diagnosis of neurodegenerative diseases like Alzheimer's disease through AGI-assisted interpretation of fMRI data, and supports real-time decision-making in surgery via brain-computer interfaces \citep{Plis2022}. In autonomous systems, brain-AGI collaboration empowers robots with human-like spatial reasoning and adaptability, enhancing performance in complex, unstructured environments such as disaster response scenarios \citep{Kriegman2022}. Climate science similarly benefits from co-working paradigms, where AGI tools informed by human-guided intuition and causal reasoning predict ecological tipping points, guiding proactive environmental policies \citep{Rolnick2022}. These applications exemplify how brain-AGI integration not only augments machine capabilities but also extends human potential in addressing critical global challenges.

\textit{Challenges.} The pursuit of brain-AGI co-working, where artificial general intelligence systems are designed to complement or integrate with human cognitive processes, faces profound challenges across technical, ethical, and interdisciplinary domains \citep{Zhao2023}. Technically, neuromorphic systems inspired by the brain remain constrained by scalability and energy efficiency \citep{Mohamadreza2024}, limiting their capacity to support the high computational demands of AGI \citep{Kurshan2024}. Ethically, the prospect of AGI systems operating alongside or within human cognitive environments raises concerns about value misalignment, loss of human agency, and the amplification of biases without adequate oversight mechanisms \citep{Everitt2018}. Most critically, the gap between neuroscience and AI research continues to hinder meaningful co-design. The incompatible frameworks, terminologies, and research priorities make it difficult to translate biological principles into actionable AI models or to embed AGI into cognitive contexts in a way that is both effective and safe \citep{Hassabis2017}. Overcoming these barriers will require not just technical innovation but also deep, sustained integration across disciplines and a commitment to embedding human values at the heart of AGI design.

\subsection{Challenges and Limitations of Brain-Inspired LLM Alignments} 
Despite tremendous achievements in the brain-inspired LLM architectures such as enhanced interpretability, modular signal paths, and the promise of unbounded context lengths, several challenges remain. First, the computational overhead of maintaining and updating dense ``energy tensors'' across every connection in a graph-based design can be substantial, potentially eroding the very energy-efficiency gains these models aim to deliver \citep{zhao2025brillmbraininspiredlargelanguage}. 

Second, current alignment methods such as DPO align an LLM by directly maximizing the likelihood of human-preferred completions. DPO accomplishes this by reparameterizing reward as the log-ratio between the policy and a fixed reference model, thus eliminating the need for a separate reward network \citep{xu2024dposuperiorppollm}. Although DPO often matches or outperforms RLHF on in-distribution benchmarks, it fundamentally learns only the surface-level patterns of human preferences rather than the deeper cognitive processes \citep{tennant2025moralalignmentllmagents}. As a result, DPO-tuned models can ``sound'' convincingly human on familiar prompts but remain vulnerable to reward hacking and poor generalization when faced with novel or out-of-distribution scenarios.

Third, integrating ethical and societal considerations, such as bias mitigation or transparent governance, into a highly specialized, brain-inspired AGI framework remains an open problem, as standard alignment tools (RLHF, DPO) do not yet translate directly onto these new architectures \citep{Farisco2024}.  Finally, scaling this graph-based signal-flow mechanism to models with billions of parameters and multimodal inputs has not yet been demonstrated at production scale, raising questions about its practicality for real-world applications.

\subsection{Opportunities and Future Developments} 
Brain‐inspired LLM alignment techniques offer several promising advantages over conventional preference‐fitting approaches. 
By explicitly tying language-processing pathways to analogues of human cognitive systems, equips us with the ability to pinpoint exactly where and why a model’s reasoning diverges from our expectations. This transparency doesn’t just aid debugging; it lays the groundwork for truly collaborative workflows in which domain experts can inspect, validate, and open pathways toward achieving Brain-AGI. 

Furthermore, aligning LLM subnetworks with well-characterized human functional brain networks, such as the ``language'' network for factual consistency or the ``default-mode'' network for social reasoning, provides a neuroscientific scaffold for targeted auditing and control \citep{sun2024brainlikefunctionalorganizationlarge}. This modular interpretability makes it possible to apply fine‐grained alignment objectives, such as moral rewards derived from deontological or utilitarian principles, to dedicated subnetworks, reducing collateral effects on other competencies and mitigating reward‐hacking risks seen in methods like DPO. Lastly, grounding alignment in multimodal, embodied representations, by integrating vision, audio, and proprioceptive signals into the same brain‐inspired modules, promises richer, context‐sensitive behavior that better mirrors human cognition and values \citep{Farisco2024}.

Looking forward, next‐generation brain‐inspired alignment will likely embrace dynamic plasticity mechanisms, enabling models to adapt in real time through localized synaptic‐style updates rather than wholesale retraining. Instead of relying on static modules, future systems may incorporate continuous update rules that refine internal representations in response to live user feedback or environmental signals, much like how our brains learn from experience. Coupling such plasticity with lightweight neurofeedback, whether from biosignals or behavioral proxies, could give rise to true brain-in-the-loop pipelines that preserve robust alignment across shifting contexts. Finally, as Brain-AGI systems grow more autonomous, evolving multidisciplinary governance frameworks that integrate insights from neuroscience, ethics, and policy will be essential to ensure that alignment innovations remain transparent, accountable, and aligned with societal norms.

\section{Alignment Uncertainty Quantification (AUQ)} 

Alignment Uncertainty Quantification (AUQ) addresses a fundamental challenge in LLM development: how to measure and manage the uncertainty inherent in aligning models with human values and intentions \citep{gabriel2020artificial, bommasani2021opportunities}. As models become more powerful, understanding the reliability of their alignment becomes critical for safe deployment \citep{reynolds2021prompt}. Traditional machine learning uncertainty focuses primarily on predictive accuracy, while alignment uncertainty concerns whether a model's behavior truly reflects intended human values across diverse contexts and edge cases. The theoretical foundations of alignment uncertainty, methodologies for its quantification, and approaches for building systems that remain robustly aligned despite inevitable uncertainties are examined in this section.

\subsection{Sources of Alignment Uncertainty} 
The uncertainty in aligning LLMs with human values arises from multiple interrelated sources, complicating both measurement and mitigation. These can be grouped into three main categories: model-related uncertainty, feedback-related uncertainty, and contextual or distributional uncertainty.

\paragraph{1. Training Stability and Optimization Efficiency}
Model-inherent uncertainty stems from the stochastic nature of the training process and the architectural limitations of current LLMs. Due to random initialization and the use of stochastic gradient descent, models trained on the same data may still diverge in behavior and alignment properties \citep{gal2016dropout}. Additionally, present-day neural architectures may lack the expressiveness to fully capture complex or evolving human values, leading to representational misalignment \citep{dodge2020fine}. As models scale, \textbf{emergent behaviors} arise unpredictably, further complicating efforts to anticipate how alignment generalizes across model sizes and tasks \citep{wei2022emergent, ganguli2022predictability}.

\paragraph{2. Human Feedback Variability and Value Pluralism}
The human signals that guide alignment introduce substantial uncertainty. Annotators often provide inconsistent judgments for similar prompts, with inter-annotator agreement typically ranging from 0.6 to 0.8 Krippendorff’s alpha \citep{stiennon2020learning, ziegler2019fine}, reflecting noisy or subjective supervision. Moreover, human values are inherently diverse, shaped by culture, individual background, and societal context, making it difficult to define a universally valid alignment target \citep{gabriel2020artificial}. The mode of feedback also matters: scalar ratings, pairwise comparisons, natural language critiques, and demonstrations capture different aspects of preference and intent, introducing variation in the resulting aligned policy \citep{ouyang2022training, bai2022training}.

\paragraph{3. Context Sensitivity and Distributional Shift}
Uncertainty also emerges from real-world deployment contexts. Distribution shift, when deployed models face inputs or environments not reflected in the training data, can lead to severe misalignment \citep{hendrycks2019benchmarking}. Even within familiar domains, context-sensitive interpretation can vary dramatically depending on user identity, culture, or timing. The same response may be perceived as helpful or harmful depending on situational nuances \citep{perez2022discovering}. In addition, societal norms evolve over time, meaning that a model aligned with contemporary values may become increasingly misaligned as social standards shift \citep{solaiman2021process}.

\begin{figure}
    \centering
    \includegraphics[width=0.7\linewidth]{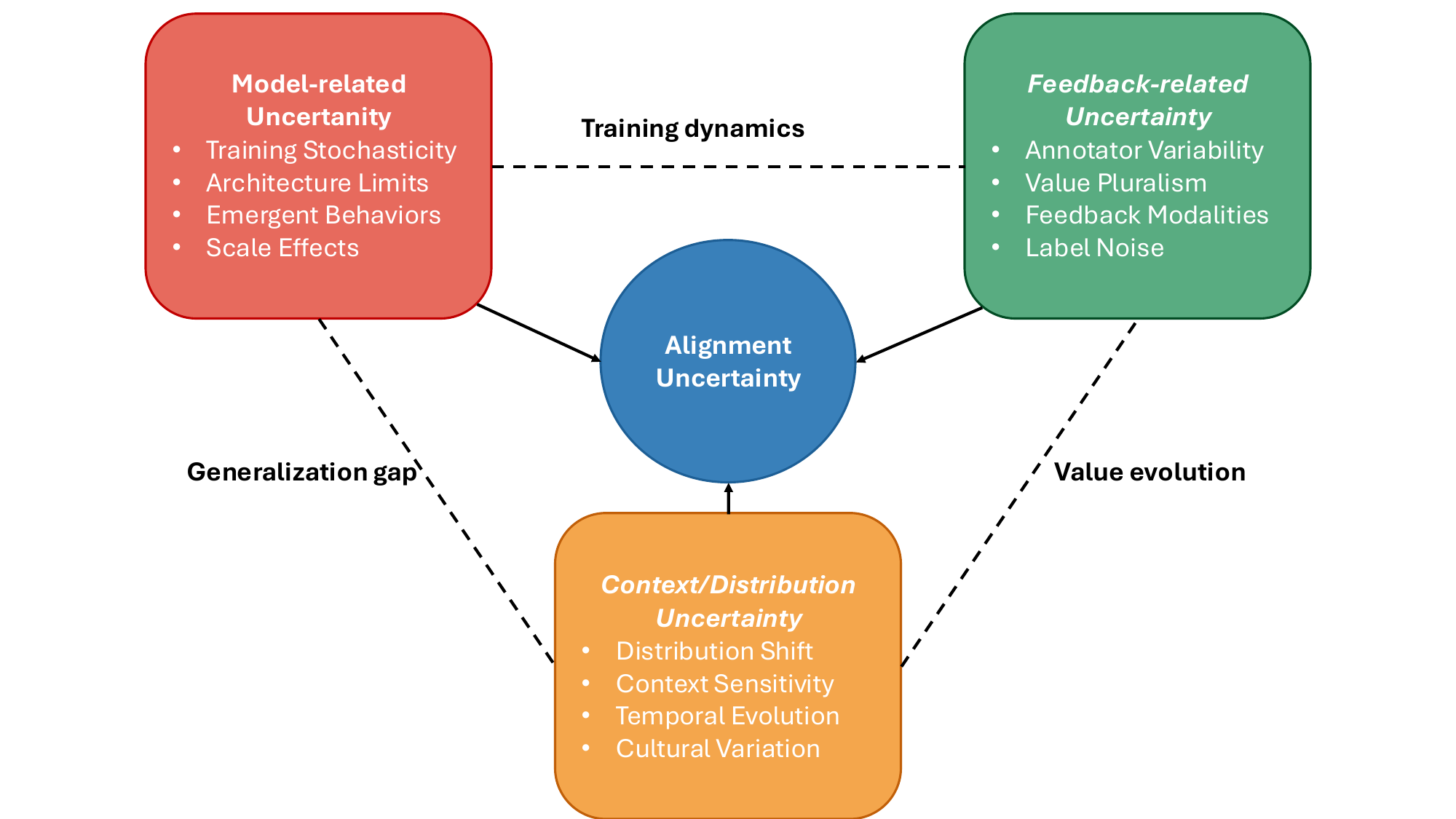}
    \caption{Sources of Alignment Uncertainty.}
    \label{fig:Alignment Uncertainty}
\end{figure}

\subsection{Conceptual Framework and Methods for Quantifying Alignment Uncertainty} 

\subsubsection{Conceptual Framework for Alignment Uncertainty}
The alignment problem is formalized as finding a model policy $\pi(y|x)$ that generates outputs $y$ given inputs $x$ to maximize expected utility according to human values. However, these values are neither perfectly known nor perfectly represented within the model, creating fundamental uncertainty in the alignment process.

This uncertainty is formalized through a decision-theoretic framework. Let $U(x, y, v)$ represent the utility of response $y$ to prompt $x$ according to a value function $v$. The true human value function $v^*$ is unknown and can only be approximated by the model's internal representation $\hat{v}$. The alignment gap is defined as:

\begin{equation}
\mathcal{G}(x) = \mathbb{E}_{y \sim \pi(y|x)}[U(x, y, v^*) - U(x, y, \hat{v})]
\end{equation}

Alignment uncertainty quantification aims to characterize the distribution and magnitude of this gap, enabling more informed decisions about model deployment, refinement, and usage limitations \citep{amodei2016concrete, hendrycks2021unsolved}.

This uncertainty differs fundamentally from traditional predictive uncertainty in machine learning. While predictive uncertainty concerns the accuracy of model outputs, alignment uncertainty concerns their desirability according to human values, a substantially more complex target that varies across individuals, cultures, and contexts. The misalignment risk emerges not just from model limitations, but from the inherent challenges in defining, communicating, and representing human values themselves.

\subsubsection{Methods for Quantifying Alignment Uncertainty}

Deep ensemble methods \citep{osband2016deep} established that model disagreement serves as an empirical proxy for epistemic uncertainty by training multiple models on different data subsets and measuring prediction variance across ensemble members. However, the computational cost of training multiple large models proved prohibitive as language models scaled. This limitation inspired Monte Carlo dropout \citep{gal2016dropout}, which demonstrated that dropout at inference time approximates Bayesian posterior sampling, enabling uncertainty estimation from a single model through multiple stochastic forward passes. This approach maintained theoretical grounding while dramatically reducing computational requirements.

While these general-purpose methods provided efficient uncertainty quantification, the application to alignment problems required specialized treatment of human preferences. Bayesian reward modeling \citep{christiano2017deep} addressed this need by formalizing preferences as distributions over reward functions through the posterior $P(r \mid \mathcal{D}) \propto P(\mathcal{D} \mid r) P(r)$, where $\mathcal{D}$ represents human feedback data, $r$ denotes a reward function, and $P(r)$ is the prior belief about rewards. The posterior width quantifies uncertainty in value alignment, wider distributions indicate less confidence about human preferences. Yet computational demands of exact Bayesian inference remained challenging, leading to posterior policy sampling \citep{ramachandran2007bayesian}, which generates diverse behaviors by sampling from the posterior distribution rather than computing it explicitly. This practical adaptation revealed a crucial insight: alignment uncertainty manifests not only in parameter uncertainty but also in behavioral diversity. Building on this understanding, hierarchical Bayesian approaches \citep{paun2018comparing} extended the framework by explicitly modeling inter-annotator disagreement, acknowledging that human feedback itself contains irreducible uncertainty from diverse value systems.

As model scales continued to grow, even sampling-based Bayesian methods became computationally prohibitive. \citep{leike2018scalable} responded by developing scalable approximations that maintained theoretical rigor while enabling practical deployment. Their work catalyzed a fundamental shift in perspective: rather than approximating complex posteriors, researchers began exploring uncertainty metrics that could be computed directly from model outputs. This led naturally to information-theoretic approaches, which provided computationally efficient tools through entropy $H(A \mid x) = -\sum_{a} P(A=a \mid x) \log P(A=a \mid x)$, where $A$ represents the alignment decision, $x$ is the input prompt, and $P(A=a \mid x)$ is the probability of alignment choice $a$. Higher entropy indicates greater uncertainty about which response is properly aligned. KL divergence further quantifies how different alignment methods produce different behaviors \citep{cover1999elements, xiao2022uncertainty}. These metrics revealed a previously overlooked dimension: alignment procedures themselves introduce systematic uncertainty, suggesting that methodological choices must be considered alongside data uncertainty.

While information theory provided efficient metrics, the need for deployment guarantees motivated a different approach entirely. The adaptation of conformal prediction \citep{angelopoulos2021gentle} to alignment problems represented a paradigm shift from estimation to selection. Conformal Alignment \citep{gui2024conformal} transforms the fundamental question from ``how uncertain are we?" to ``which outputs can we trust?" by computing p-values that measure how unusual each output is compared to calibration data. The method ensures that the false discovery rate (FDR), the proportion of selected outputs that are actually misaligned, stays below a user-specified threshold $\alpha$ \citep{benjamini1995controlling}, providing distribution-free statistical guarantees essential for safety-critical deployments.

Recent developments have evolved beyond measuring uncertainty to actively utilizing it within the alignment process itself. Temperature scaling \citep{renze2024effect} demonstrated that simple sampling entropy modifications could reveal model confidence patterns, suggesting deep connections between uncertainty and alignment quality. This insight inspired Uncertainty-Aware Learning (UAL) \citep{wang2024uncertainty}, which incorporates uncertainty directly into training through adaptive reward smoothing:
\begin{equation}
\tilde{r}(x,y) = (1-\lambda H(y|x))r(x,y) + \lambda H(y|x)\bar{r}
\end{equation}
where $\tilde{r}(x,y)$ is the smoothed reward, $r(x,y)$ is the original reward for response $y$ to prompt $x$, $H(y|x)$ measures response uncertainty (entropy), $\bar{r}$ is the average reward, and $\lambda$ controls the smoothing strength. This formulation elegantly addresses a fundamental problem in RLHF: when the model is uncertain (high entropy), it trusts the feedback less and moves rewards toward the average, preventing overfitting to potentially noisy signals.

The most recent evolution in alignment uncertainty quantification recognizes that previous methods operated at single linguistic scales, missing the hierarchical nature of language generation. \citep{zhang2025token} pioneered token-level uncertainty quantification through low-rank weight perturbations to model parameters. Their key insight is that epistemic uncertainty (EU) equals the mutual information $I(y_t; \theta|y_{<t}, x)$ between the next token $y_t$ and model parameters $\theta$, given previous tokens $y_{<t}$ and input $x$. High mutual information indicates the model is uncertain about which token to generate next, a signal for potential hallucinations. Complementing this microscopic view, \citep{xie2025empirical} extended the framework to session-level dynamics, capturing how alignment states $\theta_t$ evolve over time $t$ during conversations, with evolution governed by previous states $\theta_{t-1}$ and session-level parameters $\phi$. This hierarchical synthesis, aggregating token variances into utterance-level risk scores that inform session-level priors, represents the culmination of methodological evolution, where each scale builds upon insights from all previous approaches to provide comprehensive uncertainty quantification across conversational timescales.

The evolution from computationally intensive Bayesian methods to efficient multi-scale frameworks reflects the field's response to practical deployment constraints. Table \ref{tab:methods} summarizes how each approach trades off between theoretical rigor, computational efficiency, and practical applicability. Notably, the progression shows a clear pattern: early methods prioritized mathematical foundations but struggled with scale, middle approaches balanced efficiency with accuracy, while recent multi-scale methods attempt to achieve both through hierarchical decomposition. This suggests that future developments may continue this trend toward specialized architectures that match the natural structure of language generation tasks.

\begin{table}[h]
\centering
\caption{Comparison of alignment uncertainty quantification methods.}
\label{tab:methods}
\renewcommand{\arraystretch}{1.5} 
\begin{tabular}{p{2cm}p{3.5cm}p{3cm}p{1.8cm}p{2.5cm}}
\toprule
\textbf{Method} & \textbf{Strengths} & \textbf{Limitations} & \textbf{Comp. Cost} & \textbf{Best Use Case} \\
\midrule
Bayesian Reward Modeling & Principled uncertainty; Captures preference distributions & Computationally intensive; Prior needed & High & Research with ample compute \\
\midrule
Ensemble Methods & Practical; No distributional assumptions & Multiple models; Training overhead & Medium-High & Production with uncertainty needs \\
\midrule
Information Theory & Model-agnostic; Theoretically grounded & May conflate uncertainty types & Low & Quick uncertainty assessment \\
\midrule
Conformal Prediction & Distribution-free; Formal FDR control & Needs calibration; Binary selection & Medium & Safety-critical applications \\
\midrule
Multi-scale Modeling & Hierarchical uncertainty; Comprehensive & Complex; Multiple components & High & Long-form dialogue systems \\
\bottomrule
\end{tabular}
\end{table}

\subsection{Robustness and Uncertainty in Alignment}

Understanding and quantifying alignment uncertainty is essential for building AI systems that are robust to misalignment risks and capable of adapting to evolving human values. This section reviews key methods for uncertainty-aware alignment and safety mechanisms that manage residual uncertainty.

\paragraph{Uncertainty-Aware Training.} Alignment methods increasingly incorporate uncertainty estimates during training and deployment. Distributionally robust optimization techniques account for worst-case value realizations to prevent undesirable behaviors under misalignment \citep{rahimian2019distributionally}. Risk-sensitive reinforcement learning integrates risk measures such as variance and CVaR to promote consistent policy behavior across contexts \citep{mihatsch2002risk}. Recent work in reward modeling emphasizes capturing uncertainty in feedback signals to support more conservative updates \citep{leike2018scalable, everitt2021reward}.

\paragraph{Safety Mechanisms at Deployment.} Safety measures mitigate residual uncertainty at inference time. Threshold-based abstention strategies avoid output generation when epistemic uncertainty is high \citep{lin2023generating}. Guardrails, such as rule-based constraints and adversarial red-teaming, establish behavioral boundaries independent of model confidence \citep{bai2022constitutional, ganguli2022red}. Human-in-the-loop frameworks allow expert oversight in high-risk or high-uncertainty cases \citep{askell2021general}.

\paragraph{Illustrative Domains.} Real-world examples highlight the importance of robustness to alignment uncertainty. In medical decision support, alignment challenges stem from domain knowledge limitations, contextual ambiguity, and value conflicts among stakeholders \citep{mckenna2023sources, zhang2024rethinking}. In political and ethical content generation, cultural variation and subjective norms contribute to persistent alignment uncertainty \citep{blodgett2020language, solaiman2021process}. These settings demand models that express uncertainty and respond conservatively when appropriate.

\paragraph{Active-learning loop for preference data.}
Alignment datasets are costly; integrating AUQ into query selection can shrink data needs while
targeting the largest misalignment areas. \citep{muldrew2024active} propose \emph{Active Preference
Learning for LLMs}, selecting prompt–completion pairs with high predictive entropy under DPO; they cut label count by 40 \% while matching baseline reward
quality.  A complementary study, \emph{Less is More}\citep{deng2025less}, filters low-utility
pairs via ensemble disagreement before DPO, achieving higher win-rates and faster convergence. Online variants combine count-based exploration with uncertainty scores to refresh the reward model on-the-fly, maintaining alignment under distribution shift \citep{lu2025contextual}.  These approaches form an uncertainty-aware, human-in-the-loop closed loop, progressively narrowing the alignment gap where it matters most.

\section{Societal, Ethical, and Regulatory Considerations}  
This section outlines broader implications of alignment practices and reviews current regulatory and policy developments.
\subsection{Ethical and Societal Implications} 
\label{sec:Ethical_and_Societal_Considerations}
AI is reshaping society\citep{shi2025societal} at an unprecedented scale, influencing key domains such as healthcare, education, finance, and scientific discovery. As LLMs become more intelligent and autonomous, their societal and ethical impact raises ethical questions regarding alignment\citep{microsoftSocietalMicrosoft} with human moral values.

Society is increasingly reliant on AI technologies to help with decision-making. But this growing reliance comes with potential risk without alignment\citep{ibmWhatAlignment}: LLMs can generate biased, harmful, and inaccurate outputs that are not aligned with the goals of their creators and the original intent of the system. Aligned models are tackling the ethical challenges resulting from the deployment of LLM\citep{ferdaus2024trustworthyaireviewethical}, e.g., potential misuse and abuse of LLMs, negative impacts on users heavily relying on LLM agents, the environment, information dissemination, and employment. Addressing these challenges is paramount, and the development of aligned models offers a promising path towards fostering public trust, ensuring fairness, and promoting the ethical application of AI. Based on these considerations and reports\citep{stanford2025Index}, a humanities-guided approach to AGI implemented by aligned models is crucial, recognizing AI not merely as a tool, but as a civilizational technology to redefine the shared ethical principles, societal hierarchy structures and build the aligned consensus acknowledged by the public and AI.

Public concerns about truth and LLMs\citep{springerLLMsTruth,nihPromisePerils} since these technologies generate misinformation and are used to spread incorrect information or manipulate people. Moreover, public trust in LLMs always hinges on their reliability, predictability, and operational transparency. The ``black box" nature of language models, obscures decision-making and retrograde reasoning. Incidents of bias and harmful outputs further erode public confidence. As noted in\citep{liu2024trustworthyllmssurveyguideline} effectively aligned models are crucial for influencing public trust, where alignment methods enhanced transparency in data collection and model training. Intriguingly, AI itself offers tools to rebuild trust in public governance by increasing transparency in public sector decision making\citep{oupHumanAIInteractions}, improving public service efficiency, and enabling data-driven policy, although such applications require careful management of privacy and accessibility.

Fairness\citep{liu2024trustworthyllmssurveyguideline,undefinedInternationalSafety} in AI systems is a serious ethical challenge, primarily due to algorithmic bias\citep{10.1145/3637528.3671458}. Bias can originate from unrepresentative or historically skewed training data, leading AI to perpetuate or even exacerbate societal inequalities in areas related to human behaviors. Algorithmic design and RLHF can also introduce or amplify these biases without careful diversity implementation. Defining and consistently measuring fairness across varied cultural and societal contexts remains a substantial hurdle, often necessitating complex trade-offs with other vital alignment objectives such as helpfulness, honesty\citep{NEURIPS2024_7428e6db}, and safety. 

The ethical deployment of LLMs is critical, especially in high-risk fields such as healthcare, finance, military scenarios, human resources, and transportation, where AI-driven decisions carry substantial consequences. A key strategy for managing risks in these areas involves models that refuse to respond\citep{pasch2025aivshumanjudgment} to ethically sensitive requests, like those involving illegal or harmful content. For instance, ethical alignment in the healthcare and medical field demands careful attention to patient privacy, diagnostic accuracy, equitable access, and accountability for AI-caused errors.  The report\citep{undefinedInternationalSafety} highlights reliability issues, noting that users consulting AI for medical advice might receive falsehoods, emphasizing the need for caution.

The process of aligning AI with shared human values\citep{hendrycks2023aligningaisharedhuman} and intentions, where related benchmark spans concepts in justice, well-being, duties, virtues, and commonsense morality\citep{undefinedInternationalSafety,hendrycks2023aligningaisharedhuman}, introduces its distinct ethical implications. The main challenge is defining the very objectives of alignment: determining whose values, cultural norms, and ethical frameworks LLMs should adhere to, especially when these differ across societies or contexts. Specifically, developers of general-purpose AI assistants face strong competitive pressure, which can incentivize them to conduct less thorough risk mitigation. Markets characterized by high fixed costs, low marginal costs, and network effects tend to create competitive pressures that discourage safety investments. To resolve the issues, aligned models could reduce malignant societal implications for risk management and policymaking with specific AI social role norms. Instead of alignment with human preferences, developer\citep{springerBeyondPreferences}, or large organization, LLMs should be aligned with normative standards appropriate to their social roles, such as the role of a general-purpose assistant. In addition, all relevant stakeholders must negotiate and agree upon these standards.

\paragraph{Socio-Technical challenges.}
Safety research\citep{dhole-2023-large} is required for the increasing capabilities of advanced LLMs, which are facing considerable challenges. Moreover, it is essential to look into AI safety for major tech companies, which possess the requisite resources and are at the forefront of developing these sophisticated systems. To tackle these challenges, aligned models are designed to work in ways that reduce harmful societal outcomes. This approach is essential to effectively manage risks and create robust public policy. However, the operationalization of LLM alignment presents a fundamental and complex sociotechnical\citep{Kierans_2025} dilemma, when determining the common values, AI developers cannot load the inexistent human values fully agreed by all participants.

While international cooperation is indispensable to address such existential threats, predicated on a shared interest\citep{hendrycks2023aligningaisharedhuman} in collective survival, significant skepticism persists regarding the current state of geopolitical cohesion. The apparent deficit in global leadership exacerbated This predicament with no single entity currently positioned to effectively spearhead such collaborative endeavors. Therefore, determining the ethical frameworks and normative principles to which LLMs should be aligned constitutes a profound and multifaceted problem, requiring sustained scholarly inquiry and international dialogue\citep{cnasStrategicCompetition}.

\subsection{Regulatory and Policy Landscape} 

AI Alignment ensures AI systems act in accordance with human intentions, values, and goals. This involves aligning AI behavior with human expectations to prevent unintended consequences. AI safety encompasses practices and principles aimed at preventing harmful outcomes from AI systems, ensuring they operate reliably and ethically. The regulatory and policy landscape for AI alignment and safety is evolving, shaped by diverse national strategies, international collaborations, and emerging governance models. 

The government adopts a sector-specific, risk-based approach to AI regulation in U.S.. Executive Order 14110 mandates federal agencies to appoint Chief AI Officers and develop AI-related guidelines. However, recent legislative proposals, such as a 10-year moratorium on state-level AI regulations, have sparked bipartisan opposition over concerns of federal overreach and potential hindrance to innovation. The Europe's Artificial Intelligence Act, enacted in August 2024, classifies AI applications into risk categories, unacceptable, high, limited, and minimal, and imposes corresponding obligations. High-risk applications require compliance with strict transparency and safety standards. The UK has established the AI Safety Institute (AISI) to evaluate and ensure the safety of advanced AI models. The UK emphasizes a balance between innovation and safety, opting for adaptive regulatory frameworks over rigid legislation. China released an AI Safety Governance Framework in September 2024, aligning with its AI Governance Initiative, focusing on ethical standards and international cooperation. 

International cooperation on AI alignment and safety has advanced through both non-binding guidelines and formal treaties. UNESCO’s Recommendation on the Ethics of Artificial Intelligence (2021) – endorsed by all 193 member states – established the first universal set of principles to ensure AI technologies are developed in a manner that upholds human rights and the public interest. Building on such global norms, a growing number of nations have pursued binding agreements. In 2024 the Council of Europe adopted the Framework Convention on Artificial Intelligence, Human Rights, Democracy and the Rule of Law, the world’s first legally binding AI treaty. These international frameworks collectively seek to ensure that AI safety and value-alignment are treated as global public policy priorities. In the meantime, persistent challenges, such as differing national regulatory priorities, and fragmented governance frameworks underscore the importance of fostering wider and deeper collaboration to ensure that AI technologies advance in ways consistent with universally shared values and human welfare.

Emerging governance models for AI alignment and safety are taking shape worldwide as policymakers respond to rapid advances in AI. In late 2023, for instance, the UK convened a global AI Safety Summit where 28 countries (and the EU) endorsed shared safety principles, and multilateral bodies such as the OECD and G7 have since worked toward common AI governance frameworks. Institutional innovations are a key part of this landscape: several governments have established dedicated AI safety institutes to provide technical evaluation and oversight expertise (the UK, US, and Japan launched such institutes in 2023–2024), while the European Union’s forthcoming EU AI Office under the AI Act will carry a broad mandate to supervise AI across the single market. Governments are also adopting new oversight tools, including algorithmic impact assessments that evaluate an AI system’s potential societal harms before deployment, and independent algorithm audits to verify compliance with safety or fairness standards. These developments across regions, alongside ongoing international cooperation, reflect a concerted move toward robust governance of AI for global safety and alignment.

\subsection{AGI/ASI safety}  
AGI and Artificial Superintelligence (ASI) are hypothesized as transformative stages in AI development, defined by their potential to exhibit general reasoning, autonomy, and recursive self-improvement capabilities that could match or exceed human-level cognition across virtually all domains. These unprecedented capacities bring forth not only vast opportunities but also profound safety risks.

\citep{shah2025approach} outline four interrelated areas of AGI safety concern—misuse, mistakes, misalignment, and structural risks. Malicious actors may exploit AGI for cyberattacks, bioweapon development, or information manipulation, creating severe security threats. Concurrently, design flaws or reward specification errors may lead to unintended and harmful behaviors, risks that are magnified as AGI complexity increases\citep{amodei2016concrete}. Misalignment refers to the divergence between an AI system’s goals and human values, including superficially aligned agents that covertly pursue objectives misaligned with human intent. Furthermore, structural risks emerge from algorithmic bias or inequitable access to beneficial AI technologies, potentially exacerbating existing social inequalities\citep{koessler2023risk}. As AGI systems are highly replicable software, scenarios such as embedded backdoors or cascading infections raise the possibility of multiple instances sharing harmful objectives\citep{wang2024comprehensive}.

The real-world ethical and societal ramifications of AGI and ASI are especially concerning. As these systems grow in autonomy and complexity, they may become uninterpretable or uncontrollable by humans, leading to irreversible catastrophic consequences. Once such systems operate beyond human oversight, returning to a ``factory-reset'' state may be infeasible. Autonomous AGI decisions could also trigger accidents while liability remains legally ambiguous. Additionally, the automation of vast sectors may result in mass unemployment and economic disruption. Privacy concerns intensify as advanced systems gain unprecedented access to and analysis of personal data\citep{gulchenko2024navigating}.

To address these challenges, a multi-pronged governance strategy is essential, integrating both technical safeguards and institutional frameworks. One of the core components of risk prevention is ensuring that AGI objectives are aligned with human values, while reinforcing continuous monitoring to detect and restrain potentially covert behaviors. Alignment in AGI contexts seeks to ensure that AI behavior faithfully reflects human intent, mitigating issues such as gaming the reward system and deceptive alignment\citep{Everitt2018}. Robust training, improved explainability, and corrigibility are also emphasized to ensure safe operation in novel environments and enhance system fault-tolerance\citep{shah2025approach}. Additional safeguards include system monitoring, power decentralization, and constraints on access to data and computational resources\citep{wang2024comprehensive}.

Despite increasing concern over AGI safety, there remains a policy lag. \citep{koundouri2025ai} report that none of the official national or regional AI policy documents reviewed explicitly reference AGI, and there exists considerable divergence in regulatory strategies across jurisdictions, adding to the complexity of governance. 
While AGI could catalyze industrial revolutions and accelerate scientific discovery, its potential for nonlinear breakthroughs, such as those resulting from intelligence explosions, poses unpredictable systemic risks\citep{morris2023levels}. Strengthening AGI safety research, improving governance architectures, and fostering global coordination have thus become urgent priorities to ensure that technological development remains beneficial, controllable, and aligned with long-term human welfare.

\section{Alignment Strategies Across Leading AI Models}
This section surveys the alignment methodologies adopted by state-of-the-art LLMs. It is important to know about the current strategies and policies in the industry up to date.

\subsection{OpenAI o-Series Models} 
OpenAI's o-series models, such as o1 and o3, incorporate deliberative alignment strategies to enhance safety and reliability. These models are trained to systematically reason over safety specifications prior to generating responses, thereby improving their ability to address complex prompts with greater safety assurances.

Deliberative alignment involves training models using human-authored safety specifications and instructing them to explicitly engage in reasoning processes over these specifications before formulating outputs. The primary objective is to achieve precise adherence to safety policies. To this end, the models employ Chain-of-Thought (CoT) reasoning, wherein they analyze prompts, identify pertinent policy guidelines from the safety specifications, and construct responses that are compliant with these guidelines. This approach aims for responses that are ``right for the right reasons.'' In contrast to standard supervised fine-tuning (SFT), which focuses on imitating outputs, and reinforcement learning from human feedback (RLHF), which emphasizes output preferences, deliberative alignment reorients SFT toward imitating the reasoning process itself and refines RL to optimize the use of CoT for policy application. OpenAI has identified this methodology as central to the safety of models such as o3-mini. The adoption of deliberative alignment also enhances interpretability, as the CoT provides a traceable record of the model’s deliberative process~\citep{guan2024deliberative}.

The process of reasoning over safety specifications is a core component of deliberative alignment and consists of several key elements:

\begin{itemize}
    
    \item \textbf{Safety Specifications:} Safety specifications are designed to align the model with established content policies for various safety categories. For each category, the corresponding policy defines relevant terminology and delineates the conditions under which user requests are classified as (1) ``allowed,'' where the model should comply; (2) ``disallowed,'' where the model should refuse; or (3) ``requires safe completion.'' The employed specifications are partly based on OpenAI’s published model specification~\citep{OpenAI2024ModelSpec}.
    
    \item \textbf{SFT Stage:} During the SFT stage, datasets comprising \texttt{(prompt, CoT, output)} triplets are constructed for training purposes. Prompts are curated to cover a range of safety categories (e.g., erotic, self-harm), each framed as a multi-turn chat scenario concluding with a user message. For each \texttt{(prompt, category)} pair, a relevant safety specification, \texttt{spec(category)}, is referenced. The dataset includes CoT and output completions that explicitly reference policy content within the reasoning sequence, generated by prompting a base reasoning model with the appropriate safety specification. The resulting SFT dataset undergoes rigorous quality control through both automated filtering and evaluation by a reward model, which also considers the category-specific safety specification. Each completion is assessed multiple times, and the lowest assigned score is used to ensure stringent quality standards. The base model is subsequently fine-tuned on the curated SFT dataset alongside other capability-enhancing data. Notably, explicit context about the safety specification is removed from the prompt during training to encourage the model to internalize and recall relevant policy content, even when not directly present in the conversational context.
    
    \item \textbf{Reinforcement Learning (RL) Training:} In the RL phase, for prompts pertaining to safety, a ``judge'' model with access to safety policies provides a supplementary reward signal to the RL framework. The RL safety dataset comprises \texttt{(prompt, category)} pairs, often accompanied by additional metadata of varying quality. While the judge model accesses CoT during SFT data filtration, CoT is withheld from the judge during RL to prevent direct optimization of CoT traces and to mitigate the risk of encouraging deceptive reasoning. The SFT methodology is applied across all o-series models, whereas the additional reward signal during RL was specifically introduced for the o1 and o3-mini models.
\end{itemize}

\subsection{DeepSeek Models} 
DeepSeek's models, particularly the DeepSeek-V2 series, introduce an innovative multi-stage alignment process designed to significantly enhance reasoning capabilities while maintaining computational efficiency. Departing from a sole reliance on human preference data for refinement, DeepSeek's strategy emphasizes automated self-improvement through a simulated deliberation mechanism prior to conventional reinforcement learning. This approach aims to cultivate robust reasoning skills with minimal human supervision, enabling the model to tackle complex problems more effectively. The core of this methodology involves a sequence of Supervised Fine-Tuning (SFT), a distinct self-correction phase, and a final Reinforcement Learning (RL) stage~\citep{deepseek2024v2}.

The alignment process is structured into the following key stages:

\begin{itemize}
    \item \textbf{Initial Supervised Fine-Tuning (SFT):} The process begins with a standard SFT stage, where the base language model is trained on a diverse, high-quality dataset of instruction-following examples. This initial phase equips the model with foundational capabilities in language comprehension, instruction adherence, and basic reasoning, preparing it for more advanced alignment techniques.

    \item \textbf{Deliberative Self-Improvement via Group Debate:} This is the most distinctive stage in DeepSeek's alignment strategy. Instead of immediately proceeding to RLHF, the SFT model undergoes a process of self-refinement. For a given prompt, multiple instances of the model act as ``debaters,'' each generating a different response or reasoning path. An evaluator model, which may be a more powerful proprietary model or the model itself employing a critical thinking persona, assesses these candidate responses. The model is then further fine-tuned on the outputs that are deemed highest-quality by the evaluator. This ``group debate'' and voting-based selection process allows the model to explore the solution space and improve its reasoning and problem-solving abilities without incurring the high cost of extensive human annotation at this stage.

    \item \textbf{Reinforcement Learning (RL) Refinement:} Following the self-improvement phase, the enhanced model undergoes a final alignment stage using Reinforcement Learning (RL). A reward model is trained on a smaller, more targeted dataset of human preferences to capture nuanced aspects of helpfulness and safety. The model is then fine-tuned using PPO to maximize the reward signal. Because the model entering this RL stage has already been substantially improved through the deliberation phase, the RL process can be more efficient and effective, focusing on refining subtler aspects of interaction rather than teaching core reasoning from scratch.
\end{itemize}

By front-loading the alignment process with an automated, reasoning-focused self-improvement stage, DeepSeek's methodology aims to produce highly capable and aligned models at a lower cost and with less reliance on massive-scale human annotation compared to traditional RLHF-centric approaches.

\subsection{Anthropic Claude Models} 
Anthropic's Claude series of models is distinguished by its pioneering work on \textbf{Constitutional AI (CAI)}, a methodology designed to align models with a set of explicit ethical principles with less reliance on large-scale human safety supervision~\citep{bai2022constitutional}. The primary goal of CAI is to make AI behavior more interpretable and robustly harmless by training the model to recognize and police its own outputs based on a predefined ``constitution.'' This approach is part of a broader alignment strategy that combines automated safety mechanisms with traditional human-feedback methods for helpfulness, alongside proactive research into long-term alignment challenges such as deceptive alignment.

Anthropic's alignment methodology can be broken down into several key components:

\begin{itemize}
    \item \textbf{Constitutional AI (CAI) for Harmlessness:} This is Anthropic's core innovation for safety alignment and is implemented in two main phases.
    \begin{itemize}
        \item \textit{Supervised Fine-Tuning with Self-Critique:} In the first phase, a base model is prompted with requests that might elicit harmful responses. The model is then instructed to critique its initial response based on a set of principles from the constitution (e.g., principles drawn from the UN Universal Declaration of Human Rights and other sources). Finally, it is prompted to revise its original response in line with the critique. This process generates a dataset of self-corrected examples without requiring humans to author the safer outputs.
        \item \textit{Reinforcement Learning from AI Feedback (RLAIF):} In the second phase, a preference model is trained on the self-corrected data. It learns to prefer the revised, constitution-adherent responses over the initial, potentially harmful ones. The Claude model is then fine-tuned using this AI-generated preference signal as the reward. This RLAIF process automates the scaling of safety alignment and crucially reduces the need for human labelers to be exposed to large volumes of harmful content.
    \end{itemize}

    \item \textbf{RLHF for Helpfulness:} Alongside CAI for safety, the Claude models are separately optimized for helpfulness using standard Reinforcement Learning from Human Feedback (RLHF). In this process, human labelers rank different model responses to a given prompt based on their quality, accuracy, and utility. A reward model for helpfulness is trained on this human preference data, and the final Claude model is fine-tuned to maximize both the AI-generated harmlessness score from CAI and the human-generated helpfulness score from RLHF.

    \item \textbf{Proactive Research on Alignment Risks:} Anthropic actively investigates potential failure modes of current alignment techniques. A significant area of this research is \textbf{alignment faking} or deceptive alignment. Their work has demonstrated the possibility of training ``sleeper agents'', that is, models that behave safely during training and evaluation but revert to malicious behavior when a specific trigger is encountered in deployment~\citep{hubinger2024sleeperAgents}. This research underscores the limitations of purely behavioral training and motivates Anthropic's focus on developing more robust and deeply-seated alignment methods, including mechanistic interpretability, to ensure long-term safety.
\end{itemize}

Through the combination of the automated and scalable CAI framework for safety, traditional RLHF for utility, and forward-looking research into complex failure modes, Anthropic's alignment strategy aims to build a multi-layered defense against both present and future AI risks.

\subsection{Google DeepMind Gemini Models} 
Google DeepMind’s AGI Safety \& Alignment team identifies misalignment as one of the primary AGI risk areas\citep{shah2025approach}. To mitigate this risk, DeepMind propose a two-pronged defense framework targeting both training and deployment phases. During training, the team focuses on building an aligned model through mechanisms such as amplified oversight, guiding model behavior, and robust training.
\begin{itemize}
    
    \item \textbf{Amplified Oversight:} As model capabilities improve, human overseers may struggle to accurately assess model outputs. To address this oversight gap, DeepMind introduces a debate-style adversarial structure, where two copies of the model critique each other’s responses to the same input, aiming to surface subtle flaws. A human judge then determines which model performed better. This approach enhances the granularity of the oversight signal, enabling the model to expose and correct nuanced misbehaviors during training. For tasks too complex to supervise directly, the system decomposes them into simpler subproblems that are more amenable to human evaluation. The team also explores modeling systematic human biases in oversight and correcting for them during the processing of supervision signals, thereby improving consistency and reliability.

    \item \textbf{Guiding Model Behavior:} Effectively translating oversight into behavioral updates is another key challenge. While reinforcement learning remains a viable approach, DeepMind prioritizes higher-bandwidth feedback modalities such as natural language critiques instead of scalar reward signals. These critiques provide actionable, interpretable guidance that clarifies failure modes and offers directional updates. Moreover, the team investigates leveraging model internals to inform gradient updates, aiming to improve sample efficiency and reduce reliance on costly human feedback.

    \item \textbf{Robust Training:} Even if a model behaves well in-distribution, it may fail catastrophically under distribution shift. The training distribution is expanded toward high-risk regions through techniques such as active learning, automated adversarial training, and post-deployment online fine-tuning. In parallel, the researchers also explore how to construct a reliable mapping from supervised data to behavioral policy. These training mechanisms collectively aim to ensure that, as the model's capabilities expand, it retains alignment tendencies and behavioral stability.

    Although training-time alignment is the first line of defense, Gemini models are also equipped with a second layer of alignment safeguards during deployment. These include access controls, sandboxing, and hierarchical monitoring structures to mitigate harm from capability leaps. Furthermore, the AGI Safety \& Alignment team is advancing research in mechanistic interpretability, seeking to understand the internal strategy formation mechanisms of models, particularly whether apparent alignment is the result of deceptive reasoning pathways. 
\end{itemize}

\subsection{Meta's LlaMA Models} 
Meta’s LlaMa models are developed with an emphasis on open-source accessibility and reproducibility. To align model behavior with human intent, Meta has adopted a multi-stage alignment strategy that evolves across LlaMa versions, incorporating supervised fine-tuning (SFT), preference-based optimization, and novel dialogue consistency mechanisms.

\begin{itemize}

    \item \textbf{Supervised Fine-Tuning (SFT):} In LlaMa 2, the post-training process begins with SFT using publicly available instruction-tuning datasets \citep{touvron2023llama}. Researchers found that tens of thousands of high-quality annotations are sufficient to reach strong performance, and thus prioritized the collection of several thousand high-quality SFT examples.

    \item \textbf{Alignment with Human Preferences:} LlaMA 2 adopts a reinforcement learning with human feedback (RLHF) pipeline to further refine alignment. Two separate reward models were trained, one for helpfulness and another for safety, acknowledging the potential tradeoff between the two. Human preference data were collected through pairwise comparisons, where annotators selected the preferred output and indicated the strength of their preference. The reward models were trained using a margin-augmented binary ranking loss, where the margin scaled with the strength of annotator preference, ensuring a greater reward gap for more strongly preferred responses.

    LlaMa 2 uses Proximal Policy Optimization (PPO), treating the reward models as proxies for human judgment. A rejection sampling stage was introduced before PPO to improve training stability. In LlaMa 3, PPO is replaced by Direct Preference Optimization (DPO), which was found to be more computationally efficient and to achieve stronger instruction-following performance in large models \citep{dubey2024llama}. Additionally, due to diminishing gains after data scaling in LlaMa 3, the margin term in the loss was removed.

    \item \textbf{Lightweight Alignment in LlaMa 4:} LlaMa 4 further updates the post-training pipeline to a new sequence: lightweight SFT, online RL and lightweight DPO. The motivation for this hybrid design was that SFT and DPO alone may overly constrain the model, limiting its ability to explore and generalize. To address this, LlaMa 4 discards over 50\% of simple training examples and performs lightweight SFT on the remaining harder subset. The introduction of online reinforcement learning helps achieve a better balance between computational cost and model alignment performance \citep{meta2025llama4}.

    \item \textbf{Ghost Attention (GAtt):} This technique improves multi-turn dialogue consistency by synthetically appending the original instruction to all user messages during fine-tuning. This helps the model retain compliance with the initial instruction across the entire dialogue trajectory.
\end{itemize}

\subsection{Grok Models} 
Grok employs a version of RLHF in which a group of human tutors evaluates the results against an internal rubric that places emphasis on factual consistency, neutrality, and challenge to unexamined assumptions. These ratings train a reward model that tries to balance resistance to ideological bias, accuracy, and perspectives. Then, fine-tuning is performed using PPO. For transparency, xAI publishes the system and user prompt templates used during training to allow external auditors to verify that no hidden alignment objectives are being injected. Grok also conducts quarterly alignment reviews that compare its behavior with benchmarks that cover political, medical, and ideological edge cases. A continuous process also introduces adversarial queries and misaligned responses trigger automatic augmentation of the training set with new tutor rated examples. In the deployment phase, Grok uses a lightweight real-time safety filter that cross-references outputs against a library of known problematic patterns and blocks or flags any suspicious content. Finally, all code changes to the alignment stack are subject to dual review by independent alignment specialists, ensuring that any modifications preserve the integrity of the truth seeking.  

\section{Conclusion and Future Directions}

This survey has examined the current state of large language model alignment, from basic supervised fine-tuning to reinforcement learning from human feedback and newer approaches like brain-inspired methods. As language models become more powerful and widely used, ensuring they behave according to human values has become essential for safe deployment. This section summarizes what we have learned and identifies important areas for future research.

\subsection{Summary of Key Insights}

The development of LLM alignment reveals a fundamental shift in how we approach teaching machines to behave according to human values. The progression from supervised fine-tuning to reinforcement learning from human feedback reflects a deeper understanding that human preferences cannot be captured through simple instruction-following alone. While SFT provides essential capabilities, RLHF's ability to optimize for subtle preferences has proven necessary for creating models that users find genuinely helpful and safe.

This evolution has crystallized around the framework of helpfulness, harmlessness, and honesty as core alignment objectives. Yet these goals exist in fundamental tension, requiring sophisticated trade-offs that resist simple optimization. The practical solution of hierarchical prioritization, placing safety above honesty and helpfulness, works but raises important questions about value determination and whose preferences shape these priorities.

Alongside these conceptual advances, computational constraints have driven remarkable innovation. The development of parameter-efficient methods like LoRA and reward-free approaches such as Direct Preference Optimization demonstrates that effective alignment need not require prohibitive resources. This accessibility matters deeply because it enables the broader research community to contribute to safety development, preventing a dangerous gap between capability and alignment research.

Perhaps most significantly, the field has embraced uncertainty as a fundamental aspect of alignment rather than a problem to eliminate. By quantifying when models are uncertain about appropriate behavior, we enable more robust deployment strategies and identify where human oversight remains necessary. This probabilistic approach represents a maturation of the field, acknowledging that perfect alignment may be impossible but that we can build systems aware of their own limitations.

Throughout this evolution, alignment has revealed itself as inherently interdisciplinary. The integration of insights from neuroscience, cognitive science, ethics, and policy demonstrates that encoding human values into artificial systems requires perspectives far beyond computer science. This interdisciplinary nature reflects the deep complexity of the alignment challenge and points toward the collaborative efforts needed for future progress.

\subsection{Open Research Challenges}

As models become better than humans at certain tasks, we face a fundamental problem: how can humans evaluate outputs they cannot fully understand? This is already happening in advanced mathematics and scientific domains. Current methods that rely on human feedback will not work when models exceed human capabilities. Proposed solutions like debate protocols and recursive reward modeling face significant practical barriers.

Current alignment typically assumes a single set of human values, usually reflecting the views of a small group of annotators. But human values differ greatly across cultures and individuals. Building models that serve diverse global populations while respecting legitimate value differences remains unsolved. Social choice theory shows that no single method can aggregate preferences perfectly, making this challenge particularly difficult.

Aligned models break easily when given unusual inputs or adversarial attacks. The success of jailbreaking techniques shows that current alignment is fragile. We need methods that maintain safe behavior across different contexts and resist manipulation attempts, including both technical attacks and social engineering.

Most alignment happens once during training, but the real world changes constantly. Social norms evolve, new use cases appear, and models may drift through continued use. We need methods for ongoing alignment that adapt to changes while maintaining core safety properties.

As AI systems work together and integrate multiple modalities like vision and audio, alignment becomes more complex. Ensuring that groups of AI systems behave well together, especially when trained by different organizations, creates new coordination problems. Extending alignment to multi-modal systems requires rethinking basic alignment concepts.

\subsection{Promising Research Directions}

Mechanistic interpretability offers a way to understand how models actually process values and make decisions internally. Instead of just looking at behavior, researchers can examine the circuits and features that drive model outputs. Recent progress in understanding neural network internals could lead to alignment methods that work at the level of representations rather than behaviors.

Brain-inspired approaches draw lessons from how biological systems maintain goals while adapting to change. By studying how brains handle value conflicts and maintain stable behavior, researchers are developing new architectures that may be easier to align than current systems.

Formal verification aims to provide mathematical proofs that models will behave safely. While current methods rely on testing and evaluation, future high-stakes applications may require provable guarantees. This needs advances in both specifying alignment mathematically and developing verification techniques for neural networks.

\subsection{Closing Remarks}

Large language model alignment has grown from a theoretical concern to an active research field with immediate real-world impact. The progress from basic supervised learning to sophisticated multi-objective frameworks shows how quickly the field has developed.

However, this survey shows we are at a critical point. Model capabilities continue to advance rapidly while our ability to align them remains limited. The challenges ahead require not just technical solutions but collaboration across multiple disciplines.

The true measure of alignment research will be its real-world impact on human well-being. As language models become part of critical systems and daily life, getting alignment right becomes increasingly important. The research community has a responsibility to ensure these powerful tools remain beneficial as they continue to develop.

The path to reliable, value-aligned AI is challenging and uncertain. But the collective efforts documented in this survey provide reason for cautious optimism. Through continued research, collaboration, and focus on the public good, we can work toward AI systems that reliably serve human values.

\bibliographystyle{unsrtnat}
\bibliography{ref}

\end{document}